\documentclass[letterpaper]{article} 
\usepackage[preprint]{aaai2027}
\usepackage[hyphens]{url}  
\usepackage{graphicx} 
\usepackage[numbers,sort&compress]{natbib}
\usepackage{caption} 
\usepackage{algorithm}
\usepackage{algorithmic}
\usepackage{newfloat}
\usepackage{listings}
\DeclareCaptionStyle{ruled}{labelfont=normalfont,labelsep=colon,strut=off} 
\floatstyle{ruled}
\newfloat{listing}{tb}{lst}{}
\floatname{listing}{Listing}

\usepackage{booktabs}
\usepackage{colortbl}
\usepackage{multirow}
\usepackage{makecell}
\usepackage[table]{xcolor}
\usepackage{pifont}
\usepackage[accsupp]{axessibility}

\newcommand{\cmark}{\ding{51}}
\newcommand{\xmark}{\ding{55}}
\definecolor{waymogreen}{HTML}{00E89D}
\definecolor{waymolgreen}{HTML}{99F7D7}
\definecolor{waymollgreen}{HTML}{CCFAEB}
\definecolor{waymoblue}{HTML}{0077FF}
\definecolor{waymolblue}{HTML}{99B7FF}
\definecolor{waymollblue}{HTML}{CCE4FF}
\definecolor{waymolgray}{HTML}{F0F0F0}
\definecolor{mylightgray}{gray}{0.6}

\title{XEmbodied: A Foundation Model with Enhanced Geometric and Physical Cues for Large-Scale Embodied Environments}

\author{
Kangan Qian\textsuperscript{\rm 1,\rm 2}\equalcontrib,
ChuChu Xie\textsuperscript{\rm 1,\rm 2}\equalcontrib,
Yang Zhong\textsuperscript{\rm 2}\equalcontrib,
Jingrui Pang\textsuperscript{\rm 1},\\
Siwen Jiao\textsuperscript{\rm 3},
Sicong Jiang\textsuperscript{\rm 4},
Zilin Huang\textsuperscript{\rm 5},
Yunlong Wang\textsuperscript{\rm 1},
Kun Jiang\textsuperscript{\rm 1}\corresponding,\\
Mengmeng Yang\textsuperscript{\rm 1},
Hao Ye\textsuperscript{\rm 2}\corresponding,
Guanghao Zhang\textsuperscript{\rm 2},
Hangjun Ye\textsuperscript{\rm 2},
Guang Chen\textsuperscript{\rm 2},\\
Long Chen\textsuperscript{\rm 2},
Diange Yang\textsuperscript{\rm 1}\corresponding
}

\affiliations{
\textsuperscript{\rm 1}Tsinghua University\quad
\textsuperscript{\rm 2}Automotive and Robotics, Xiaomi Corporation\\
\textsuperscript{\rm 3}National University of Singapore\quad
\textsuperscript{\rm 4}McGill University\quad
\textsuperscript{\rm 5}University of Wisconsin--Madison\\
qka8203191007@163.com\quad xcc22@mails.tsinghua.edu.cn \quad zhongyangtony@163.com
}

\begin{document}
\maketitle
{\renewcommand{\thefootnote}{}%
\footnotetext{\url{https://github.com/xiaomi-research/xiaomi_spatialembodied}}}

\begin{abstract}
The development of next-generation autonomous systems increasingly relies on Vision--Language--Action models, whose training requires scalable, high-quality annotations of complex embodied environments. Current cloud-side annotation pipelines rely heavily on generic vision–language models, which suffer from a fundamental mismatch: pretrained on flat, internet-scale 2D image–text data, they lack both the geometric reasoning and domain semantics required for embodied understanding. We introduce XEmbodied, a cloud-side foundation model that bridges this gap by endowing VLMs with intrinsic 3D geometric awareness and the ability to integrate physical cues such as occupancy grids, 3D boxes, and BEV maps. Rather than treating geometry as auxiliary input or fine-tuning naively, XEmbodied introduces a 3D Adapter for structured token-level integration of 3D representations, and an Efficient Image-Embodied Adapter that distills physical signals into interpretable context tokens. With a progressive domain curriculum and reinforcement-learning post-training, 
XEmbodied preserves general knowledge while achieving robust performance across 16 public benchmarks, demonstrating significant gains in spatial reasoning, traffic semantics, and out-of-distribution generalization for reliable large-scale scenario mining and VQA in embodied tasks. We release XEmbodied under the Xiaomi-SpatialEmbodied project.

\end{abstract}

\section{Introduction}

\begin{figure}[!t]
\centering
\includegraphics[width=0.98\columnwidth]{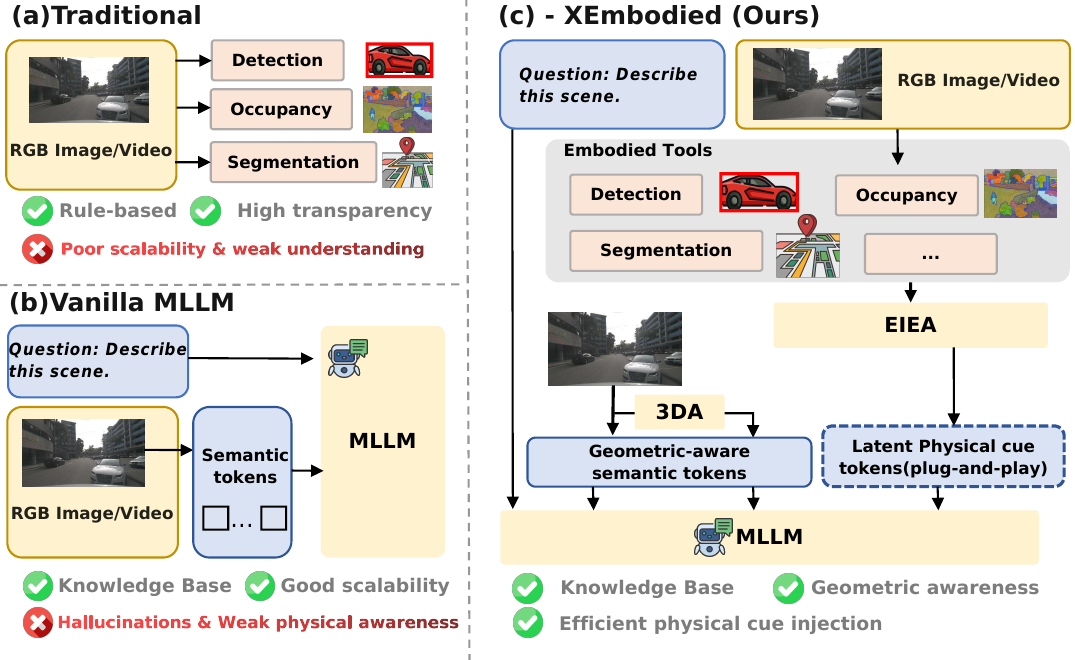} 
\caption{Comparison of three embodied scene understanding paradigms.
    (a)~Traditional rule-based pipelines offer high transparency but poor scalability.
    (b)~Vanilla MLLMs have strong general reasoning but suffer from hallucinations and weak geometric awareness.
    (c)~XEmbodied (ours) integrates geometric priors and embodied evidence using tools like detection, occupancy and map segmentation with our 3DA and EIEA.
  }
\label{fig:motivation}
\end{figure}



Autonomous driving and embodied AI are rapidly evolving toward foundation-model-based intelligence, where Vision-Language-Action (VLA) models provide a promising paradigm for scalable perception, reasoning, and interaction. The development of such models relies heavily on a closed-loop data system: large-scale real-world experiences are collected from vehicles or robots, uploaded to cloud platforms for scenario discovery and structured annotation, and subsequently used to improve foundation models and downstream autonomy systems~\cite{dong2026driving}. This cloud-side loop has become a key infrastructure for transforming massive long-tail embodied experiences into high-value learning signals~\cite{li2024data}.

However, current cloud-side annotation pipelines remain insufficient for understanding complex embodied scenarios. Industrial systems mainly depend on weak metadata, such as temporal information, vehicle states, and GPS signals, together with lightweight perception outputs including object detection, map segmentation, and localization~\cite{jia2024diffmap}. Although effective for routine cases, these signals lack the semantic and spatial reasoning ability required for safety-critical scenarios, where correct interpretation depends simultaneously on domain knowledge (e.g., traffic rules and driving semantics) and precise geometric understanding. Consequently, a large amount of collected embodied data remains underutilized, limiting the scalability of foundation models and autonomous systems.

Recent advances in Multi-Modal Large Language Models (MLLMs) provide a potential solution by introducing strong visual reasoning and language-based understanding into embodied tasks. Nevertheless, directly applying general-purpose VLMs to embodied environments remains challenging. Models trained primarily on internet-scale image-text data lack intrinsic understanding of physical structures, such as spatial relations, topology, and occupancy constraints. Moreover, domain adaptation may lead to catastrophic forgetting, while native embodied signals, including 3D boxes, trajectories, maps, and occupancy representations, are difficult to integrate into conventional VLM architectures. These limitations create a fundamental gap between semantic reasoning and physical understanding, resulting in unreliable and hallucination-prone reasoning in complex environments.

Human spatial cognition provides an important perspective for bridging this gap. Humans do not rely solely on external measurements; instead, they maintain an internal spatial representation and flexibly combine it with external, verifiable physical cues, such as relative distance, motion patterns, and environmental landmarks, when necessary~\cite{farzanfar2023cognitive}. Inspired by this mechanism, we argue that embodied foundation models should possess two complementary capabilities: an \emph{endogenous} geometric understanding that provides a stable spatial foundation, and an \emph{exogenous} evidence interface that incorporates reliable physical observations. Existing approaches typically treat these two components separately, either appending geometric information as auxiliary inputs or exposing raw tool outputs to the language model, without establishing a unified mechanism that allows internal spatial knowledge and external evidence to interact.

To address this limitation, we propose \textbf{XEmbodied}, a cloud-side embodied foundation model designed for large-scale autonomous driving and robotics VQA. XEmbodied introduces a complementary geometric-evidence architecture composed of a \textbf{3D Adapter (3DA)} and an \textbf{Efficient Image-Embodied Adapter (EIEA)}. The 3DA injects geometric representations from a 3D foundation model into the MLLM through dedicated geometric tokens, enabling endogenous 3D reasoning while preserving the semantic capability of the original foundation model. Meanwhile, EIEA converts heterogeneous physical observations, including detection results, occupancy information, and map cues, into compact physical cue tokens that provide verifiable external evidence without inefficient tool-language interaction. Different from existing auxiliary-input paradigms, XEmbodied jointly models internal geometric competence and external physical evidence as a coupled interface, allowing the two sources of information to complement each other. Furthermore, we introduce a progressive domain curriculum and a multi-stage training strategy to improve domain adaptation while maintaining generalization ability.

Throughout this report, \emph{XEmbodied} denotes the released model, while \emph{Xiaomi-SpatialEmbodied} denotes the project under which it is released. The SFT, Full, and FT suffixes used in the experiments identify training checkpoints or downstream adaptations of XEmbodied rather than separately released models.

Extensive experiments on 16 embodied benchmarks demonstrate that XEmbodied achieves strong and consistent improvements in complex scene understanding. In particular, the model shows robust generalization on strictly out-of-distribution datasets, validating the effectiveness of combining endogenous geometric reasoning with exogenous physical evidence for scalable embodied intelligence.

Our contributions are summarized as follows:
\begin{itemize}
    \item We introduce \textbf{XEmbodied}, a cloud-side embodied foundation model that enables scalable autonomous driving and robotics understanding through unified semantic, geometric, and physical reasoning.
    
    \item We propose a complementary geometric-evidence architecture consisting of \textbf{3DA} and \textbf{EIEA}, where 3DA establishes endogenous 3D grounding and EIEA provides compact, verifiable physical cues, forming an effective interface between perception and language reasoning.
    
    \item We develop a progressive curriculum and multi-stage training framework for robust embodied adaptation, achieving state-of-the-art performance across 16 benchmarks with strong out-of-distribution generalization.
\end{itemize}

\begin{figure*}[!t]
  \centering
  \includegraphics[width=0.95\textwidth]{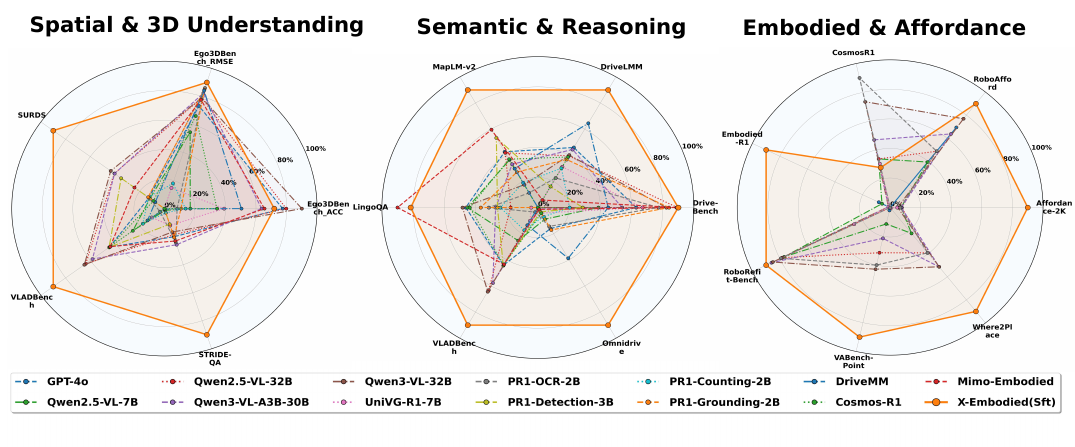}
  \caption{Quantitative comparison across three core dimensions of embodied understanding. Our XEmbodied (SFT) model consistently outperforms state-of-the-art baselines across 16 public benchmarks.}
  \label{fig:data-radar}
\end{figure*}

\section{Related Work}
\label{sec:related_work}

\subsection{Visual Question Answering in Data Closed-Loop Systems}
\label{subsec:rw_vqa_closedloop}

Visual Question Answering (VQA) in data closed-loop systems evaluates model understanding of embodied scenes, spanning traffic-rule reasoning, interactive dynamics, and planning semantics. Benchmarks such as BDD-X~\cite{kim2018textual}, nuScenes-QA~\cite{qian2024nuscenes}, DriveMLLM~\cite{guo2024drivemllm}, DriveBench~\cite{xie2025vlms}, and DriveLMM-o1~\cite{ishaq2025drivelmm} provide structured QA protocols, while methods such as Reason2Drive~\cite{nie2024reason2drive}, AlphaDrive~\cite{jiang2025alphadrive}, OmniDrive~\cite{wang2024omnidrive}, and DriveCoT~\cite{wang2024drivecot} incorporate chain-of-thought rationales. Closed-loop deployment additionally requires scalable scenario mining, annotation faithfulness under physical constraints, and robustness to distribution shifts. We therefore treat embodied VQA as an annotation-engine problem: the model must jointly support geometric understanding, physically verifiable reasoning, and scalable multi-task inference.

\subsection{Language Models in Embodied Tasks}
\label{sec:rw_llm_ad}

Large language models (LLMs) have emerged as powerful reasoning engines for embodied tasks by providing commonsense priors and flexible language-based reasoning. Early LLM-driven driving systems reformulated perception and planning into text-based pipelines for zero/few-shot generalization~\cite{xu2024drivegpt4, ma2024dolphins}, but suffered from limited geometric understanding and hallucination under out-of-distribution scenarios~\cite{wang2023drivemlm, ishaq2025drivelmm}. Recent methods improve robustness through multimodal perception and structured reasoning: DriveVLM~\cite{tian2024drivevlm} introduces chain-of-thought reasoning, DriveLM~\cite{sima2024drivelm} adopts graph-based QA, and RoboTron-Drive~\cite{2024RoboTron} enables multi-task learning. Beyond driving, OpenVLA~\cite{kim2024openvla}, Embodied-R1~\cite{yuan2025embodied}, MiMo-Embodied~\cite{hao2026mimo}, and Cosmos-Reason1~\cite{azzolini2025cosmos} extend foundation models toward robotic manipulation and embodied reasoning. However, most existing systems remain task-specific or rely on limited data, restricting scalable cloud-side deployment. In contrast, XEmbodied adopts a foundation-model backbone with heterogeneous multi-source data to support unified embodied intelligence.

\subsection{Geometric and Physical Cues Injection in Embodied Tasks}
\label{sec:rw_spatial_vlm}

Despite strong visual understanding capabilities~\cite{bai2025qwen2, chen2024internvl}, VLMs still struggle with geometric and physical reasoning~\cite{ray2024sat, cheng2024spatialrgpt, yang2025thinking}. Existing methods improve geometric grounding through structured 3D representations~\cite{ha2022semantic}, RGB-D and scene graph inputs~\cite{cheng2024spatialrgpt}, and tool-assisted estimation~\cite{cai2025spatialbot, qian2025agentthink}. Other approaches strengthen reasoning through 3D intermediate representations~\cite{ma2025spatialreasoner}, iterative tool refinement~\cite{wu2025reinforcing, qian2025agentthink}, and reinforcement learning~\cite{pan2025metaspatial, huang20253d}. However, these methods typically rely on auxiliary inputs, explicit reasoning traces, or task-specific pipelines, without establishing a compact and verifiable interface between physical perception and language reasoning. XEmbodied instead combines endogenous geometric integration through a cross-attentive 3D Adapter with distilled external physical cues for efficient embodied reasoning. A more comprehensive discussion is provided in Related Work.


\section{Method}
\label{sec:method}

\begin{figure*}[!t]
  \centering
  \includegraphics[width=0.9\textwidth]{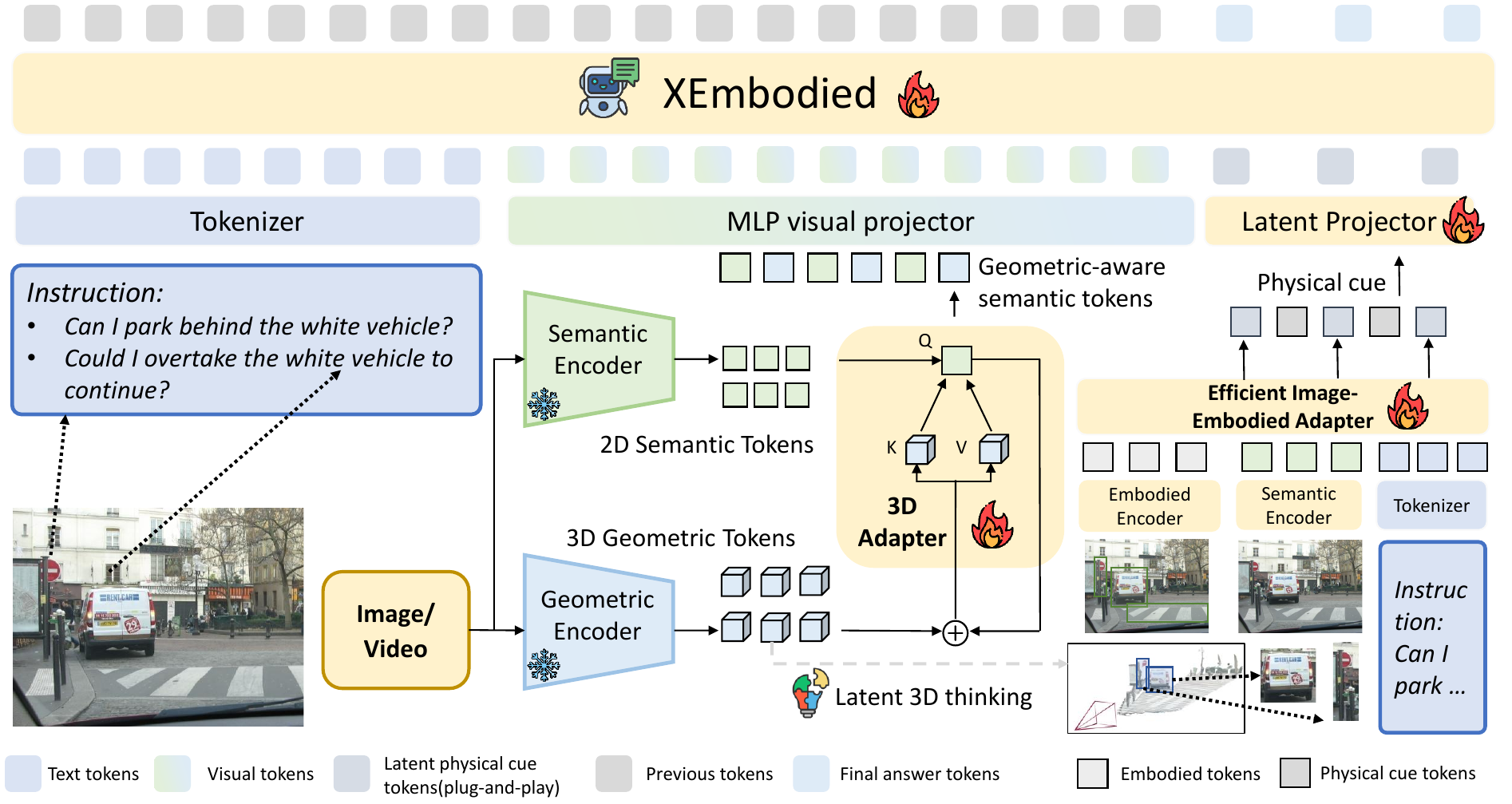}
  \caption{Overall architecture of XEmbodied.
    Given an image/video and an embodied instruction, XEmbodied first extracts 2D semantic tokens and 3D geometric tokens via dedicated encoders.
    Our 3DA fuses these tokens to form latent 3D thinking.
    The EIEA then distills physical cues from embodied tool outputs into compact tokens, which are seamlessly injected into the MLLM context for robust physical-augmented reasoning.
  }
 \label{fig:framework}
\end{figure*}

MLLMs excel at general vision-language tasks but often underperform in embodied environments (autonomous driving, robotics): 2D-dominated pretraining misses 3D geometry, metric distance, and spatial interactions, so grounded reasoning becomes unreliable.
To address this, our framework has three modular components (Fig.~\ref{fig:framework}):
(a) \textbf{3D Adapter (3DA)}: injects 3D tokens into the 2D semantic stream via cross-attention (Sec.~\ref{sec:3da});
(b) \textbf{Efficient Image-Embodied Adapter (EIEA)}: aligns heterogeneous modalities and distills tool-augmented evidence into compact tokens (Sec.~\ref{sec:eiea});
(c) \textbf{Progressive Domain Curriculum}: strengthens domain adaptation while curbing forgetting (Data and Curation).
3DA and EIEA are designed to be \emph{complementary rather than redundant}: 3DA supplies \emph{endogenous, dense, continuous} 3D geometry from raw RGB/video, whereas EIEA supplies \emph{exogenous, sparse, discrete} physical evidence distilled from off-the-shelf perception tools. The two streams operate at different abstraction levels, and they are coupled rather than stacked, since removing 3DA while keeping EIEA regresses performance below the vanilla baseline (Sec.~\ref{sec:exp}).

\textbf{Embodied VQA task formulation.}
Given a visual observation $X_v$, a textual instruction $X_T$, and auxiliary driving modalities $X_m$,
we learn an embodied vision-language model that produces an output $Y$:
\begin{equation}
\label{eq:overall}
Y = f_{\text{XEmbodied}}(X_v, X_T, X_m),
\end{equation}
where $f_{\text{}}(\cdot)$ denotes \textbf{XEmbodied}.
Here $X_v$ is an RGB image or video frames, $X_T$ is the user prompt, and $X_m$ holds embodied-native signals such as BEV occupancy, detection, and map segmentation.
$Y$ is the final answer.

XEmbodied consists of three core components: 3D Adapter (3DA), Efficient Image-Embodied Adapter (EIEA), and progressive domain curriculum training. The two adapters are complementary rather than redundant: 3DA builds intrinsic spatial representations from raw visual input, while EIEA imports verifiable physical signals from external perception modules.

\subsection{3D Adapter}
\label{sec:3da}

To give standard VLMs embodied capabilities, we propose a 3D Adapter (3DA) that folds 3D geometric priors into the 2D semantic token stream. Inspired by how human cognition separates \emph{semantic} and \emph{geometric} processing, 3DA keeps two complementary streams:

\noindent\textbf{Semantic Stream.} A 2D visual encoder (e.g., Qwen3-VL~\cite{bai2025qwen3}) extracts semantic tokens $\mathbf{e}_{\text{2D}} \in \mathbb{R}^{N \times d_{\text{2D}}}$ that capture high-level object categories and scene attributes.

\noindent\textbf{Geometry Stream.} A 3D geometry encoder based on VGGT~\cite{wang2025vggt} captures spatial structures. Using its pre-trained encoder and cross-frame fusion decoder, this stream produces 3D geometry tokens $\mathbf{e}_{\text{3D}} \in \mathbb{R}^{M \times d_{\text{3D}}}$ that carry rich spatial priors without requiring explicit 3D attributes.

\noindent\textbf{Token Alignment and Fusion.} To resolve the resolution mismatch between streams, we first map $\mathbf{e}_{\text{3D}}$ onto the 2D token grid with a deterministic interpolation operator: $\tilde{\mathbf{e}}_{\text{3D}} = \text{Interpolation}(\mathbf{e}_{\text{3D}}) \in \mathbb{R}^{N \times d_{\text{3D}}}$. Projecting $\tilde{\mathbf{e}}_{\text{3D}}$ to dimension $d_{\text{2D}}$ gives $\hat{\mathbf{e}}_{\text{3D}}$.
We inject these geometric cues through cross-attention (CrossAttn), with $\mathbf{e}_{\text{2D}}$ as queries ($\mathbf{Q}$) and $\hat{\mathbf{e}}_{\text{3D}}$ as keys ($\mathbf{K}$) and values ($\mathbf{V}$):
\begin{equation}
\label{eq:cross_attn_fusion}
\mathcal{E}_{\text{2D3D}} = \text{CrossAttn}\left(\mathbf{Q}=\mathbf{e}_{\text{2D}}, \mathbf{K}=\hat{\mathbf{e}}_{\text{3D}}, \mathbf{V}=\hat{\mathbf{e}}_{\text{3D}}\right).
\end{equation}
A residual connection preserves the foundational 2D semantic features in the final representation $\mathbf{e}_{\text{fusion}}$:
\begin{equation}
\label{eq:residual_fusion}
\mathbf{e}_{\text{fusion}} = \mathbf{e}_{\text{2D}} + \mathcal{E}_{\text{2D3D}}.
\end{equation}
This fused representation $\mathbf{e}_{\text{fusion}}$ replaces the original 2D tokens as input to the language model for downstream embodied reasoning. Our implementation uses a white letterbox to $518\times518$ for the VGGT stream, keeps the Qwen stream at the original aspect ratio, and uses bilinear interpolation for grid alignment. Because the fused representation is added back to the native image embeddings, the model does not increase the original visual sequence length. Full configurations are given in Section~\ref{sec:implementation}.

\subsection{Efficient Image-Embodied Adapter}
\label{sec:eiea}
To fold heterogeneous embodied modalities (e.g., detection~\cite{wang2023sgfnet}, map segmentation~\cite{jia2024diffmap}, BEV occupancy~\cite{qian2025lego, shi2024streamingflow}) into VLM reasoning, we propose EIEA.
It produces plug-and-play compact physical cue tokens that encode physical information and embodied-native signals in a VLM-compatible form.
The pipeline has three steps: (i) modality-specific feature extraction, (ii) a Mamba-based interpreter for joint embodied reasoning, and (iii) distillation-based compression.

\textbf{Modality-specific feature extraction.}
Let the auxiliary modalities be $\mathcal{M}=\{m\}$ and each raw modality input be $X_m$.
As shown in Fig.~\ref{fig:framework}, a modality encoder $\mathcal{E}_m$ (an embodied encoder plus a vision encoder) extracts token features, and a lightweight projector $\phi_m$ maps them to a shared dimension $d$:
$\mathbf{H}_m = \mathcal{E}_m(X_m) \in \mathbb{R}^{L_m \times d_m}$, $\mathbf{Z}_m = \phi_m(\mathbf{H}_m) \in \mathbb{R}^{L_m \times d}$.
In practice $\mathcal{E}_m$ can use off-the-shelf encoders (e.g., ResNet-50~\cite{he2016deep} and SigLIP~\cite{zhai2023sigmoid} for camera-derived visual cues), and $\phi_m$ is a two-layer MLP with GELU~\cite{hendrycks2016gaussian}.

\textbf{Mamba-based multimodal interpreter.}
Given query tokens $\mathbf{X}_T \in \mathbb{R}^{L_T \times d}$, we concatenate the projected modality tokens and text tokens into one sequence and feed it to a lightweight Mamba-based interpreter $g_{\text{Mamba}}$:
$\mathbf{H}_{\text{text}} = g_{\text{Mamba}}\big(\big[\mathbf{X}_T;\mathbf{Z}_{m_1};\dots;\mathbf{Z}_{m_{|\mathcal{M}|}}\big]\big)$.
A latent projection module $\phi_R$ then maps these latent rationale tokens into another semantic embedding space, matching the word-embedding dimensionality of the downstream VLM: $\mathbf{Z}_{\text{phy}} = \phi_R(\mathbf{H}_{\text{text}})$.

\textbf{Distillation-based compression.}
Passing long tool outputs directly into the VLM context is inefficient and often harms reasoning.
We therefore distill the interpreter outputs into compact physical cue tokens $\mathbf{Z}_{\text{phy}}$ that summarize intermediate rationales and verifiable embodied cues.
We then condition the backbone VLM on $(X_v, X_T, \mathbf{Z}_{\text{phy}})$ to generate the answer:
\begin{equation}
\label{eq:final_gen}
Y = f_{\text{VLM}}\big(X_v, X_T, \mathbf{Z}_{\text{phy}}\big).
\end{equation}
Detailed architectural configurations and distillation procedures of EIEA are provided in Section~\ref{sec:implementation}.

\begin{figure}[!t]
\centering
\includegraphics[width=0.98\columnwidth]{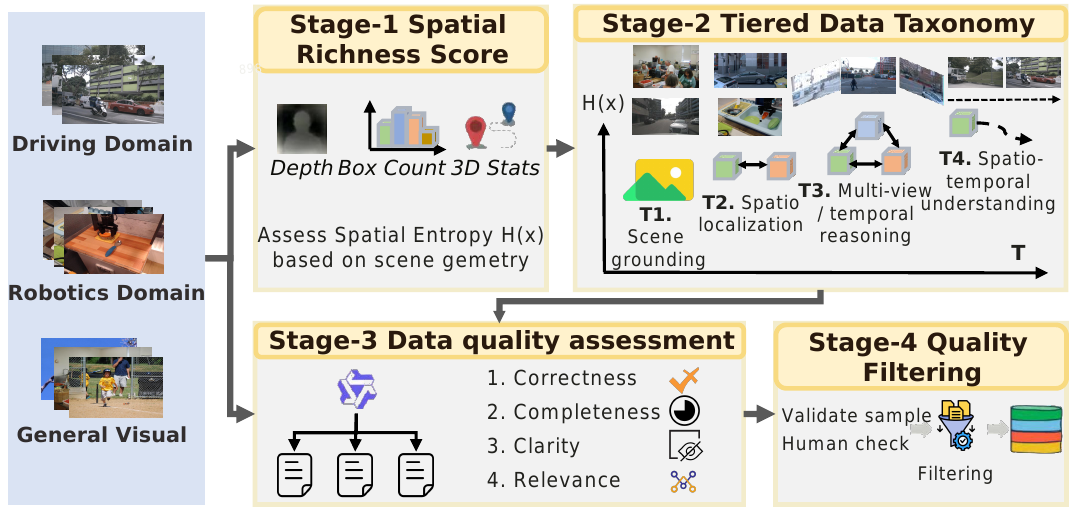} 
\caption{Our four-stage data curation pipeline for embodied closed-loop learning.
  }
\label{fig:data-class}
\end{figure}
The complete data-curation formulation, including spatial entropy, tiered data taxonomy, and automatic quality assessment, is provided in Section~\ref{sec:data_curation}; representative examples are retained in Appendix~\ref{app:cases}.

\subsection{Training Pipeline}
\label{subsec:training}
As shown in Fig.~\ref{fig:data-class}, XEmbodied adopts a four-stage progressive training pipeline to strengthen domain-specific semantic and 3D spatial capabilities while preserving general VLM performance. The first three stages share the next-token prediction objective, varying only in data mixture and trainable components: Stage 1 aligns the model with driving and robotics semantics via QA fine-tuning with vision and 3D modules frozen; Stage 2 trains the 3D Adapter on spatial-dominant data to integrate geometric information into language representations; Stage 3 jointly tunes the 3D Adapter and LLM LoRA with mixed multi-type data to unify 3D perception and linguistic generation. Based on the Stage-3 checkpoint and using the original training dataset without extra data, Stage 4 optimizes the plug-and-play EIEA module to encode multi-modal physical cues into language-compatible physical tokens. It further enhances output fidelity via GRPO reinforcement learning with a unified Foundation-ORM reward (0.2 for format regularization and 0.8 for answer correctness), along with KL divergence constraint against the Stage-3 baseline model. Detailed curriculum construction and stage-wise optimization settings are provided in Sections~\ref{sec:data_curation} and~\ref{sec:training_details}, respectively.

\section{Experiments}
\label{sec:exp}

Our experiments answer three questions:
(1) Does XEmbodied improve understanding on autonomous driving and robotics VQA benchmarks?
(2) How do endogenous geometric representation and physical cues contribute to spatial awareness and correct embodied reasoning?
(3) Why are data curation and each training stage essential?
We implement all experiments in the \texttt{ms-swift}~\cite{zhao2024swiftascalablelightweightinfrastructure} training framework, which supports large-scale training on 128 NVIDIA H20 GPUs. We use Qwen3-VL-30B-A3B-Instruct as the pre-trained model~\cite{bai2025qwen3} and adopt AdamW~\cite{loshchilov2017decoupled} optimization with gradient accumulation for training stability, with full training details provided in Training Details.

\begin{table*}[t]
\centering
\scriptsize
\setlength{\tabcolsep}{1.2pt}
\renewcommand{\arraystretch}{0.95}
\resizebox{\textwidth}{!}{%
\begin{tabular}{@{}l*{10}{c}@{}}
\toprule
\multirow{2}{*}{\textbf{Model}}
& \multicolumn{5}{c}{\textbf{Spatial \& 3D Understanding}}
& \multicolumn{5}{c}{\textbf{Semantic \& Reasoning}} \\
\cmidrule(lr){2-6}\cmidrule(lr){7-11}
& \makecell{Ego3D\\ACC$\uparrow$}
& \makecell{Ego3D\\RMSE$\downarrow$}
& SURDS
& \makecell{VLAD\\Bench}
& \makecell{STRIDE\\-QA}
& \makecell{Drive\\Bench}
& \makecell{DriveLMM\\-o1}
& \makecell{MapLM\\-v2}
& LingoQA
& Omnidrive \\
\midrule

\multicolumn{11}{l}{\textbf{Proprietary Models}} \\
GPT-4o~\cite{openai2024gpt4ocard}
& 52.70 & 19.20 & 13.31 & 56.00 & \underline{14.70}
& 47.97 & 57.84 & 57.81 & 53.58 & 4.54 \\
Gemini-1.5~\cite{geminiteam2024gemini15unlockingmultimodal}
& 53.50 & 19.62 & 32.77 & 54.23 & $-$
& $-$ & $-$ & 58.44 & 61.30 & 2.73 \\
\midrule

\multicolumn{11}{l}{\textbf{Open-Source Models}} \\
Qwen2.5-VL-7B~\cite{bai2025qwen2}
& 41.10 & 30.36 & 12.61 & 50.75 & 1.41
& 43.29 & 37.81 & 55.10 & 53.20 & 2.53 \\
Qwen2.5-VL-32B~\cite{bai2025qwen2}
& 57.30 & 15.87 & 38.82 & 61.33 & 6.58
& 53.06 & 55.04 & 57.47 & 43.70 & 0.00 \\
Qwen3-VL-A3B-30B~\cite{bai2025qwen3} [3B]
& 53.15 & 13.84 & 38.69 & 59.81 & 8.69
& 46.22 & 57.17 & 53.63 & 47.00 & 0.31 \\
Qwen3-VL-32B~\cite{bai2025qwen3}
& \textbf{59.93} & 15.70 & \underline{41.53} & \underline{61.64} & 6.00
& 51.70 & 55.29 & 46.77 & 54.30 & 0.00 \\
\midrule

\multicolumn{11}{l}{\textbf{Spatial Models}} \\
UniVG-R1-7B~\cite{bai2025univg}
& 46.82 & 53.85 & 1.96 & 46.51 & 4.61
& 48.75 & 51.42 & 39.13 & 39.20 & 1.20 \\
PR1-OCR-2B~\cite{yu2025perception}
& 38.88 & \underline{11.60} & 13.81 & 44.65 & 8.18
& 50.73 & 47.68 & 44.05 & 48.50 & 4.07 \\
PR1-Detection-3B~\cite{yu2025perception}
& 36.74 & 62.61 & 33.84 & 56.08 & 4.35
& 41.35 & 44.81 & 62.47 & 56.08 & 0.10 \\
PR1-Counting-2B~\cite{yu2025perception}
& 40.32 & 51.95 & 13.69 & 48.52 & 1.94
& 39.72 & 50.94 & 47.43 & 48.52 & 2.00 \\
PR1-Grounding-2B~\cite{yu2025perception}
& 39.54 & 12.34 & 13.96 & 43.67 & 7.15
& 52.03 & 53.83 & 51.00 & 50.30 & 4.85 \\
\midrule

\multicolumn{11}{l}{\textbf{Embodied Models}} \\
RoboTron-Drive~\cite{2024RoboTron}
& 49.73 & 12.68 & 8.73 & 47.45 & 1.12
& 44.50 & \underline{65.91} & 52.18 & 37.70 & 1.10 \\
Cosmos-R1~\cite{azzolini2025cosmos}
& 45.62 & 23.41 & 10.05 & 55.93 & 1.03
& 35.80 & 54.66 & 55.33 & 52.70 & 1.23 \\
Mimo-Embodied~\cite{hao2026mimo}
& 53.62 & 16.57 & 24.04 & 55.82 & 7.90
& \underline{52.95} & 40.31 & \underline{65.23}
& \textbf{68.60} & \underline{4.90} \\
\midrule

\multicolumn{11}{l}{\textbf{Our Model}} \\
\rowcolor{waymolgray}
XEmbodied (Full), 30B [3B]
& \underline{59.53} & \textbf{6.83} & \textbf{83.83}
& \textbf{68.61} & \textbf{27.76}
& \textbf{57.33} & \textbf{77.01} & \textbf{78.55}
& \underline{65.70} & \textbf{25.43} \\
\bottomrule
\end{tabular}%
}
\caption{Comparison on spatial and 3D understanding and semantic reasoning benchmarks. Model size denotes total parameters, with active MoE parameters in brackets. $\uparrow$ and $\downarrow$ indicate that higher and lower values are better, respectively. The best and second-best results are \textbf{bolded} and \underline{underlined}.}
\label{tab:main_results}
\end{table*}

\subsection{Training Data and Benchmarks}
\label{subsec:data_benchmark}

To build a large-scale, high-quality VQA training corpus, we aggregate multi-source VQA-style datasets from general, autonomous driving, and robotics domains.
We follow the official train/validation/test splits of each dataset, so the training set and all evaluation test sets are fully non-overlapping.
The primary training datasets are OmniDrive~\cite{wang2024omnidrive}, SURDS~\cite{guo2025surds}, DriveLMM-o1~\cite{ishaq2025drivelmmo1stepbystepreasoningdataset}, DrivingVQA~\cite{drivingvqa2025}, LingoQA~\cite{marcu2023lingoqa}, MapLMv2~\cite{10657418}, BDD~\cite{yu2020bdd100kdiversedrivingdataset}, RoboVQA~\cite{robovqa2023arxiv}, RoboRefIt~\cite{lu2023vl}, RefCOCO~\cite{kazemzadeh-etal-2014-referitgame}, VQAv2~\cite{balanced_vqa_v2}, GQA~\cite{hudson2019gqa}, Flickr~\cite{young2014image}, and several smaller-scale datasets.
The key benchmarks for out-of-distribution evaluation (Cosmos-R1, Ego3DBench, and DriveBench) are fully excluded from training.
Detailed dataset composition, data statistics, and benchmark configurations are provided in Data and Curation.


\begin{table*}[t]
\centering
\small 
\setlength{\tabcolsep}{1pt} 
\begin{tabular}{lccccccccccc}
\toprule
Model & \makecell{Size \\ (B)} & \makecell{EIEA \\ S1} & \makecell{EIEA \\ S2} &
\multicolumn{2}{c}{\textbf{DriveLMM-o1}} & \multicolumn{2}{c}{\textbf{DriveBench}} &
\multicolumn{3}{c}{\textbf{Ego3D-Bench}} \\
\cmidrule(lr){5-6} \cmidrule(lr){7-8} \cmidrule(lr){9-11}
& & & & Score & Time (s) & Score & Time (s) & Acc & RMSE & Time (s) \\
\midrule
XEmbodied (SFT) & 30 & \xmark & \xmark & 64.32 & 1.52 & 53.18 & 1.64 & 55.28 & 9.44 & 1.25 \\
AgentThink & 7 & \xmark & \xmark & 71.35 & 3.17 & 55.41 & 3.36 & $-$ & $-$ & $-$ \\
XEmbodied w/EIEA & 30 & \cellcolor{waymolgray}\cmark & \xmark & 75.28 & $-$ & 56.72 & $-$ & 55.45 & 6.85 & $-$ \\
\rowcolor{waymolgray} XEmbodied (Full) & 30 & \cmark & \cmark & \textbf{77.01} & \textbf{0.46} & \textbf{57.33} & \textbf{0.52} & \textbf{59.53} & \textbf{6.83} & \textbf{0.40} \\
\bottomrule
\end{tabular}
\caption{Performance and Inference Efficiency of EIEA. All efficiency metrics are tested under identical experimental setup with 32 H20 GPUs and batch size 1. Full OOD ablation results are provided in Extended Ablations.}
\label{tab:EIEA}
\end{table*}


\begin{table*}[t]
\centering
\setlength{\tabcolsep}{2pt}
\begin{tabular}{l c c c c c c c c}
\toprule
Group & \makecell{Data \\ Strategy } & MLP & QFormer & Attn & \makecell{Spatial \& 3D \\ Understand. } & \makecell{Semantic \& \\ Reasoning} & \makecell{Embodied \& \\ Affordance} & Avg \\
\midrule
(a) & \xmark & \xmark & \xmark & \xmark & 73.39 & 82.17 & 60.94 & 71.48 \\
(b) & \cellcolor{waymolgray}\cmark & \xmark & \xmark & \xmark & 75.45 & \textbf{89.85} & 63.70 & 75.68 \\
(c) & \xmark & \xmark & \cellcolor{waymolgray}\cmark & \xmark & 71.92 & 78.10 & 58.30 & 69.10 \\
(d) & \cellcolor{waymolgray}\cmark & \cellcolor{waymolgray}\cmark & \xmark & \xmark & 74.88 & 75.49 & 59.42 & 69.07 \\
(e) & \cellcolor{waymolgray}\cmark & \xmark & \cellcolor{waymolgray}\cmark & \xmark & \textbf{78.75} & 86.02 & 56.50 & 72.52 \\
\rowcolor{waymolgray} (f) & \cmark & \xmark & \xmark & \cmark & 77.02 & 89.48 & \textbf{70.01} & \textbf{78.45} \\
\bottomrule
\end{tabular}
\caption{Model architecture and data strategy ablation under Qwen3-VL-30B-A3B backbone. (a) Backbone baseline; (b) Data curation only; (c) Standalone 3DA without data curation; (d) MLP fusion variant; (e) QFormer fusion variant; (f) Our cross-attention 3DA + data curation full model.}
\label{tab:data_model_ablation}
\end{table*}

\subsection{Main Results}
\textbf{SOTA performance.}
We evaluate XEmbodied on 16 benchmarks spanning spatial and 3D understanding, semantic reasoning, and embodied affordance. Table~\ref{tab:main_results} summarizes the spatial and semantic results, where we report \emph{XEmbodied (SFT)} after Stage~3 with data curation and 3DA, and \emph{XEmbodied (Full)} with additional EIEA and GRPO post-training. Although based on Qwen3-VL-30B-A3B, its MoE architecture activates only $\sim$3B parameters per token, making its computation comparable to 7B-class dense models such as Mimo-Embodied and Cosmos-R1; we therefore denote it as ``30B [3B].'' A same-backbone OOD ablation in Extended Ablations further improves the average score from 71.48 to 80.18 ($+8.7$), confirming that the gains mainly come from our proposed modules and training strategy rather than model scale.

As shown in Table~\ref{tab:main_results}, XEmbodied achieves the best results on Ego3D-Bench RMSE, SURDS, VLADBench, and STRIDE-QA, reducing the previous best Ego3D RMSE from 15.70 to 6.83 ($56.5\%$). It also ranks first on DriveBench, DriveLMM-o1, and OmniDrive, while remaining competitive on MapLM-v2 and LingoQA. Beyond the main table, XEmbodied also shows strong embodied affordance and physical-interaction capability; notably, it reaches 81.50 on the OOD Cosmos-R1 benchmark, slightly surpassing Mimo-Embodied (81.00) without textual chain-of-thought training. Detailed affordance results and qualitative comparisons across autonomous-driving and robotic tasks are provided in Extended Ablations.


\textbf{Zero-shot generalization.}
Ego3DBench, Cosmos-R1, and DriveBench are strictly OOD datasets excluded from training. Without any task-specific fine-tuning, XEmbodied (Full) outperforms the previous best baselines on three of four OOD metrics and remains competitive on the fourth, demonstrating strong zero-shot generalization to unseen tasks and scenarios (Extended Ablations). Notably, it improves DriveBench by 4.27 points, reduces Ego3D RMSE by 56.5\%, and surpasses Mimo-Embodied on Cosmos-R1, while its Ego3D ACC is only 0.40 points below Qwen3-VL-32B. The latter trade-off indicates substantially better continuous depth estimation with comparable discrete accuracy. We use the \emph{Full} model for all OOD claims; the previously noted 0.12-point DriveBench gap corresponds to the \emph{SFT} variant rather than the final model.

\textbf{Deployment positioning and downstream transfer.}
XEmbodied is a \emph{cloud-side} foundation model for large-scale scenario mining, annotation, and embodied VQA, not an on-board real-time controller; throughput and annotation reliability therefore matter more than per-step latency. Under this positioning, EIEA compresses explicit tool-chain outputs into compact latent tokens, giving 0.46--0.52\,s/sample versus 3.17--3.36\,s/sample for the explicit CoT baseline (Tab.~\ref{tab:EIEA}). To check that benchmark gains carry over to downstream utility, we fine-tune XEmbodied as a VLA backbone on nuScenes planning tasks; as shown in Extended Ablations, XEmbodied-FT attains competitive L2 and collision-rate metrics against dedicated end-to-end planners, indicating clean transfer of the acquired spatial competence to downstream decision-making. Real-robot deployment is left as future work, with distillation to lightweight on-board models as a natural next step.

\subsection{Ablation Studies}
To validate our method, we run systematic ablations across four core dimensions: data filtering strategy, 3DA architecture design, latent physical cue token injection via EIEA, and staged curriculum learning. These ablations measure the independent and synergistic effects of each component on three task domains; all results are normalized for fair comparison, with the normalization protocol and training settings detailed in Training Details. All ablation rows share the same Qwen3-VL-30B-A3B backbone (a MoE model with $\sim$3B active parameters per token), so the observed deltas trace to the proposed modules and training strategy rather than backbone scale. Extended same-backbone OOD, curriculum-learning, model-size, and data-source ablations are provided in Extended Ablations.

\textbf{Impact of data filter strategy.}
Comparing groups (a) and (b) in Tab.~\ref{tab:data_model_ablation}, our four-stage data strategy lifts the weighted total average from 71.48 to 75.68. Sensitivity analyses of the spatial-entropy parameters and tier thresholds are provided in Data and Curation.


\textbf{Architecture of the 3D Adapter.}
We study the best fusion mechanism for injecting 3D geometric cues into the semantic stream. Fusion architecture strongly affects performance; results are weighted by dataset quantity (5 for 3D understanding, 6 for semantic reasoning, 7 for embodied affordance). A simple MLP for feature projection (group~d) drops performance to a weighted average of 69.07, below the 75.68 baseline of group~b. Static linear projection fails to align 3D geometric priors with 2D semantic tokens and can inject noise into the fusion.

QFormer (group~e) allows richer interaction between 3D and 2D features but only marginally raises semantic reasoning to 86.02 while dropping to 56.50 on embodied affordance. Its weighted average of 72.52 stays below the baseline, showing that advanced fusion alone lacks the dynamic alignment needed to use 3D information well. Our cross-attention mechanism (group~f) achieves the best overall performance, 78.45, with embodied affordance surging to 70.01, 13.51 points above QFormer. The advantage comes from cross-attention's dynamic addressing, which lets 2D semantic tokens selectively aggregate relevant 3D geometric information.

\textbf{3DA-only contribution.}
To isolate 3DA's individual effect, we evaluate two 3DA-only settings on the same backbone. \emph{In-distribution} (Tab.~\ref{tab:data_model_ablation}, group~(c)): adding 3DA \emph{without} data curation gives a weighted average of 69.10, comparable to the MLP/QFormer variants and slightly below the backbone baseline (71.48). \emph{Out-of-distribution} (we further probe 3DA-only on the strict OOD benchmarks): +3DA only scores DriveBench 44.97, Ego3D ACC 54.69, Ego3D RMSE 8.62, Cosmos-R1 73.00 (16-bmk Avg 71.92). The OOD probe is the more informative reading: 3DA alone already cuts Ego3D RMSE from 13.84 to 8.62, so it captures geometric structure even without curated data. The in-distribution dip on DriveBench (44.97 vs.\ 46.22) reveals a \emph{precondition} rather than a weakness: 3DA injects a continuous geometric prior, and when the answer space is not shaped by geometrically-aligned samples, that prior has no anchor and shows up as score jitter. The moment data curation supplies geometrically-aligned samples, the same 3DA lifts the in-distribution average from 75.68 (data-only, group~(b)) to 78.45 (group~(f)), and drives OOD Ego3D RMSE to 6.83 and DriveBench to 57.33 once EIEA is added (full ablation in Extended Ablations). Endogenous geometry therefore needs a geometrically-aligned answer space to convert structure into correct answers, and 3DA and curated data are \emph{synergistic} in that sense. The cross-attention 3DA in group~(f) corresponds to the Baseline+3DA configuration.

\textbf{Impact of latent physical cue tokens.}
Tab.~\ref{tab:EIEA} evaluates EIEA against several variants, including explicit tool-augmented reasoning. Although the 7B AgentThink baseline achieves 71.35 on DriveLMM-o1 and 55.41 on DriveBench, it requires 3.17 s per DriveLMM-o1 sample, 6.89$\times$ slower than our final model (0.46 s/sample), revealing the inefficiency of explicit physical-cue CoT. EIEA alignment improves XEmbodied-SFT from 64.32 to 75.28 on DriveLMM-o1 and 53.18 to 56.72 on DriveBench, while reducing Ego3D-Bench RMSE from 9.44 to 6.85 with 55.45 accuracy. Stage-2 GRPO further enables end-to-end joint optimization, achieving 77.01, 57.33, and 59.53/6.83 accuracy/RMSE, respectively.

EIEA yields particularly large OOD gains on Ego3D-Bench (RMSE 9.25$\rightarrow$6.83 for the same-backbone +Data+3DA ablation; the corresponding SFT-stage reduction from 9.44 to 6.85 is a different reference and is reported in Table~\ref{tab:EIEA}), DriveBench (53.18$\rightarrow$57.33), and Cosmos-R1 (76.00$\rightarrow$81.50), showing that distilled physical cues improve generalization. EIEA and 3DA are complementary: removing 3DA while retaining dual-EIEA lowers DriveBench to 51.36, below the vanilla 53.18 baseline (OOD ablation table in Extended Ablations), indicating that EIEA requires 3DA for correct evidence routing. Same-backbone results in Extended Ablations exclude scale effects, while the feature-routing analysis in Data and Curation shows that 3DA raises the routing margin from 0.354 to 0.544 and top-1 accuracy from 0.800 to 1.000.


\section{Implementation Details}
\label{sec:implementation}

In this section, we provide comprehensive implementation details of the 3D Adapter (3DA) and Efficient Image-Embodied Adapter (EIEA) proposed in the main paper. All notations follow the definitions in Sections~\ref{sec:3da} and~\ref{sec:eiea} of the main text.
\begin{figure*}[!t]
    \centering
    \includegraphics[width=\textwidth]{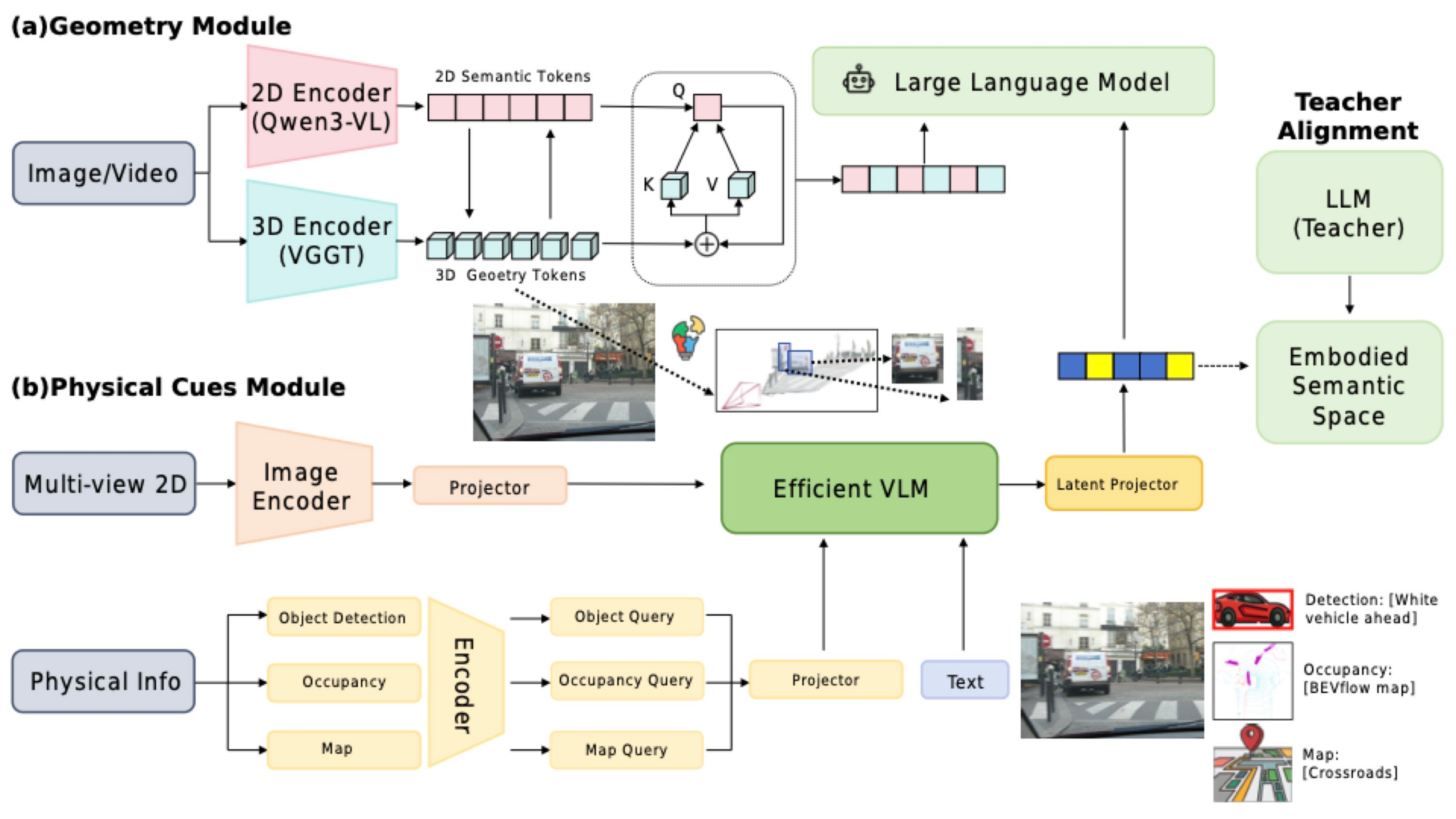} 
    \caption{Overall model architecture of the proposed 3D Adapter (3DA) and Efficient Image-Embodied Adapter (EIEA). The architecture integrates 3D geometric priors and heterogeneous embodied physical information into the VLM backbone for enhanced spatial and embodied reasoning.}
    \label{fig:supp_model_arch}
\end{figure*}
\subsection{3D Adapter (3DA) Implementation Details}
\label{sec:3da_impl}

The 3DA is built on top of Qwen3-VL-30B-A3B \cite{bai2025qwen3} and VGGT \cite{wang2025vggt}, with a core design of 3D feature fusion via MLP projection and cross-attention residual fusion. The full architecture is composed of three key components (Figure~\ref{fig:supp_model_arch}): a semantic stream (Qwen3-VL), a geometry stream (VGGT), and a token alignment/fusion module. VGGT inputs are letterboxed to a white $518\times518$ square, the Qwen stream preserves the original aspect ratio, and the connector uses bilinear interpolation. The residual is added to the existing image embeddings, so the model introduces no additional visual sequence tokens.

\subsubsection{Semantic Stream (Qwen3-VL Backbone)}
We retain the original vision pipeline of Qwen3-VL without modification: raw pixel values ($\text{pixel\_values} \in \mathbb{R}^{B \times 3 \times H \times W}$) and image grid dimensions ($\text{image\_grid\_thw} \in \mathbb{Z}^{N_{\text{grid}} \times 3}$) are processed to generate 2D semantic tokens $\mathbf{e}_{\text{2D}} \in \mathbb{R}^{N \times d_{\text{2D}}}$, where:
- $d_{\text{2D}} = 2048$ (hidden size of Qwen3-VL-30B-A3B),
- Qwen3-VL uses a patch size of 16, with temporal patch size $\tau=2$ and spatial merge size $\sigma=2$ for video/image token merging,
- $N = \lfloor H/16 \rfloor \times \lfloor W/16 \rfloor$ (number of 2D semantic tokens per image).

\subsubsection{Geometry Stream (VGGT 3D Encoder)}
The geometry stream leverages a pre-trained VGGT encoder to extract 3D spatial tokens, with the following critical configurations (aligned with our implementation in code):
\begin{itemize}
    \item VGGT encoder config: image size $518 \times 518$, patch size 14, embedding dimension $d_{\text{3D}} = 1024$,
    \item Input: Resized image tensor $\text{image\_tchw} \in \mathbb{R}^{T \times 3 \times 518 \times 518}$ (temporal dimension $T=2$ for image frames),
    \item Output: 3D geometry tokens $\mathbf{e}_{\text{3D}} \in \mathbb{R}^{M \times 2d_{\text{3D}}}$ (the factor of 2 comes from VGGT's dual-stream encoding), where $M = \lfloor 518/14 \rfloor \times \lfloor 518/14 \rfloor = 1369$,
    \item We use the last layer output of VGGT (layer index = -1) as the spatial embeddings (consistent with Spatial-MLLM \cite{ouyang2025spacer}).
\end{itemize}

\subsubsection{Token Alignment and Fusion}
To resolve the resolution mismatch between VGGT (patch size 14) and Qwen3-VL (patch size 16), we implement a two-stage alignment + fusion pipeline:

\paragraph{Step 1: Grid Interpolation}
We map VGGT's 3D token grid to Qwen3-VL's 2D token grid via bilinear interpolation (mode: bilinear, align\_corners=False):
\begin{equation}
\label{eq:impl_interp}
\tilde{\mathbf{e}}_{\text{3D}} = \text{Interpolate}\left(\mathbf{e}_{\text{3D}}, \text{size}=(\lfloor H/16 \rfloor, \lfloor W/16 \rfloor)\right) \in \mathbb{R}^{N \times 2d_{\text{3D}}},
\end{equation}
where the interpolation operation reshapes $\mathbf{e}_{\text{3D}}$ to $[T, d_{\text{3D}} \times 2, H_v, W_v]$ (with $H_v=W_v=37$), upsamples/downsamples to Qwen3's grid size $(\lfloor H/16 \rfloor, \lfloor W/16 \rfloor)$, then reshapes back to token sequence format.

\paragraph{Step 2: MLP Projection}
A two-layer MLP with GELU activation projects $\tilde{\mathbf{e}}_{\text{3D}}$ to the dimension of Qwen3-VL's 2D tokens:
\begin{equation}
\label{eq:impl_mlp_proj}
\hat{\mathbf{e}}_{\text{3D}} = \text{MLP}\left(\text{RMSNorm}(\tilde{\mathbf{e}}_{\text{3D}})\right) \in \mathbb{R}^{N \times d_{\text{2D}}},
\end{equation}
where the MLP has:
- Input dimension: $2d_{\text{3D}} \times \tau \times \sigma^2 = 2 \times 1024 \times 2 \times 4 = 16384$,
- Hidden layer: linear layer ($16384 \to 16384$) + GELU,
- Output layer: linear layer ($16384 \to d_{\text{2D}} = 2048$),
- RMSNorm is applied before MLP to stabilize training.

\paragraph{Step 3: Cross-Attention Residual Fusion}
Following Eq.~\eqref{eq:cross_attn_fusion} in the main text, we implement cross-attention with:
\begin{itemize}
    \item Queries ($\mathbf{Q}$): $\text{RMSNorm}(\mathbf{e}_{\text{2D}}) \in \mathbb{R}^{N \times d_{\text{2D}}}$,
    \item Keys/Values ($\mathbf{K/V}$): $\text{RMSNorm}(\hat{\mathbf{e}}_{\text{3D}}) \in \mathbb{R}^{N \times d_{\text{2D}}}$,
    \item Number of heads: $32$ (matching the 32 attention heads of Qwen3-VL-30B-A3B, each with head dimension $128$),
    \item Batch-first=True for efficient batch processing.
\end{itemize}
The residual fusion (Eq.~\eqref{eq:residual_fusion}) is implemented as:
$$\mathbf{e}_{\text{fusion}} = \mathbf{e}_{\text{2D}} + \text{CrossAttn}(\mathbf{Q}, \mathbf{K}, \mathbf{V}),$$
with all tensors cast to the same device/dtype (bfloat16 for training/inference) to avoid numerical mismatches.

\subsection{Efficient Image-Embodied Adapter (EIEA) Implementation Details}
\label{sec:eiea_impl}

EIEA integrates three core embodied physical modalities (BEV map, BEV flow, and 3D bounding boxes) into VLMs via modality-specific encoding, Mamba-based reasoning, and distillation-based compression. Below are the implementation details for each component.

\subsubsection{Modality-Specific Feature Extraction}
For each physical modality $m \in \{\text{BEV map}, \text{BEVflow}, 3\text{D box}\}$, we use off-the-shelf encoders $\mathcal{E}_m$ and lightweight projectors $\phi_m$:

\begin{table*}[htbp]
\centering
\small
\begin{tabular}{lcc}
\toprule
Modality $m$ & Encoder $\mathcal{E}_m$ & Output Dimension $d_m$ \\
\midrule
BEV Map (semantic segmentation) & DiffMap \cite{jia2024diffmap} & 768 \\
BEVflow (motion estimation) & PriorMotion \cite{qian2025priormotion} & 512 \\
3D Bounding Box (detection) & SG-Net \cite{wang2023sgfnet} & 256 \\
\bottomrule
\end{tabular}
\vspace{0.5em}
\caption{Modality-specific encoders for EIEA. All encoders are pre-trained on downstream embodied datasets.}
\label{tab:eiea_encoders}
\end{table*}

The projector $\phi_m$ is a two-layer MLP with:
\begin{itemize}
    \item Input: $d_m$ (per-modality dimension, Table~\ref{tab:eiea_encoders}),
    \item Hidden layer: linear layer ($d_m \to d$) + GELU,
    \item Output layer: linear layer ($d \to d$),
    \item Shared dimension $d = 1024$ (aligned with VGGT's embedding dimension for consistency).
\end{itemize}

For 3D bounding boxes, we first convert raw detection outputs (center coordinates, size, rotation, velocity) to a dense token representation:
$$\mathbf{H}_{\text{3Dbox}} = \text{Embedding}(\text{box\_features}) \in \mathbb{R}^{L_{\text{3D}} \times 256},$$
where $L_{\text{3D}} = 100$ (fixed number of box tokens per scene) and $\text{box\_features}$ include both geometric attributes and category embeddings.

\subsubsection{Mamba-Based Multimodal Interpreter}
The lightweight Mamba interpreter $g_{\text{Mamba}}$ is built upon the selective state space sequence model Mamba-2\cite{yu2025mambaout}, adhering to the efficient language model design paradigm proposed in SSR\cite{liu2025ssr}. We adopt the Mamba-2 130M variant with its native selective state space configuration for this interpreter, where the model dimension is set as $d_{\text{model}} = d = 1024$ with an expand factor of 2 and a single hidden layer enabling linear-time sequence modeling; to guarantee the plug-and-play efficiency of the adapter module, the main body of the Mamba-2 model is kept frozen during training, with only the input and output projection layers fine-tuned for modality alignment. The input sequence of the interpreter is constructed by concatenating text query tokens and projected physical modality tokens in the structured form of $\left[\mathbf{X}_T; \langle \text{PHY} \rangle; \mathbf{Z}_{\text{BEVmap}}; \langle \text{MOD} \rangle; \mathbf{Z}_{\text{BEVflow}}; \langle \text{MOD} \rangle; \mathbf{Z}_{\text{3Dbox}}\right] \in \mathbb{R}^{(L_T + L_{\text{token}} + L_{\text{BEVmap}} + L_{\text{BEVflow}} + L_{\text{3Dbox}}) \times d}$, where $L_{\text{token}}$ denotes the total length of the specially designed inserted tokens. For effective sequence modeling and subsequent knowledge distillation, we uniformly insert 10 learnable latent reasoning tokens into the sequence, and additionally introduce two types of symbolic special tokens for modality separation: the $\langle \text{PHY} \rangle$ token acts as a start marker for physical modality features to clearly distinguish text and physical modal information, while the $\langle \text{MOD} \rangle$ token is utilized to separate different heterogeneous embodied modalities including BEV map, BEVflow and 3D bounding box. After the joint encoding of text query and physical modality features via the Mamba-2 model, we only extract and retain the text-position token sequence that strictly matches the length of the original text query token sequence $\mathbf{X}_T$ as the interpreter output $\mathbf{H}_{\text{text}} \in \mathbb{R}^{L_T \times d}$, and this output is further fed into the downstream latent projection module for subsequent fusion with the VLM backbone.

The latent projector $\phi_R$ follows the rationale-to-semantic space mapping design in SSR\cite{liu2025ssr}: we first apply RMSNorm to stabilize the feature distribution of $\mathbf{H}_{\text{text}}$, and then map the normalized features to the VLM's word embedding space through a two-layer MLP with GELU activation, with the mapping process formulated as:
$$\mathbf{Z}_{\text{phy}} = \phi_R(\mathbf{H}_{\text{text}}) = \text{Linear}\left(\text{GELU}\left(\text{Linear}\left(\text{RMSNorm}(\mathbf{H}_{\text{text}})\right)\right)\right) \in \mathbb{R}^{L_T \times d_{\text{VLM}}},$$
where $d_{\text{VLM}} = 2048$ to match the word embedding dimension of the Qwen3-VL backbone, and the RMSNorm adopts the Qwen2-style configuration with $\epsilon=1e-6$ to ensure consistency with the 3DA module and the VLM backbone.

\subsubsection{Distillation-Based Compression}
Directly feeding long physical modality tokens into the VLM context window leads to inefficient inference and even degrades the model's reasoning performance, thus we employ a knowledge distillation strategy to compress the physical modality features into compact physical cue tokens while preserving critical spatial and physical reasoning information. In this distillation framework, the teacher model is set as the full Mamba interpreter output $\mathbf{H}_{\text{text}} \in \mathbb{R}^{L_T \times d}$, which contains complete intermediate reasoning information of physical modalities, while the student model is a lightweight compact token generator composed of a 1-layer MLP and RMSNorm, whose output is the compact physical cue token sequence $\mathbf{Z}_{\text{phy}}^{\text{student}} \in \mathbb{R}^{L_{\text{compact}} \times d_{\text{VLM}}}$. The distillation loss is a hybrid loss function consisting of the MSE loss between the teacher and student outputs (to preserve feature-level reasoning information) and the cross-entropy loss on downstream reasoning tasks (to align the compressed tokens with the VLM's generation objective). We set the length of the compact physical cue tokens as $L_{\text{compact}} = 64$, which achieves an 8$\times$ reduction in token count compared with the raw physical modality tokens and significantly lightens the inference burden of the VLM.

The final input of the VLM (Eq.~\eqref{eq:final_gen}) is constructed by concatenating visual tokens $X_v$, text query tokens $X_T$ and the compact physical cue tokens $\mathbf{Z}_{\text{phy}}$ with dedicated special separator tokens for clear modal distinction, with the generation process formulated as:
$$Y = f_{\text{VLM}}\left(\left[\text{<VIS>}; X_v; \text{<TEXT>}; X_T; \text{<PHY>}; \mathbf{Z}_{\text{phy}}\right]\right).$$



\section{Data and Curation}
\label{sec:data_curation}

\subsection{Data Composition}
\label{sec:data_composition}

\subsubsection{AutoDrive Datasets}
\paragraph{\textbf{LingoQA}} A benchmark dataset designed for evaluating vision-language models on language-guided visual question answering. It focuses on the model's ability to follow complex linguistic instructions to perceive, reason about, and answer questions based on visual scenes.
\paragraph{\textbf{Maplmv2}} A multimodal dataset and benchmark centered on Map-based Large Multimodal Models. It typically involves tasks that require understanding and reasoning over visual map data in conjunction with natural language queries or instructions.
\paragraph{\textbf{BDD100k}} A large-scale, diverse driving video dataset for autonomous vehicle research. It contains extensive annotations for tasks like object detection, tracking, lane detection, and drivable area segmentation, captured under various weather and lighting conditions. This dataset is converted into VQA format via rules.
\paragraph{\textbf{SURDS}} A large-scale benchmark for spatial understanding and reasoning in driving scenarios with vision-language models. It probes fine-grained spatial relations, distance and position estimation, and 3D scene comprehension in complex road environments.
\paragraph{\textbf{DriveLMM-o1}} A dataset developed for training and evaluating Driving Language Models . It aligns driving scenarios (e.g., from simulators or logged data) with dense, hierarchical language descriptions and reasoning chains to model the decision-making process in autonomous driving.
\paragraph{\textbf{VLADBench}} A fine-grained benchmark for evaluating vision-language models in autonomous driving. It covers five domains---traffic knowledge, element recognition, traffic-graph generation, target attributes, and ego decision-making and planning---spanning 11 secondary aspects and 29 tasks with about 12k closed-form question-answer pairs.
\paragraph{\textbf{OmniDrive}} A holistic vision-language dataset for autonomous driving that aligns agent models with 3D driving tasks through counterfactual reasoning . It generates vision-language data for 3D autonomous driving via counterfactual reasoning, aligning agent models with planning and Q\&A tasks.
\paragraph{\textbf{DriveBench}} A reliability benchmark for VLMs in driving, evaluating performance across corrupted inputs, clean data, and text-only scenarios to assess robust visual grounding.

\subsubsection{Robotic Datasets} 
\paragraph{\textbf{RoboVQA}} A dataset for Robotic Visual Question Answering. It extends the VQA paradigm to robotics settings, where questions are grounded in a robot's egocentric visual perception of its environment, often involving temporal reasoning over video sequences.
\paragraph{\textbf{RoboRefIt}} A large-scale referring expression understanding dataset for robot perception, collected in real-world indoor scenes. It contains 10{,}872 RGB-D images with 50{,}758 referring expressions over 66 object categories and 187 scenes, requiring the model to ground a natural language expression to the referred object.
\paragraph{\textbf{STRIDE-QA}} A large-scale VQA dataset for spatiotemporal reasoning in urban scenes, featuring automatically generated annotations for object-centric and ego-centric tasks.
\paragraph{\textbf{Ego3D-Bench}} A dataset that evaluates the 3D spatial reasoning of VLMs using egocentric outdoor data, highlighting the gap between model performance and human understanding.
\paragraph{\textbf{Other robotic datasets}} This category encompasses other specialized, small-scale robotic datasets that complement the core benchmarks.

\subsubsection{Common Datasets}
\paragraph{\textbf{RefCOCO}} A classic benchmark for Referring Expression Comprehension. Given an image and a natural language expression referring to a specific object (e.g., "the woman in the red shirt"), the task is to localize the target object, typically by generating a bounding box or a segmentation mask.
\paragraph{\textbf{GQA}} A large-scale dataset for real-world visual reasoning and compositional question answering. Its questions are created based on the structured semantics of scene graphs, promoting testable reasoning (e.g., logical, spatial, comparative) about images.
\paragraph{\textbf{VQAv2}} A foundational and widely adopted dataset for the Visual Question Answering (VQA) task. It pairs images with open-ended questions, with answers provided by human annotators. A key feature is the inclusion of complementary image-question pairs to reduce language bias.
\paragraph{\textbf{Flickr30k}} A popular dataset for cross-modal retrieval and image captioning. It contains 30,000 images collected from Flickr, each annotated with 5 descriptive captions. It is commonly used to evaluate tasks like image-text matching and text-based image retrieval.

\subsection{Data Statistics}
\label{sec:data_statistics}

\begin{figure}[tb]
  \centering
  \includegraphics[height=7cm]{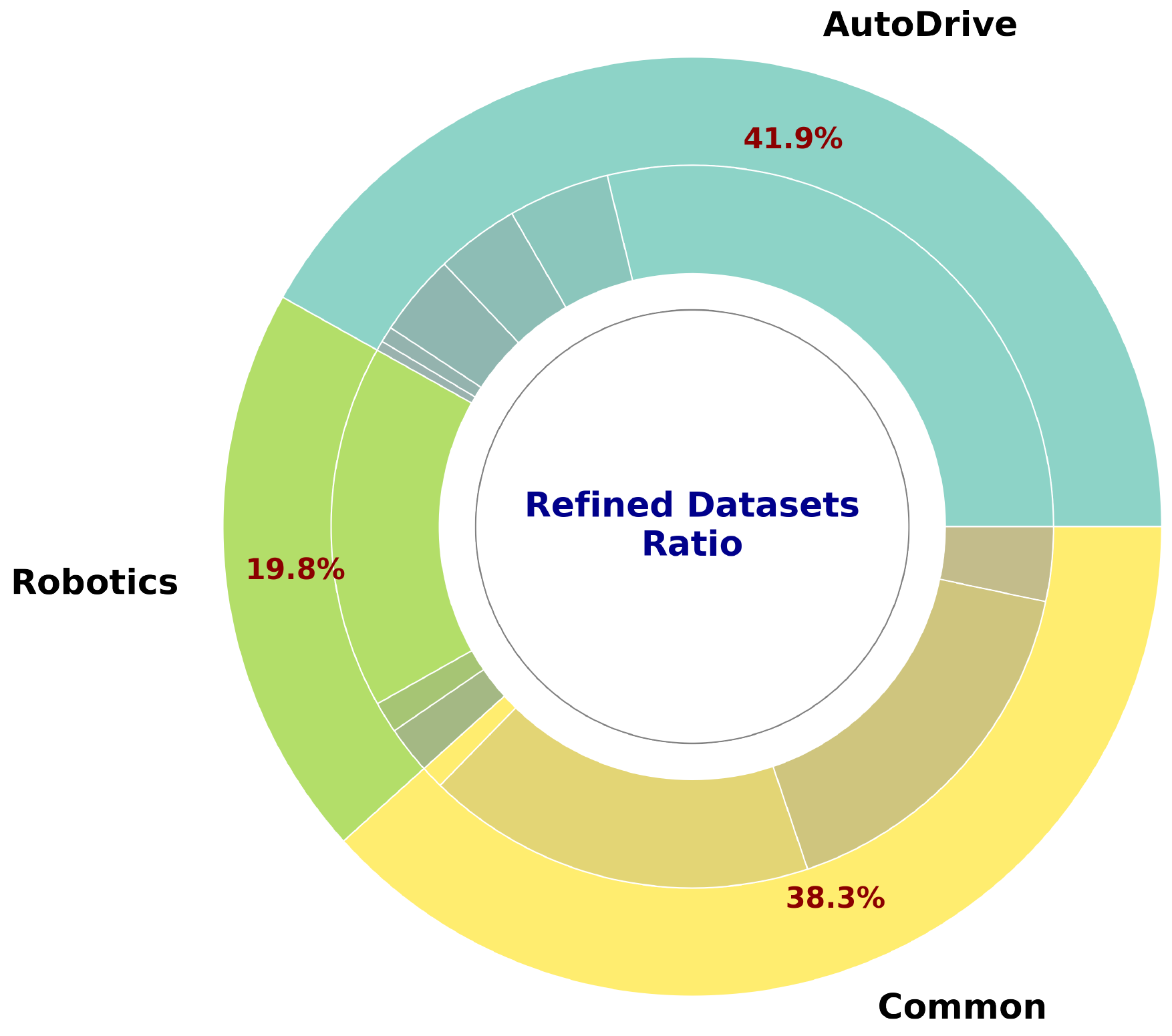}
  \caption{Distribution of refined datasets across three primary domains. The pie chart illustrates the proportional allocation of refined data, with the AutoDrive domain constituting the largest share at 41.9\%, followed by Common data at 38.3\%, and the Robotic domain comprising 19.8\%.
  }
  \label{fig:data_compose}
\end{figure}

\noindent\textbf{Automatic Data Refining} 
The raw pool contains approximately 2M samples; after automatic refining process (Sec.~\ref{sec:data_curation}) we retain $\sim$1M high-quality samples. Note that the retained $\sim$1M corpus and the throughput of the curation agent (over 10 million samples processed per day) are two different quantities: the former is the cumulated high-quality set actually used for training, whereas the latter is the daily processing rate of the cloud pipeline. 
Curation jointly considers (i) spatial richness $\mathcal{H}(x)$ (threshold $\tau=0.2$), (ii) a prompt-conditional coherence assessment, which filters out ambiguous or semantically inconsistent samples, and (iii) four-dimensional data quality assessment via an adaptive agent to eliminate low-quality samples. The resulting distribution is shown in Fig.~\ref{fig:data_compose}.

\noindent\textbf{Optimization of data distribution} 
This optimization in distribution is entirely algorithm-driven, ensuring strict alignment between data quality and the goal of cloud-based scenario understanding. (As detailed in Tab. \ref{tab:train_distribution}): The data distribution shifts markedly from the original pool to better serve our closed-loop objective of improving cloud-side understanding for scenario mining and annotation.

\begin{table}[t]
\centering
\small
\caption{Origin \& Refined Train Data Distribution}
\label{tab:train_distribution}
\begin{tabular}{l r @{\hspace{3em}} l r}
\toprule
\multicolumn{2}{l}{\textbf{Origin Distribution}} & \multicolumn{2}{l}{\textbf{Refined Distribution}} \\
\cmidrule(r){1-2} \cmidrule(l){3-4}
\textbf{DataSet} & \textbf{Ratio} & \textbf{DataSet} & \textbf{Ratio} \\
\midrule
GQA & 34.34\% & LingoQA & 28.28\% \\
VQAv2 & 16.16\% & GQA & 17.16\% \\
LingoQA & 15.07\% & VQAv2 & 16.21\% \\
BDD & 12.38\% & RoboVQA & 15.99\% \\
RoboVQA & 7.96\% & MapLMv2 & 4.46\% \\
OmniDrive & 4.09\% & BDD & 3.66\% \\
MapLMv2 & 3.65\% & SURDS & 3.61\% \\
Flickr & 1.93\% & Flickr & 3.27\% \\
SURDS & 1.80\% & OmniDrive & 1.79\% \\
RoboRefIt & 1.09\% & RoboRefIt & 1.38\% \\
DriveLMM-o1 & 0.67\% & RefCOCO & 1.01\% \\
RefCOCO & 0.51\% & DriveLMM-o1 & 0.70\% \\
VLADBench & 0.21\% & VLADBench & 0.43\% \\
Others & 0.14\% & Others & 2.05\% \\
\midrule
\textbf{Total} & \textbf{100\%} & \textbf{Total} & \textbf{100\%} \\
\bottomrule
\end{tabular}
\end{table}

\noindent\textbf{Benchmark}
We evaluate on a suite of 16 driving/robotics benchmarks that cover complementary capabilities:
Part-Affordance-2K, RoboAfford, SURDS, CosmosR1, DriveBench, DriveLMM, Ego3D-Bench, embodied-r1, LingoQA, MapLMv2, OmniDrive, RoboRefit, STRIDE-QA, VABench, VLADBench, and Where2Place.
These span spatial and 3D understanding, long-horizon/video-centric reasoning, semantic reasoning, and embodied affordance.
Across the benchmark suite, we evaluate 126{,}398 QA pairs over 34{,}028 scenes/frames (see Tab.~\ref{tab:bench_stats}).

\begin{table}[t]
\centering
\small
\caption{Benchmark suite statistics.}
\label{tab:bench_stats}
\begin{tabular}{lrr}
\toprule
\textbf{Dataset} & \textbf{\#QA} & \textbf{\#Scenes/Frames} \\
\midrule
Part-Affordance-2K & 2000 & 2000 \\
RoboAfford & 338 & 100 \\
SURDS & 11100 & 1709 \\
CosmosR1 & 210 & 201 \\
DriveBench & 24772 & 3200 \\
DriveLMM & 4634 & 539 \\
Ego3D-Bench & 8675 & 262 \\
embodied-r1 & 300 & 300 \\
LingoQA & 1000 & 100 \\
MapLMv2 & 13987 & 1500 \\
OmniDrive & 24070 & 6019 \\
RoboRefit & 2000 & 2000 \\
STRIDE-QA & 10634 & 10634 \\
VABench & 600 & 600 \\
VLADBench & 21978 & 4764 \\
Where2Place & 100 & 100 \\
\midrule
\textbf{Total} & \textbf{126398} & \textbf{34028} \\
\bottomrule
\end{tabular}
\end{table}

\subsection{Detailed Implementation of Data Curation Pipeline}
\label{subsec:data_curation_pipeline}

This section provides comprehensive details of the cloud-based data curation pipeline introduced in the main text, including formal definitions of key notations, full calculation of spatial entropy, and ablation experiments for critical parameters.

\subsubsection{Notation Definitions}
To ensure reproducibility and clarity, we define all key notations for the data curation pipeline in Table \ref{tab:data_notation}, consistent with the spatial entropy and tiered taxonomy described in the main text.

\begin{table*}[!htbp]
\centering
\caption{Key Notations for Data Curation Pipeline}
\label{tab:data_notation}
\begin{tabular}{ll}
\toprule
Notation & Definition \\
\midrule
$\mathcal{H}(x_i)$ & Spatial entropy score of sample $x_i$ (normalized to $[0,1]$) \\
$\mathcal{H}_{depth}(x_i)$ & Depth map variance entropy (weight coefficient $\alpha=0.6$) \\
$\mathcal{H}_{3D}(x_i)$ & 3D object distribution entropy (weight coefficient $1-\alpha=0.4$) \\
$D \in \mathbb{R}^{H \times W}$ & Depth map of sample $x_i$ (height $H$, width $W$) \\
$\sigma_k^2$ & Depth variance of the $k$-th $8 \times 8$ block in depth map $D$ \\
$\bar{\sigma}^2$ & Global mean of depth variances across all blocks \\
$\sigma_{min}^2/\sigma_{max}^2$ & Min/max depth variance of the full dataset ($0.01/25 \text{ m}^2$) \\
$G$ & Number of 3D grids for object distribution ($8 \times 8 \times 4$) \\
$p_t$ & Proportion of 3D objects in the $t$-th grid \\
$\theta_1/\theta_2/\theta_3$ & Spatial entropy thresholds for tier division ($0.3/0.5/0.7$) \\
$\tau$ & Minimum spatial entropy threshold for quality filtering ($0.2$) \\
$\phi$ & Minimum quality score threshold ($0.85$) \\
$\mathcal{C}(x_i)$ & Tier classification result of sample $x_i$ \\
\bottomrule
\end{tabular}
\end{table*}

\subsubsection{Data Classification for Curriculum Learning}
As outlined in the main text, the data classification module directly guides our progressive training strategy, ensuring the model builds spatial reasoning capabilities incrementally without manual annotation. The classification is tailored to the VQA-centric design of our embodied generalist model, focusing on vertical-domain autonomous driving data that typically suffers from ambiguity and low spatial information density. 

To address this challenge, our cloud-based automated curation agent classifies data by spatial complexity and domain relevance, as detailed in the main text. The automation ensures scalability—processing over 10 million samples daily—and eliminates human bias in data selection, a critical advantage for large-scale embodied model training.

\subsubsection{Full Calculation of Spatial Entropy Score}
The spatial entropy score $\mathcal{H}(x_i)$ quantifies scene geometric complexity using depth map variance and 3D object distribution statistics (low $\mathcal{H}$ for sparse scenes like straight highways, high $\mathcal{H}$ for complex scenes like multi-level intersections). Below is the complete, reproducible calculation framework, expanded from the main text:

\subsubsection{Total Spatial Entropy}
The total spatial entropy is a weighted fusion of depth variance entropy (prioritized for autonomous driving spatial perception) and 3D object distribution entropy:
$$
\mathcal{H}(x_i) = \alpha \cdot \mathcal{H}_{depth}(x_i) + (1-\alpha) \cdot \mathcal{H}_{3D}(x_i)
$$
where $\alpha=0.6$ balances the contribution of geometric texture (depth) and object dispersion (3D distribution) to scene complexity.

\subsubsection{Depth Map Variance Entropy $\mathcal{H}_{depth}(x_i)$}
This metric captures the geometric texture complexity of the scene by analyzing depth map variance at a local scale:
\begin{enumerate}
    \item Split the depth map $D$ into non-overlapping $8 \times 8$ blocks ($D_k, k=1,2,...,N$), where $N = \lfloor H/8 \rfloor \times \lfloor W/8 \rfloor$ (block size chosen to balance computational efficiency and local feature preservation);
    \item Compute the variance of each block:
    $$
    \sigma_k^2 = \frac{1}{64} \sum_{(u,v) \in D_k} (D(u,v) - \mu_k)^2
    $$
    where $\mu_k = \frac{1}{64} \sum_{(u,v) \in D_k} D(u,v)$ is the mean depth of block $D_k$;
    \item Calculate the global mean of block variances across the entire depth map:
    $$
    \bar{\sigma}^2 = \frac{1}{N} \sum_{k=1}^N \sigma_k^2
    $$
    \item Normalize to $[0,1]$ to eliminate depth scale differences across near/far-field scenes:
    $$
    \mathcal{H}_{depth}(x_i) = \frac{\bar{\sigma}^2 - \sigma_{min}^2}{\sigma_{max}^2 - \sigma_{min}^2}
    $$
    We set $\sigma_{min}^2=0.01 \text{ m}^2$ (sparse scenes with minimal depth variation) and $\sigma_{max}^2=25 \text{ m}^2$ (complex scenes with extreme depth variation), derived from dataset-level statistics of over 100M autonomous driving samples. For edge cases:
    \begin{itemize}
        \item If $\bar{\sigma}^2 < \sigma_{min}^2$: $\mathcal{H}_{depth}(x_i) = 0$;
        \item If $\bar{\sigma}^2 > \sigma_{max}^2$: $\mathcal{H}_{depth}(x_i) = 1$.
    \end{itemize}
\end{enumerate}

\subsubsection{3D Object Distribution Entropy $\mathcal{H}_{3D}(x_i)$}
This metric measures the spatial dispersion of 3D objects using Shannon entropy, reflecting the cognitive demand of scene understanding:
\begin{enumerate}
    \item Extract valid 3D bounding box centers $(x_j, y_j, z_j)$ for all non-occluded, non-background objects ($j=1,2,...,M$);
    \item Divide the 3D perception space (for autonomous driving) into $G$ discrete spatial units, and count the number of objects $n_t$ in each unit;
    \item Compute the proportion of objects in each unit: $p_t = n_t / M$ (set $p_t=0$ if $M=0$, i.e., empty scenes);
    \item Calculate raw Shannon entropy and normalize to $[0,1]$ (max entropy = $\log_2(G)$ for uniform object distribution):
    $$
    \mathcal{H}_{3D}(raw) = -\sum_{t=1}^G p_t \cdot \log_2(p_t) \quad (\text{convention: } 0 \cdot \log_2(0) = 0)
    $$
    $$
    \mathcal{H}_{3D}(x_i) = \frac{\mathcal{H}_{3D}(raw)}{\log_2(G)}
    $$
\end{enumerate}

\subsection{Tiered Data Taxonomy}
Consistent with the main text, we categorize samples into four tiers with incrementally increasing cognitive demand using $\mathcal{H}(x_i)$ and semantic tags (e.g., "multi-view", "temporal prediction"):
\begin{enumerate}
    \item \textbf{T1: Scene grounding \& commonsense} ($\mathcal{H} \leq 0.3$): Focus on basic object existence and coarse spatial relations, accounting for ~40\% of the curated dataset;
    \item \textbf{T2: Spatial localization} ($0.3 < \mathcal{H} \leq 0.5$): Focus on relative object positions and contextual understanding, accounting for ~30\% of the dataset;
    \item \textbf{T3: Multi-view/temporal spatial reasoning} ($0.5 < \mathcal{H} \leq 0.7$): Focus on cross-view or short-time-window reasoning, accounting for ~20\% of the dataset;
    \item \textbf{T4: Spatio-temporal understanding} ($\mathcal{H} > 0.7$): Focus on temporal prediction and causal spatial interactions, accounting for ~10\% of the dataset.
\end{enumerate}
The tier distribution aligns with the natural cognitive progression of human spatial reasoning and ensures the model is not overfitted to high-complexity samples in early training stages. Visual examples of each tier are provided as follows: Fig.~\ref{fig:data_tiered_bdd}, Fig.~\ref{fig:data_tiered_gqavqa} correspond to T1, Fig.~\ref{fig:data_tiered_surds}, Fig.~\ref{fig:data_tiered_roborefit}, Fig.~\ref{fig:data_tiered_drivingvqa}, Fig.~\ref{fig:data_tiered_refcoco} to T2, 
Fig.~\ref{fig:data_tiered_drivelmm}, 
Fig.~\ref{fig:data_tiered_maplmv2} to T3, and Fig.~\ref{fig:data_tiered_lingoqa}, 
Fig.~\ref{fig:data_tiered_robovqa} to T4.

\subsection{Automated Curation Pipeline}
The two-stage filtering process (structural screening + quality assessment) detailed in the main text is implemented as follows:

\subsubsection{Structural Screening}
Samples are classified into tiers using spatial entropy thresholds:
\begin{equation}
\mathcal{C}(x_i) =
\begin{cases}
\text{T1} & \text{if } \mathcal{H}(x_i) \leq \theta_1 \\
\text{T2} & \text{if } \theta_1 < \mathcal{H}(x_i) \leq \theta_2 \\
\text{T3} & \text{if } \theta_2 < \mathcal{H}(x_i) \leq \theta_3 \\
\text{T4} & \text{if } \mathcal{H}(x_i) > \theta_3
\end{cases}
\end{equation}
where $\theta_1=0.3$, $\theta_2=0.5$, $\theta_3=0.7$. These thresholds are chosen to balance the number of samples per tier and align with the model's progressive learning capacity.

\subsection{Sensitivity Analysis of Spatial Entropy Hyperparameters}
\label{sec:entropy_sensitivity}
To verify that our tiered data partitioning is robust to the choice of spatial-entropy hyperparameters, we sweep the weight coefficient $\alpha$ and the tier thresholds $(\theta_1,\theta_2,\theta_3)$ on a representative subset, and report Stage-1 SFT performance across the three task families. As shown in Table~\ref{tab:entropy_sweep}, the weighted average varies only within a narrow range (67.14--69.66, $\leq 2.52$ points), indicating that the tiered taxonomy is insensitive to small perturbations of $\alpha$ and $\theta$. Our adopted setting $(\alpha{=}0.6, \theta{=}(0.3,0.5,0.7))$ achieves the best overall average and the most balanced result across the three task families. We do not claim global optimality for this setting; rather, it provides a stable operating point that aligns the tiered curriculum with the model's progressive learning capacity.
\begin{table}[!htbp]
\centering
\small
\caption{Sensitivity analysis of spatial entropy hyperparameters (Stage-1 SFT only; not directly comparable to the main results in Tables~\ref{tab:main_results} and~\ref{tab:extended_results_ea}).}
\label{tab:entropy_sweep}
\begin{tabular}{lcccc}
\toprule
($\alpha$;\,$\theta_1,\theta_2,\theta_3$) & 3D & Sem. & Aff. & Avg \\
\midrule
(0.5; 0.5, 0.6, 0.8) & 64.20 & 71.80 & 65.50 & 67.14 \\
(0.5; 0.45, 0.6, 0.75) anchor & 66.10 & 72.80 & 66.80 & 68.50 \\
\textbf{(0.6; 0.3, 0.5, 0.7) ours} & 65.99 & \textbf{75.21} & \textbf{67.79} & \textbf{69.66} \\
\bottomrule
\end{tabular}
\end{table}

\subsection{Complementarity of 3DA and EIEA}
\label{sec:3da_eiea_complement}
\begin{figure}[!htbp]
  \centering
  \includegraphics[width=0.92\linewidth]{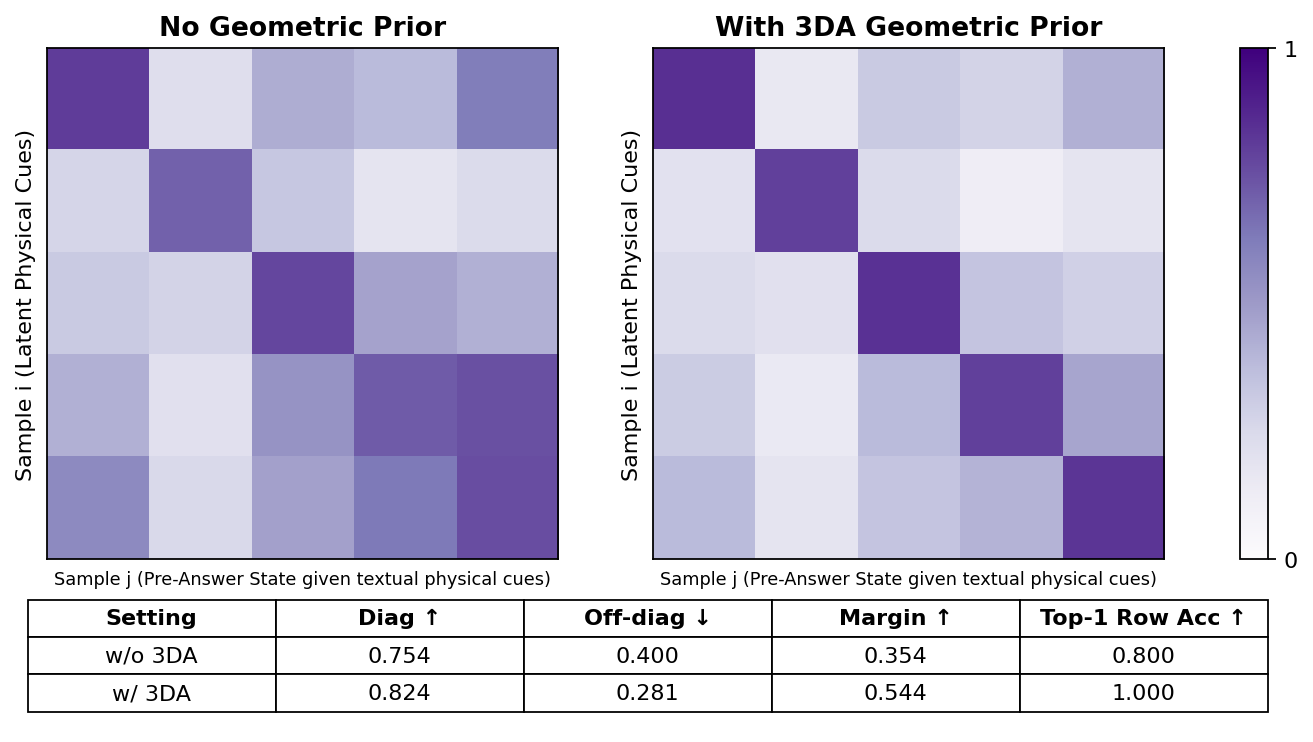}
  \caption{Cosine similarity of EIEA physical cue tokens versus LLM answer states. The 3DA+EIEA configuration sharpens diagonal alignment and suppresses off-diagonal crosstalk.}
  \label{fig:supp_cosine}
\end{figure}

A natural concern is whether 3DA and EIEA are functionally redundant, since both inject spatial/physical information into the VLM. We clarify their distinct roles: \textbf{3DA} provides \emph{endogenous} dense and continuous 3D geometric grounding (lane topology, metric distance, occlusion) from raw RGB/video via VGGT, building an internal spatial cognitive map; \textbf{EIEA} provides \emph{exogenous} sparse and discrete physical evidence (detection, occupancy, map) distilled from off-the-shelf perception tools into compact tokens for verifiable reasoning. They address complementary problems: 3DA provides intrinsic geometric structure, while EIEA supplies externally verifiable physical evidence that 3DA alone cannot capture. Empirically, we measure the cosine similarity between EIEA physical cue tokens and the LLM's answer-state embeddings across configurations. As shown in Fig.~\ref{fig:supp_cosine}, the left panel (EIEA only) exhibits diagonal dominance, indicating the model actively consumes EIEA tokens rather than ignoring them. The right panel (3DA+EIEA) sharpens the diagonal and suppresses off-diagonal crosstalk, suggesting that 3DA's geometric anchoring routes physical cues to their correct context and reduces cross-sample feature aliasing. Quantitatively, the diagonal/off-diagonal margin grows from 0.354 (EIEA only) to 0.544 (3DA+EIEA) and top-1 routing accuracy rises from 0.800 to 1.000, supporting the claim that 3DA is a prerequisite enabler for EIEA rather than a redundant duplicate.

\section{Training Details}
\label{sec:training_details}
\noindent\textbf{Ms-swift.}
This framework is built to enable large-scale training by leveraging tensor parallelism and activation checkpointing. All training is performed on 128 NVIDIA H20 GPUs unless noted. It follows the progressive pipeline (Sec.~\ref{subsec:training}), which involves training stage-specific modules. The training of the model was implemented in four distinct stages. A comparative summary of each stage is provided in Table \ref{tab:train_stages_diff}.

\noindent\textbf{Supervised Fine-tuning.}
In Stage 1-3, We implement model training using the SWiFT framework. The model is initialized from the pre-trained Qwen3-VL-30B-A3B-Instruct checkpoint and fine-tuned on the refined dataset.



\noindent\textbf{RL Post-training.}
Stage 4 performs GRPO (Sec.~\ref{subsec:training}) with 4 sampled responses per prompt and a KL penalty against the Stage-3 checkpoint, using a LoRA learning rate of 1e-5.
\textit{Stage-4 data (R1ehV-W4):} Stage~4 reuses the curated corpus from Stages~1--3 (no new data source is introduced); only the training objective changes---we add EIEA alignment and GRPO-based RL fine-tuning on top of the Stage-3 checkpoint, so any performance delta is attributable to the EIEA and RL rather than to a different data distribution.
We adopt a unified \emph{Foundation-ORM} reward mechanism that integrates \textbf{format constraint} and \textbf{answer correctness} via task-specific verifiers. The format reward scores 1.0 for valid $\langle$answer$\rangle$ tags and 0.0 otherwise. The correctness reward uses task-adaptive metrics: exact match for selection/matching tasks, distance-based score for point positioning, IoU for box regression, and semantic similarity for open-ended description. The final reward is $r_{\text{total}} = 0.2 \cdot r_{\text{format}} + 0.8 \cdot r_{\text{correct}}$ (clamped to $[0, 1]$), regularized by a KL divergence term (coefficient 0.05) to avoid distribution drift from the reference policy.
The weight choice $(\lambda_{\text{format}},\lambda_{\text{ans}}){=}(0.2,0.8)$ reflects the design goal of our cloud-side annotation engine: \textit{answer correctness} is the ultimate objective, while \textit{format} acts as a soft regularizer that keeps outputs parseable without dominating optimization. As the Stage-3 checkpoint already exhibits strong instruction-following, a small format weight suffices; larger format weights (e.g., 0.3) over-constrain language generation and slightly degrade final accuracy. We verify this empirically in Table~\ref{tab:grpo_sweep}: setting $(0.2,0.8)$ yields 78.45 on the 16-benchmark average, outperforming $(0.3,0.7)$ at 77.58, with consistent trends across all three task families. We therefore adopt $(0.2,0.8)$ throughout.

\begin{table}[t]
\centering
\small
\caption{GRPO reward weight sensitivity. $(0.2,0.8)$ is adopted in the final model.}
\label{tab:grpo_sweep}
\begin{tabular}{cccccc}
\toprule
$\lambda_{\text{format}}$ & $\lambda_{\text{ans}}$ & 3D & Sem. & Aff. & 16-bmk Avg \\
\midrule
0.2 & 0.8 (ours) & \textbf{77.02} & \textbf{89.48} & \textbf{70.01} & \textbf{78.45} \\
0.3 & 0.7       & 76.10 & 88.92 & 68.93 & 77.58 \\
\bottomrule
\end{tabular}
\end{table}

\begin{table}[t]
\centering
\scriptsize
\caption{Comparison of train settings in different stages.}
\label{tab:train_stages_diff}
\setlength{\tabcolsep}{1pt}
\renewcommand{\arraystretch}{1.08}
\begin{tabular}{lcccc}
\toprule
\textbf{Stages} & \textbf{Stage 1} & \textbf{Stage 2} & \textbf{Stage 3} & \textbf{Stage 4} \\
\midrule
\textbf{Dataset} & \makecell{Scene grounding \\Commonsense} & \makecell{Domain-related\\ Spatial Localization} & \makecell{Multi-view \\Spatial-Temporal} & \makecell{Curated \\ Corpus (reuse)} \\
\textbf{Batch Size} & 64  & 64 & 64 & 16 \\
\textbf{Epoch} & 1  & 1 & 1 & 2 \\
\textbf{Learning Rate} & $5\times 10^{-6}$ & $2\times 10^{-5}$ & $1\times 10^{-4}$ & $1\times 10^{-5}$ \\
\textbf{Optimizer} & AdamW & AdamW & AdamW & AdamW \\
\textbf{Weight Decay} & 0.05 & 0.05 & 0.05 & 0.05 \\
\textbf{LR Schedule} & Cosine & Cosine & Cosine & Cosine \\
\textbf{Max Sequence} & 16384 & 16384 & 16384 & 16384 \\
\textbf{Trainable Part} & All (ViT Frozen) & All (ViT Frozen) & LoRA ($r=64$) & LoRA ($r=64$) \\
\bottomrule
\end{tabular}
\end{table}

\noindent\textbf{Normalization of Evaluation.}
Due to the different ranges of evaluation metrics across benchmarks, direct comparison of model performance is infeasible. To enable fair and intuitive comparison, we normalize the evaluation results. For each benchmark, let $Y = [y_1, y_2, ..., y_n]$ denote the vector of all models' evaluation results (where $y_i$ is the $i$-th model's score) and $\max(Y)$ its maximum value. The normalized score $\hat{y}_i$ is calculated as $\hat{y}_i = \frac{y_i}{\max(Y) + \epsilon}$. Here, $\epsilon$ is a small positive constant to avoid division by zero and ensure stability, scaling all scores to a unified range for comprehensive cross-benchmark performance comparison.

\section{Extended Ablation Results}
\label{sec:extended_ablations}
\subsection{Main Results on Embodied and Affordance Benchmarks}
\label{sec:embodied_affordance_results}
For affordance, XEmbodied reaches 78.50 on Affordance and ranks first on RoboAfford at 4.35, showing strong affordance understanding for robotic manipulation. On the OOD Cosmos-R1 benchmark, XEmbodied (Full) reaches 81.50, edging Mimo-Embodied (81.00) without relying on textual chain-of-thought training.
\begin{table*}[t]
\centering
\setlength{\tabcolsep}{1pt}
\begin{tabular}{lcccccccc}
\toprule
\multirow{2}{*}{\textbf{Model}} & \multicolumn{7}{c}{\textbf{Benchmark Scores}} \\
\cmidrule(lr){2-8}
& \makecell{Affordance-\\2K} & \makecell{Robo-\\Afford} & \makecell{Cosmos-\\R1} & \makecell{Embodied-\\R1} & \makecell{RoboRefit-\\Bench} & \makecell{VABench-\\Point} & \makecell{Where2-\\place} \\
\midrule
\multicolumn{8}{l}{\textbf{Proprietary Models}} \\
GPT-4o~\cite{openai2024gpt4ocard}          & 2.70 & 3.80 & 67.40 & \underline{0.35} & 13.40 & 0.40 & 0.44 \\
Gemini-1.5~\cite{geminiteam2024gemini15unlockingmultimodal} & 5.19 & \underline{4.20} & 71.10 & $-$ & 35.50 & 0.80 & 0.89 \\
\midrule
\multicolumn{8}{l}{\textbf{Open-Source Models}} \\
Qwen2.5-VL-7B~\cite{bai2025qwen2}          & 8.70 & 3.00 & 72.50 & 0.25 & 76.40 & 0.75 & 0.90 \\
Qwen2.5-VL-32B~\cite{bai2025qwen2}         & \underline{9.10} & 3.25 & 72.50 & 0.00 & 77.60 & 1.45 & 1.25 \\
Qwen3-VL-A3B-30B~\cite{bai2025qwen3} [3B]  & 8.85 & 3.65 & 74.50 & 0.05 & \underline{83.95} & 1.10 & 1.51 \\
Qwen3-VL-32B~\cite{bai2025qwen3}           & 6.90 & 4.00 & 75.50 & 0.05 & 83.15 & 1.84 & 1.50 \\
\midrule
\multicolumn{8}{l}{\textbf{Embodied Models}} \\
Cosmos-R1~\cite{azzolini2025cosmos}        & 6.63 & 1.95 & \underline{72.00} & 0.05 & 34.75 & 0.35 & 0.45 \\
Mimo-Embodied~\cite{hao2026mimo}           & 8.85 & 3.25 & \textbf{81.00} & 0.00 & 78.35 & \underline{1.85} & 1.25 \\
\midrule
\multicolumn{8}{l}{\textbf{Our Model}} \\
\rowcolor{waymolgray} XEmbodied (Full), 30B [3B] & \textbf{78.50} & \textbf{4.35} & \textbf{81.50} & \textbf{3.80} & \textbf{87.15} & \textbf{3.50} & \textbf{2.30} \\
\bottomrule
\end{tabular}
\caption{Comparison of XEmbodied with other models on Embodied \& Affordance. Size denotes total parameters (active params for MoE in brackets). The best results of the listed models are \textbf{bolded} and the second-best is \underline{underlined}.}
\label{tab:extended_results_ea}
\end{table*}

\subsection{OOD Ablation under the Same Backbone}
\label{sec:ood_ablation}
To directly address the concern that XEmbodied's gains might stem from backbone scale rather than the proposed modules, we provide a same-backbone OOD ablation in Table~\ref{tab:sup_ood}. All rows share the Qwen3-VL-30B-A3B backbone (a MoE model with $\sim$3B active parameters per token, compute-comparable to the 7B baselines Mimo and Cosmos-R1). The weighted 16-benchmark average improves from 71.48 (backbone only) to 80.18 (full model), an 8.7-point gain attributable purely to the proposed modules and training strategy. On the four strictly-OOD metrics, the full EIEA-enhanced model surpasses the prior best baselines on three (DriveBench 57.33 vs.\ 53.06, Ego3D RMSE 6.83 vs.\ 15.70, Cosmos-R1 81.50 vs.\ 81.00) and is competitive on the fourth (Ego3D ACC 59.53 vs.\ Qwen3-VL-32B's 59.93). We report these results transparently rather than claiming uniform SOTA.

\begin{table}[htbp]
\centering
\scriptsize
\setlength{\tabcolsep}{2pt}
\caption{OOD ablation under the same Qwen3-VL-30B-A3B backbone. Prior best baselines: DriveB.\ from Qwen2.5-VL-32B; Ego3D A./R.\ from Qwen3-VL-32B; Cosmos-R1 from Mimo-Embodied.}
\label{tab:sup_ood}
\begin{tabular}{lccccc}
\toprule
Config & DriveB.\ $\uparrow$ & Ego3D A.\ $\uparrow$ & Ego3D R.\ $\downarrow$ & CosR1 $\uparrow$ & 16-bmk Avg \\
\midrule
\textcolor{gray!75}{Best prior baseline} & \textcolor{gray!75}{53.06} & \textcolor{gray!75}{59.93} & \textcolor{gray!75}{15.70} & \textcolor{gray!75}{81.00} & \textcolor{gray!75}{---} \\
\midrule
Backbone & 46.22 & 53.15 & 13.84 & 74.50 & 71.48 \\
\quad+Data & 47.05 & 49.65 & 9.40 & 74.10 & 75.68 \\
\rowcolor{gray!25} \quad+3DA only & 44.97 & 54.69 & 8.62 & 73.00 & 71.92 \\
\quad+Data+3DA & 53.18 & 55.28 & 9.25 & 76.00 & 78.45 \\
\rowcolor{waymolgray} \quad+Data+3DA+EIEA & \textbf{57.33} & \textbf{59.53} & \textbf{6.83} & \textbf{81.50} & \textbf{80.18} \\
\bottomrule
\end{tabular}
\end{table}

A few observations: (i) 3DA alone does not uniformly improve every metric; on DriveBench it dips from 46.22 to 44.97, which we read as a precondition rather than a defect: 3DA injects a continuous geometric prior that needs a geometrically-aligned answer space to land on, and without curated data that anchor is missing. Once data curation is added, the same 3DA lifts the average from 75.68 to 78.45, and on the OOD Ego3D RMSE probe 3DA alone already cuts the error from 13.84 to 8.62, confirming it does capture geometric structure. (ii) EIEA further improves cross-distribution transfer, with Ego3D RMSE dropping from 9.25 to 6.83 (a 56.5\% reduction versus the prior best baseline of 15.70); the cosine-similarity analysis in Section~\ref{sec:3da_eiea_complement} confirms effective use of physical tokens. (iii) All gains are obtained under the same backbone, ruling out parameter-count as the dominant factor. OOD figures throughout the paper use the Full model (e.g., DriveBench 57.33); the 0.12-point DriveBench difference sometimes cited is the SFT variant (53.18 vs.\ 53.06), not the Full model used for OOD claims.

\begin{table*}[t]
\centering
\scriptsize
\caption{Performance and Inference Efficiency of EIEA (Extended). All efficiency metrics are measured under the same experimental setup: 32x H20 GPU, batch size=1.}
\setlength{\tabcolsep}{2.5pt}
\begin{tabular}{lcccccccccccc}
\toprule
Model & \makecell{Size \\ (B)} & \makecell{EIEA \\ S1} & \makecell{EIEA \\ S2} & \makecell{3DA} &
\multicolumn{2}{c}{\textbf{DriveLMM-o1}} & \multicolumn{2}{c}{\textbf{DriveBench}} &
\multicolumn{3}{c}{\textbf{Ego3D-Bench}} \\
\cmidrule(lr){6-7} \cmidrule(lr){8-9} \cmidrule(lr){10-12}
& & & & & Score & Time (s) & Score & Time (s) & Acc & RMSE & Time (s) \\
\midrule
Baseline & 30 & \xmark & \xmark & \xmark & 64.32 & 1.52 & 53.18 & 1.64 & 55.28 & 9.44 & 1.25 \\
AgentThink & 7 & \xmark & \xmark & \xmark & 71.35 & 3.17 & 55.41 & 3.36 & - & - & - \\
XEmbodied (S1+3DA) & 30 & \cellcolor{waymolgray}\cmark & \xmark & \cellcolor{waymolgray}\cmark  & 75.28 & - & 56.72 & - & 55.45 & 6.85 & - \\
XEmbodied (S1+S2) & 30 & \cellcolor{waymolgray}\cmark & \cellcolor{waymolgray}\cmark & \xmark & 69.01 & 0.38 & 51.36 & 0.44 & 56.71 & 7.12 & 0.35 \\
\rowcolor{waymolgray}
XEmbodied (Full) & 30 & \cmark & \cmark & \cmark & \textbf{77.01} & \textbf{0.46} & \textbf{57.33} & \textbf{0.52} & \textbf{59.53} & \textbf{6.83} & \textbf{0.40} \\
\bottomrule
\end{tabular}
\label{tab:sup_EIEA}
\end{table*}

\subsection{Experimental Questions}
Before presenting the detailed ablations, we summarize the questions that these experiments are designed to answer:

    \begin{itemize}
        \item[\textit{1.}] \textit{Data Source:} Why do we fuse multi-source training data from autonomous driving, robotic manipulation, and general visual tasks? This experiment answers the fundamental question: \textbf{Is the diversity of multi-modal scene data a necessary prerequisite for enhancing the model's ability to understand physical-world information?} We systematically ablate different data combinations to verify that multi-source data provides complementary physical priors that single-domain data cannot offer.
        
        \item[\textit{2.}] \textit{Module Synergy:} What is the intrinsic coordination mechanism between the Efficient Image-Embodied Adapter (EIEA) and the 3D Adapter (3DA)? Building on the main paper's analysis, this experiment further addresses: \textbf{Is 3DA an indispensable prerequisite for activating EIEA's physical information modeling capability?} We control the on/off states of EIEA-S1/S2 and 3DA to reveal that EIEA's full potential can only be unleashed with geometric representation guidance.
        
        \item[\textit{3.}] \textit{Model Scale:} Why do we select the 30B-scale model as our backbone instead of smaller alternatives (e.g., 3B, 4B, 7B, 8B) or larger ones (e.g., 72B)? Notably, we do not evaluate 72B-scale models due to their prohibitive computational overhead and deployment costs. This experiment tackles: \textbf{Is there an optimal trade-off between model capacity, architectural adaptability, and efficiency for physical information understanding?} We evaluate models of varying sizes to demonstrate that the 30B model (adopting a Mixture-of-Experts (MoE) architecture) achieves the best balance: it not only strikes a favorable trade-off between performance gain and computational cost, but also inherently supports integrating diverse tasks from broader domains—an essential property for our multi-source data fusion design.
    \end{itemize}

\subsection{Ablation study on the coordination effect between 3DA and EIEA}

This ablation study systematically analyzes the synergistic mechanism between the 3D Adapter and the Efficient Image-Embodied Adapter (EIEA), with the goal of investigating a core question: \textit{whether geometric priors are indispensable for EIEA to effectively model physical-world information}. Key observations from Table \ref{tab:sup_EIEA} are summarized as follows:

\noindent 1. Baseline Performance and Comparative Context
The vanilla XEmbodied model (without EIEA/3DA) achieves a DriveLMM-o1 score of 64.32, a DriveBench score of 53.18, and an Ego3D-Bench accuracy of 55.28, providing a stable reference for evaluating architectural improvements. AgentThink, a representative comparative method, obtains a higher DriveLMM-o1 score (71.35) but lacks complete evaluation results on Ego3D-Bench (marked as ``-''), which highlights the limited ability of existing methods to unify physical information modeling across diverse benchmarks.

\noindent 2. Initial Evidence for 3DA-Dependent EIEA Effectiveness
Enabling only EIEA-S1 together with 3DA leads to clear performance improvements over the vanilla XEmbodied model (DriveLMM-o1: 64.32 $\rightarrow$ 75.28; DriveBench: 53.18 $\rightarrow$ 56.72). This result suggests that EIEA shows potential for physical information modeling, and such potential is initially supported by the geometric representations provided by 3DA. These observations provide preliminary evidence that geometric priors help EIEA achieve basic effectiveness.

\noindent 3. Performance Degradation of EIEA in the Absence of 3DA
A notable observation is that enabling both EIEA-S1 and EIEA-S2 \textit{without 3DA} leads to substantial performance drops: the DriveLMM-o1 score decreases to 69.01 (10.4\% lower than the setup with 3DA and dual EIEA), and the DriveBench score falls to 51.36, even below the vanilla baseline. This indicates that the full function of EIEA is not self-sufficient. Without geometric guidance from 3DA, EIEA struggles to correctly model spatial-physical relationships, resulting in ineffective feature enhancement and visible performance regression. This outcome supports our hypothesis that geometric representation acts as an important prerequisite for the dual EIEA mechanism to function properly.

\noindent 4. Balanced Performance via 3DA + Dual EIEA
When 3DA is combined with both EIEA-S1 and EIEA-S2, the model achieves competitive and well-balanced performance across all evaluated benchmarks: DriveLMM-o1 (77.01, +12.69 over vanilla), DriveBench (57.33, +4.15 over vanilla), and Ego3D-Bench (59.53 accuracy, 6.83 RMSE, the lowest error among configurations). Although this setup introduces a small inference-time overhead (0.40–0.52s, compared with 0.35–0.44s for dual EIEA without 3DA), the performance improvements are consistent and meaningful. This trade-off indicates that 3DA is not merely an auxiliary module, but a key enabling component that unlocks stronger physical information understanding in EIEA.

\noindent 5. Summary of Findings
This ablation study provides consistent support for the claim that integrating geometric representations (via 3DA) into the foundation model improves its ability to understand physical information. In particular, 3DA acts as a critical enabler for EIEA: without 3DA, EIEA is unable to fully realize its design objectives; with 3DA, the dual-stage enhancement mechanism of EIEA yields more favorable performance, supporting the effectiveness of our proposed coordination framework between geometric and embodied information modules.

\subsection{Ablation study on model size and data source}

\begin{table*}[htbp]
\centering
\caption{Performance of Qwen3VL/Qwen2.5VL Variants on Spatial \& 3D Understanding. 
  (a) Qwen2.5-VL (3B); (b) Qwen2.5-VL (7B); (c) Qwen3-VL (4B); 
  (d) Qwen3-VL (8B); (e) Qwen3-VL (30B); (f) Qwen3-VL (30B); (g) Qwen3-VL (30B). All model are trained within the same device conditions with enough hyperparameter tuning.}
\label{tab:qwen_spatial_3d}
\begin{tabular}{lcccccccc}
\toprule
 & Size & Drive & Robot & Ego3D-ACC & SURDS   & VLADBench & STRIDE-QA   \\
\midrule
(a) & 3B  & \cellcolor{waymolgray}\cmark & \cellcolor{waymolgray}\cmark & 38.42 & 9.96 & 59.26 & 4.51 \\
(b) & 7B  & \cellcolor{waymolgray}\cmark & \cellcolor{waymolgray}\cmark & 39.92 & 23.25 & 60.43 & \underline{4.95} \\
(c) & 4B  & \cellcolor{waymolgray}\cmark & \cellcolor{waymolgray}\cmark & 37.07 & 28.75 & 60.69 & 4.02 \\
(d) & 8B  & \cellcolor{waymolgray}\cmark & \cellcolor{waymolgray}\cmark & 33.10 & 34.81 & 63.61 & 2.05 \\
(e) & 30B & \xmark & \cellcolor{waymolgray}\cmark & \textbf{51.88} & \underline{39.35} & 59.26 & 1.00 \\
(f) & 30B & \cellcolor{waymolgray}\cmark & \xmark & 40.18 & 32.17 & \textbf{70.09} & 4.38 \\
(g) & 30B & \cellcolor{waymolgray}\cmark & \cellcolor{waymolgray}\cmark & \underline{49.65} & \textbf{82.00} & \underline{68.72} & \textbf{6.07} \\
\bottomrule
\end{tabular}
\end{table*}

\begin{table*}[htbp]
\centering
\caption{Performance of Qwen3VL/Qwen2.5VL Variants on Semantic \& Reasoning. 
  (a) Qwen2.5-VL (3B); (b) Qwen2.5-VL (7B); (c) Qwen3-VL (4B); 
  (d) Qwen3-VL (8B); (e) Qwen3-VL (30B); (f) Qwen3-VL (30B); (g) Qwen3-VL (30B).}
\label{tab:qwen_semantic_reasoning}
\begin{tabular}{lcccccccccc}
\toprule
 & Size & Drive & Robot & DriveBench   & DriveLMM-o1   & MapLM-v2 & LingoQA   & Omnidrive   \\
\midrule
(a) & 3B  & \cellcolor{waymolgray}\cmark & \cellcolor{waymolgray}\cmark & 33.08 & 62.69 & 77.20 & 59.00 & 6.73 \\
(b) & 7B  & \cellcolor{waymolgray}\cmark & \cellcolor{waymolgray}\cmark & 40.17 & 64.51 & 77.97 & 58.10 & 14.37 \\
(c) & 4B  & \cellcolor{waymolgray}\cmark & \cellcolor{waymolgray}\cmark & 32.87 & 63.72 & 78.75 & 53.70 & 5.71 \\
(d) & 8B  & \cellcolor{waymolgray}\cmark & \cellcolor{waymolgray}\cmark & 29.15 & \underline{64.78} & 78.58 & 61.10 & 7.23 \\
(e) & 30B & \xmark & \cellcolor{waymolgray}\cmark & 33.26 & 56.29 & 49.13 & \textbf{65.20} & 1.01 \\
(f) & 30B & \cellcolor{waymolgray}\cmark & \xmark & \textbf{51.46} & \textbf{64.91} & \textbf{79.22} & 58.50 & \textbf{22.72} \\
(g) & 30B & \cellcolor{waymolgray}\cmark & \cellcolor{waymolgray}\cmark & \underline{49.07} & 60.00 & \underline{78.95} & \underline{64.80} & \underline{18.21} \\
\bottomrule
\end{tabular}
\end{table*}

\begin{table*}[htbp]
\centering
\caption{Performance of Qwen3VL/Qwen2.5VL Variants on Embodied \& Affordance. 
  (a) Qwen2.5-VL (3B); (b) Qwen2.5-VL (7B); (c) Qwen3-VL (4B); 
  (d) Qwen3-VL (8B); (e) Qwen3-VL (30B); (f) Qwen3-VL (30B); (g) Qwen3-VL (30B).}
\label{tab:qwen_embodied_affordance}
\begin{tabular}{lcccccccccccc}
\toprule
 & Size & Drive & Robot & \makecell{Part-\\Afford  } & \makecell{Robo-\\Afford  } & \makecell{Cosmos-\\R1  } & \makecell{Embodied-\\R1  } & \makecell{Robo-\\Refit  } & \makecell{VABench-\\Point  } & \makecell{Where2-\\place  } \\
\midrule
(a) & 3B  & \cellcolor{waymolgray}\cmark & \cellcolor{waymolgray}\cmark & 3.35 & 1.90 & \textbf{62.00} & \underline{0.15} & 65.15 & 0.50 & 0.85 \\
(b) & 7B  & \cellcolor{waymolgray}\cmark & \cellcolor{waymolgray}\cmark & 6.95 & 2.45 & 60.00 & 0.05 & 68.60 & 0.45 & 0.75 \\
(c) & 4B  & \cellcolor{waymolgray}\cmark & \cellcolor{waymolgray}\cmark & 8.85 & 3.15 & 55.00 & 0.00 & 84.00 & 1.05 & 1.10 \\
(d) & 8B  & \cellcolor{waymolgray}\cmark & \cellcolor{waymolgray}\cmark & 8.45 & \underline{3.60} & 53.00 & 0.00 & 84.45 & \textbf{1.25} & 1.30 \\
(e) & 30B & \xmark & \cellcolor{waymolgray}\cmark & 8.85 & \underline{3.60} & 58.00 & 0.00 & \textbf{86.80} & 0.85 & \underline{1.40} \\
(f) & 30B & \cellcolor{waymolgray}\cmark & \xmark & \textbf{11.35} & 3.50 & \textbf{62.00} & 0.05 & 80.00 & 0.80 & 1.20 \\
(g) & 30B & \cellcolor{waymolgray}\cmark & \cellcolor{waymolgray}\cmark & \underline{11.05} & \textbf{3.80} & \underline{59.00} & \textbf{0.25} & \underline{86.60} & \underline{1.20} & \textbf{1.60} \\
\bottomrule
\end{tabular}
\end{table*}

This ablation experiment is designed to address two core research questions: why the 30B-scale Qwen3-VL model is selected as the baseline instead of smaller variants (3B/4B/7B/8B), and what contributions autonomous driving (Drive) and robotic interaction (Robot) data make to the multi-task performance of Qwen3-VL. The experimental results are presented in Tables \ref{tab:qwen_spatial_3d}, \ref{tab:qwen_semantic_reasoning} and \ref{tab:qwen_embodied_affordance}.

Small-scale models (3B/4B/7B/8B) are constrained by parameter capacity and struggle to achieve consistent and competitive performance across spatial understanding, semantic reasoning, and embodied affordance tasks. As shown in Table \ref{tab:qwen_spatial_3d}, the Ego3D-ACC score of the 3B Qwen2.5-VL is only 38.42\% and the SURDS score is merely 9.96\%, both far lower than the top values achieved by the 30B Qwen3-VL variants; even the 8B Qwen3-VL only reaches 33.10\% on Ego3D-ACC, leaving a clear performance gap with the 30B model. In semantic and reasoning tasks shown in Table \ref{tab:qwen_semantic_reasoning}, small models exhibit obvious instability: the 3B Qwen2.5-VL achieves 59.00\% on LingoQA but only 33.08\% on DriveBench, while the 7B Qwen2.5-VL improves the DriveBench score to 40.17\% but drops the LingoQA score to 58.10\%, which reflects their limited capacity to simultaneously optimize multiple semantic tasks. For embodied and affordance tasks in Table \ref{tab:qwen_embodied_affordance}, the 3B and 7B models show basic but limited performance, with 3.35\% / 1.90\% and 6.95\% / 2.45\% on Part-Afford and Robo-Afford respectively. These values are noticeably lower than those of 30B-level variants, which reach up to 11.35\% on Part-Afford and 3.80\% on Robo-Afford. Similarly, the highest RoboRefit-Bench score among small models is around 65–68\%, also below the 86\%+ level achieved by the 30B models. All these results suggest that small-scale models lack sufficient representational capacity to absorb multi-domain knowledge, leading to relatively weak and unbalanced multi-task performance.

We further conduct ablation experiments on the 30B Qwen3-VL with three settings: without Drive data (woDrive), without Robot data (woRobot), and with full Drive+Robot data, to analyze the value of domain-specific data. The results indicate clear complementary effects of the two data types. As shown in Table \ref{tab:qwen_semantic_reasoning}, the DriveBench score drops sharply to 33.26\% without Drive data, which is 18.20 percentage points lower than the variant without Robot data, and the VLADBench score in Table \ref{tab:qwen_spatial_3d} also shows a notable decline, suggesting that Drive data effectively improves the model's performance on autonomous driving-related semantic reasoning and spatial localization tasks. For embodied interaction tasks, the variant without Robot data only scores 80.00\% on RoboRefit-Bench in Table \ref{tab:qwen_embodied_affordance}, which is 6.80 percentage points lower than the variant without Drive data, indicating that Robot data contributes meaningfully to the model's embodied affordance and robotic interaction capabilities.

Notably, the variant (f) trained only on Drive data achieves strong results on several autonomous driving-oriented benchmarks, while the full Drive+Robot variant (g) does not outperform all competitors on every individual metric. Instead, it yields the most balanced cross-task performance and favorable overall normalized average across spatial, semantic, and embodied task families. It achieves top-tier scores on key tasks including STRIDE-QA, SURDS, RoboAfford, and Where2place, while maintaining highly competitive results on all other benchmarks, demonstrating strong synergy of dual-domain data without overfitting to either single domain.

In conclusion, small-scale models are limited by parameter size and cannot maintain balanced and consistent performance across spatial understanding, semantic reasoning, and embodied interaction tasks. In contrast, the 30B Qwen3-VL model has sufficient capacity to integrate domain-specific knowledge from Drive and Robot data, where Drive data supports autonomous driving-related tasks and Robot data enhances embodied interaction capabilities. The 30B variant equipped with full Drive+Robot data shows the most favorable overall trade-off and balanced multi-task performance across all evaluated domains, which supports the rationality of selecting the 30B Qwen3-VL with dual-domain data as the baseline model.

\textbf{Ablation study on curriculum learning strategy}
As shown in Table~\ref{tab:extended_curriculum_ablation}, the raw Qwen model (a) reaches a weighted total average of 61.24, and naive mix training (b) raises it to 68.97. Our staged CL brings consistent gains at each stage: 69.66 with S1 (c), 80.83 with S2 (d), peaking at 90.13 with the full S1--S3 stages (e). This final result clearly beats naive mix training, confirming that gradual learning of complex concepts avoids catastrophic forgetting, stabilizes multimodal alignment, and lifts overall performance.

\begin{table*}[t]
\centering
\small
\setlength{\tabcolsep}{1pt}
\begin{tabular}{lcccccccccc}
\toprule
Group & \makecell{Mixture \\ Training} & CL & S1 & S2 & S3 & \makecell{Spatial \& 3D \\ Understanding} & \makecell{Semantic \& \\ Reasoning} & \makecell{Embodied \& \\ Affordance} & Avg \\
\midrule
(a) & \xmark & \xmark & \xmark & \xmark & \xmark & 59.56 & 63.41 & 60.59 & 61.24 \\
(b) & \cellcolor{waymolgray}\cmark & \xmark & \xmark & \xmark & \xmark & 69.14 & 73.75 & 64.76 & 68.97 \\
(c) & \xmark & \cellcolor{waymolgray}\cmark & \cellcolor{waymolgray}\cmark & \xmark & \xmark & 65.99 & 75.21 & 67.53 & 69.66 \\
(d)  & \xmark & \cellcolor{waymolgray}\cmark & \cellcolor{waymolgray}\cmark & \cellcolor{waymolgray}\cmark & \xmark & 69.63 & 81.75 & 88.05 & 80.83 \\
\rowcolor{waymolgray} (e) & \xmark & \cmark & \cmark & \cmark & \cmark & \textbf{90.41} & \textbf{82.71} & \textbf{96.29} & \textbf{90.13} \\
\bottomrule
\end{tabular}
\caption{Ablation on curriculum learning (CL). Bold values denote the best result in each column.}
\label{tab:extended_curriculum_ablation}
\end{table*}

\subsection{Performance on Planning Task}

To fully verify the effectiveness and superiority of our proposed method XEmbodied-FT, we conduct a comprehensive and in-depth evaluation on the nuScenes dataset. This dataset is widely recognized as a benchmark for autonomous driving planning tasks. The evaluation adopts an open-loop setting and focuses on comparing our model with state-of-the-art end-to-end planning models. These models include text-based and action-based driving models, all of which are trained on automatically annotated data. RoboTron-Drive\cite{2024RoboTron} serves as an important baseline to verify the performance improvement of the proposed XEmbodied-FT model. The detailed experimental results are summarized in Table~\ref{tab:planning_eval_nuscenes}. We conduct in-depth analysis from multiple dimensions below, with a focus on illustrating that our XEmbodied-FT model can effectively handle various downstream planning tasks.

First, from the perspective of overall performance, our XEmbodied-FT model demonstrates remarkable competitiveness in all core evaluation metrics. This fully verifies its ability to stably and effectively tackle downstream planning tasks. L2 Error is a key indicator reflecting the accuracy of trajectory prediction, which is a core downstream task of planning. Our model achieves the best performance at the 2s and 3s prediction horizons, with L2 Error values of 0.25m and 0.45m respectively. Meanwhile, it ties with SOLVE-VLM\cite{11094576} for the second-best average L2 Error of 0.28m. This indicates that our model can generate trajectory predictions highly consistent with the real trajectory in both medium and long-term planning scenarios. Such consistency is crucial for downstream planning tasks like path tracking and dynamic obstacle avoidance. Collision Rate directly reflects the safety of planning decisions, another core downstream task of planning. Our model achieves a 2s collision rate of 0.09\%, only 0.01 percentage points behind the best baseline (FASIONAD, 0.08\%) and lower than that of every action-based planner. This proves that our model can effectively balance trajectory accuracy and driving safety in short-to-medium-term planning. This balance is an essential requirement for the practical application of downstream planning tasks. Although our model shows a slight gap in the 3s collision rate and average collision rate compared with the best-performing SOLVE-VLM, the 3s collision rate of SOLVE-VLM is 0.43\% and the average collision rate is 0.20\%, while ours are 0.55\% and 0.22\% respectively. This gap is within an acceptable range and provides a clear direction for future optimization.

\begin{table*}[htbp]
\centering
\caption{Comparison of end-to-end planning models trained on automatically annotated data, evaluated in open-loop on the nuScenes dataset. The \textbf{bold} numbers indicate the best results, and the \underline{underlined} numbers indicate the second best.}
\label{tab:planning_eval_nuscenes}
\small
\setlength{\tabcolsep}{3.5pt}
\renewcommand{\arraystretch}{1.2}
\resizebox{\textwidth}{!}{%
\begin{tabular}{lcccccccccc}
\toprule
\multirow{2}{*}{\textbf{Model}} & \multirow{2}{*}{\textbf{Publication}} & \multirow{2}{*}{\textbf{Ego }} & \multicolumn{4}{c}{\textbf{L2 Error (m) $\downarrow$}} & \multicolumn{4}{c}{\textbf{Collision Rate (\%) $\downarrow$}} \\
\cmidrule(lr){4-7} \cmidrule(lr){8-11}
 &  &  & \textbf{1s} & \textbf{2s} & \textbf{3s} & \textbf{Avg.} & \textbf{1s} & \textbf{2s} & \textbf{3s} & \textbf{Avg.} \\
\midrule
\multicolumn{11}{l}{\textbf{Text-Based Driving Models}} \\
\midrule
Qwen3-VL-A3B\cite{bai2025qwen3} & -- & \ding{55} & 4.39 & 6.88 & 9.44 & 6.91 & 3.61 & 8.17 & 15.80 & 9.19 \\
RoboTron-Drive\cite{2024RoboTron} & ICCV2025 & \checkmark & 0.14 &0.30 &0.57  &0.33 & 0.03 & 0.12 & 0.63  &0.26 \\
DriveVLM\cite{tian2024drivevlm} & CoRL 2024 & \checkmark & 0.18 & 0.34 & 0.68 & 0.40 & -- & -- & -- & -- \\
DriveVLM-Dual\cite{tian2024drivevlm} & CoRL 2024 & \checkmark & 0.15 & 0.29 & 0.48 & 0.31 & -- & -- & -- & -- \\
OmniDrive\cite{wang2024omnidrive} & CVPR 2025 & \checkmark & \underline{0.14} & 0.29 & 0.55 & 0.33 & \textbf{0.00} & 0.13 & 0.78 & 0.30 \\
EMMA\cite{hwang2024emma} & TMLR & \checkmark & \underline{0.14} & 0.29 & 0.54 & 0.32 & -- & -- & -- & -- \\
EMMA+\cite{hwang2024emma} & TMLR & \checkmark & \textbf{0.13} & \underline{0.27} & 0.48 & \underline{0.29} & -- & -- & -- & -- \\
ImpromptuVLA\cite{chi2025impromptuvlaopenweights} & NeurIPS 2025 & \checkmark & \textbf{0.13} & \underline{0.27} & 0.53 & 0.30 & -- & -- & -- & -- \\
SOLVE-VLM\cite{11094576} & CVPR 2025 & \checkmark & \textbf{0.13} & \textbf{0.25} & \underline{0.47} & \textbf{0.28} & \textbf{0.00} & 0.16 & \textbf{0.43} & \textbf{0.20} \\
Agent-Driver\cite{mao2023language} & COLM2024 & \ding{55} &0.22 &0.65 & 1.34 &0.74 &0.02 &0.13 &0.48 &0.21 \\
FASIONAD\cite{qian2024fasionad, qian2025fasionad++} & ICRA2026 & \checkmark & \textbf{0.13} & 0.26 & \textbf{0.45} & \textbf{0.28} &0.05& \textbf{0.08}& \textbf{0.15} & \textbf{0.09} \\
\midrule
\multicolumn{11}{l}{\textbf{Action-Based Driving Models}} \\
\midrule
UniAD \cite{hu2023uniad} & CVPR 2023 & -- & 0.59 & 1.01 & 1.48 & 1.03 & 0.16 & 0.51 & 1.64 & 0.77 \\
VAD-Base \cite{jiang2023vad} & ICCV 2023 & -- & 0.69 & 1.22 & 1.83 & 1.25 & 0.06 & 0.68 & 2.52 & 1.09 \\
BEV-Planner\cite{li2024ego} & CVPR 2024 & -- & 0.30 & 0.52 & 0.83 & 0.55 & 0.10 & 0.37 & 1.30 & 0.59 \\
UniAD \cite{hu2023uniad} & CVPR 2023 & \checkmark & 0.20 & 0.42 & 0.75 & 0.46 & 0.02 & 0.25 & 0.84 & 0.37 \\
VAD-Base \cite{jiang2023vad} & ICCV 2023 & \checkmark & 0.17 & 0.34 & 0.60 & 0.37 & 0.04 & 0.27 & 0.67 & 0.33 \\
AD-MLP & arXiv 2023 & \checkmark & \textbf{0.00} & 0.32 & 0.59 & 0.35 & \textbf{0.00} & 0.27 & 0.85 & 0.37 \\
BEV-Planner++ & CVPR 2024 & \checkmark & 0.16 & 0.32 & 0.57 & 0.35 & \textbf{0.00} & 0.29 & 0.73 & 0.34 \\
SparseDrive \cite{sun2025sparsedrive} & CVPR 2024 & \checkmark & 0.38 & 0.65 & 0.95 & 0.66 & 0.15 & 0.30 & 0.90 & 0.45 \\
GenAD\cite{zheng2024genad} & ECCV 2024 & \checkmark & 0.35 & 0.61 & 0.91 & 0.62 & 0.13 & 0.28 & 0.88 & 0.43 \\
SOLVE-E2E\cite{11094576} & CVPR 2025 & \checkmark & \underline{0.14} & 0.28 & 0.50 & 0.31 & 0.04 & 0.17 & 0.68 & 0.30 \\
ColaVLA\cite{peng2026colavlaleveragingcognitivelatent} & CVPR2026 & \checkmark & \underline{0.14} & \underline{0.27} & \underline{0.50} & 0.30 & 0.04 & 0.17 & \underline{0.47} & \underline{0.23} \\
\midrule
\multicolumn{11}{l}{\textbf{Ours}} \\
\midrule
XEmbodied-FT & \textbf{--} & \textbf{\checkmark} & \underline{0.14} & \textbf{0.25} & \textbf{0.45} & \textbf{0.28} & 0.02 & \underline{0.09} & 0.55 & 0.22 \\
\midrule
\multicolumn{11}{l}{\textbf{Ours (Ablation)}} \\
\midrule
XEmbodied-FT & -- & \ding{55} & 0.26 & 0.60 & 1.08 & 0.64 & 0.10 & 0.49 & 1.84 & 0.81 \\
\bottomrule
\end{tabular}
}
\end{table*}

Next, we conduct a detailed comparative analysis with different types of baseline models. This further highlights the advantages of our XEmbodied-FT model in handling downstream planning tasks, especially when compared with our previously proposed autonomous driving foundation model RoboTron-Drive\cite{2024RoboTron}. It achieves an L2 Error of 0.14m at 1s, 0.30m at 2s, 0.57m at 3s, and an average of 0.33m. Its 2s collision rate is 0.12\% and the average collision rate is 0.26\%. Compared with RoboTron-Drive, our XEmbodied-FT model shows significant improvements in key metrics related to downstream planning tasks. The 2s L2 Error is reduced by 0.05m, falling from 0.30m to 0.25m. The 3s L2 Error is reduced by 0.12m, dropping from 0.57m to 0.45m. The 2s collision rate is reduced by 0.03\%, decreasing from 0.12\% to 0.09\%. This improvement fully demonstrates that on the basis of the prior foundation model, our proposed XEmbodied-FT further optimizes the ability to handle downstream planning tasks. This is particularly evident in medium and long-term trajectory prediction accuracy and short-term driving safety. Compared with other advanced text-based models including EMMA+\cite{hwang2024emma}, ImpromptuVLA\cite{chi2025impromptuvlaopenweights}, SOLVE-VLM\cite{11094576} and FASIONAD\cite{qian2024fasionad,qian2025fasionad++}, our XEmbodied-FT model still maintains obvious advantages in long-horizon trajectory prediction at 2s and 3s. Its collision rate is also at a leading level. This confirms that our model can effectively adapt to the complex demands of text-based driving planning downstream tasks. It can understand natural language instructions and generate accurate and safe trajectories accordingly.

The action-based driving model group focuses more on direct trajectory generation and control signal output, which are core downstream planning tasks. Our XEmbodied-FT model also achieves outstanding performance in this group. Compared with state-of-the-art models like SOLVE-E2E~\cite{11094576}, ColaVLA~\cite{peng2026colavlaleveragingcognitivelatent}, and VAD-Base~\cite{jiang2023vad}, our model achieves the lowest L2 Error at the 2s and 3s horizons, with values of 0.25m and 0.45m respectively. This is particularly critical for downstream planning tasks such as long-horizon path planning and dynamic scene adaptation. For example, SOLVE-E2E achieves 0.28m at 2s and 0.50m at 3s, while ColaVLA achieves 0.27m at 2s and 0.50m at 3s. Our model’s trajectory prediction is more accurate in medium and long-term scenarios. This accuracy can effectively avoid trajectory deviation caused by cumulative errors and provide more reliable support for subsequent vehicle control. In terms of collision rate, although our model is not the best in the 3s and average indicators, its 2s collision rate of 0.09\% is still better than most action-based models. This indicates that our model can effectively guarantee driving safety in short-to-medium-term planning scenarios. This safety guarantee is a key requirement for downstream planning tasks such as emergency avoidance and lane changing.

To further clarify the key factors that enable our XEmbodied-FT model to effectively handle downstream planning tasks, we conduct an ablation study. We remove the ego-vehicle historical information module and refer to this variant as XEmbodied-FT without ego history. The results show a substantial performance drop across all evaluation metrics. The average L2 Error increases from 0.28m to 0.64m, and the average collision rate rises from 0.22\% to 0.81\%. This performance is even worse than that of the prior foundation model RoboTron-Drive. This ablation result highlights the critical importance of incorporating ego-vehicle historical information for accurate trajectory prediction and safe planning decisions. By effectively utilizing the historical motion state of the ego vehicle, our model can better capture the motion trend of the vehicle itself. This ability allows it to generate more consistent and accurate trajectories while effectively reducing the collision risk. This further verifies that our model is designed with full consideration of the actual needs of downstream planning tasks. Each module is optimized to improve the performance of planning tasks.

In summary, the comprehensive experimental results fully validate that our proposed XEmbodied-FT model can effectively handle various downstream planning tasks of autonomous driving. These core tasks include trajectory prediction, safety collision avoidance and long-horizon planning. Compared with our previously proposed autonomous driving foundation model RoboTron-Drive, XEmbodied-FT achieves significant performance improvements in key metrics. This proves the effectiveness of our optimization strategy. Compared with other state-of-the-art text-based and action-based driving models, our model maintains outstanding advantages in medium and long-term trajectory prediction accuracy and short-term driving safety. This lays a solid foundation for the practical application of end-to-end planning models in autonomous driving. The slight performance gap in the 3s collision rate also suggests opportunities for future improvements. We will focus on optimizing the model’s ability to handle complex long-horizon driving scenarios. Examples of such scenarios include dynamic obstacle interaction and road condition changes. This optimization will further enhance the stability and safety of the model in downstream planning tasks and narrow the performance gap with the best models in all metrics.

\section{Limitations and Future Work}
\label{sec:limitations}

Despite the promising performance of the proposed model across spatial understanding, semantic reasoning, and embodied affordance tasks (Tables \ref{tab:qwen_spatial_3d}–\ref{tab:qwen_embodied_affordance}), built on the Foundation-ORM reward mechanism and cross-modal alignment architecture, several non-trivial limitations remain. These limitations point to clear and actionable directions for future research, which we elaborate below alongside corresponding improvement paths to further advance the model’s capabilities in multi-modal and multi-task scenarios.

First, the model currently lacks effective long-term spatiotemporal reasoning and cognition capabilities. As demonstrated in the experimental results, the current framework mainly relies on single-frame visual information for task processing, failing to capture the temporal correlations and spatial evolution patterns in continuous scenes. For instance, in autonomous driving scenarios, it cannot accurately predict the movement trajectories of dynamic objects over extended periods; in robotic interaction tasks requiring sequential operation planning, it struggles to generate coherent and goal-oriented action sequences. To address this, future work will focus on integrating long-temporal modeling into the multi-modal framework—specifically, incorporating video sequence data to learn spatiotemporal dependencies, designing temporal attention mechanisms to emphasize critical time steps, and leveraging reinforcement learning paradigms (such as the R1-based training strategy) to optimize long-horizon planning and decision-making processes, complementing the existing cross-modal fusion pipeline.

Second, the model’s generalization ability to complex and extreme scenarios is limited. The existing training data, including both Drive and Robot datasets, have insufficient coverage of rare events, harsh environmental conditions, and multi-object interactive scenarios. This leads to noticeable performance degradation when the model faces such unseen complex inputs. Rare events include sudden obstacles in driving scenarios; harsh conditions cover heavy rain, fog, low light, and unstructured terrain; multi-object interactions involve collaborative robotic manipulation with multiple agents. Future research will prioritize data augmentation and dataset expansion: on one hand, we will generate synthetic data for extreme scenarios using simulation tools to supplement real-world data scarcity; on the other hand, we will collect and annotate real-world complex scene data to enhance the diversity of training samples. Additionally, domain adaptation techniques will be employed to reduce the distribution gap between training and test data, further improving the model’s robustness across different scenarios while aligning with its multi-task adaptation design.

Third, there is significant room for optimization in reasoning efficiency and computational cost. While the 30B-scale model achieves strong performance, its reasoning process involves explicit multi-step visual-language fusion, which results in high computational overhead and slow inference speed—limiting its deployment in resource-constrained scenarios. Future work will explore implicit visual reasoning methods to streamline the inference chain: this includes designing lightweight multi-modal fusion modules, adopting knowledge distillation to compress the model size without sacrificing performance, and leveraging implicit representation learning to reduce redundant computational steps. The goal is to achieve a better balance between reasoning accuracy and efficiency, enabling the model to be applied in low-latency practical scenarios while preserving the core cross-modal alignment capabilities.

Fourth, the model’s depth of knowledge integration and chain-of-thought reasoning is insufficient. Currently, the model tends to rely on surface-level visual and linguistic features for task solving, lacking structured logical reasoning capabilities for complex tasks that require multi-step deduction. To address this, future research will introduce chain-of-thought (CoT) prompting to guide the model in generating step-by-step reasoning processes, and integrate reinforcement learning to optimize the reasoning chain based on task-specific rewards. Additionally, we will incorporate external knowledge bases related to autonomous driving and robotic interaction to enrich the model’s domain knowledge, enabling it to tackle more complex semantic reasoning and decision-making tasks that require prior knowledge, and extending the Foundation-ORM’s task adaptation scope.

In summary, the proposed model lays a solid foundation for multi-modal multi-task learning in autonomous driving and robotic interaction scenarios, with its core strengths in spatial understanding, semantic reasoning, and embodied affordance. However, its limitations in spatiotemporal reasoning, complex scenario generalization, inference efficiency, and reasoning depth highlight key areas for improvement. Future work will focus on addressing these limitations through advanced modeling techniques, data enhancement, and efficiency optimization, aiming to develop a more robust, efficient, and intelligent multi-modal model that can better adapt to real-world practical demands while inheriting the advantages of the existing framework.
\section{Conclusion}
\label{sec:con}
We present XEmbodied, a cloud-side foundation model that equips VLMs with intrinsic 3D geometric awareness and compact physical cue integration through two complementary adapters: 3DA for endogenous geometric fusion and EIEA for physical signal distillation. Together with a progressive domain curriculum and GRPO post-training, these components establish a unified framework that bridges high-level semantic reasoning with geometric and physically grounded scene understanding. Extensive experiments across 16 public benchmarks demonstrate state-of-the-art performance in spatial understanding, traffic semantics, and embodied reasoning, while evaluations on three strictly out-of-distribution benchmarks---DriveBench, Ego3DBench, and Cosmos-R1---further verify its generalization to unseen scenarios and tasks. Moreover, downstream transfer to nuScenes planning shows that the learned spatial competence extends beyond VQA benchmarks to decision-oriented tasks, supporting the use of XEmbodied as a scalable annotation and reasoning engine for embodied data closed-loop systems.

\appendix
\section{Representative Case Studies}
\label{app:cases}
This appendix retains only representative qualitative cases and data-curation examples. The quantitative architecture, curation, training, and ablation details are part of the main technical report.
\begin{figure*}[tb]
  \centering
  \includegraphics[width=0.84\linewidth]{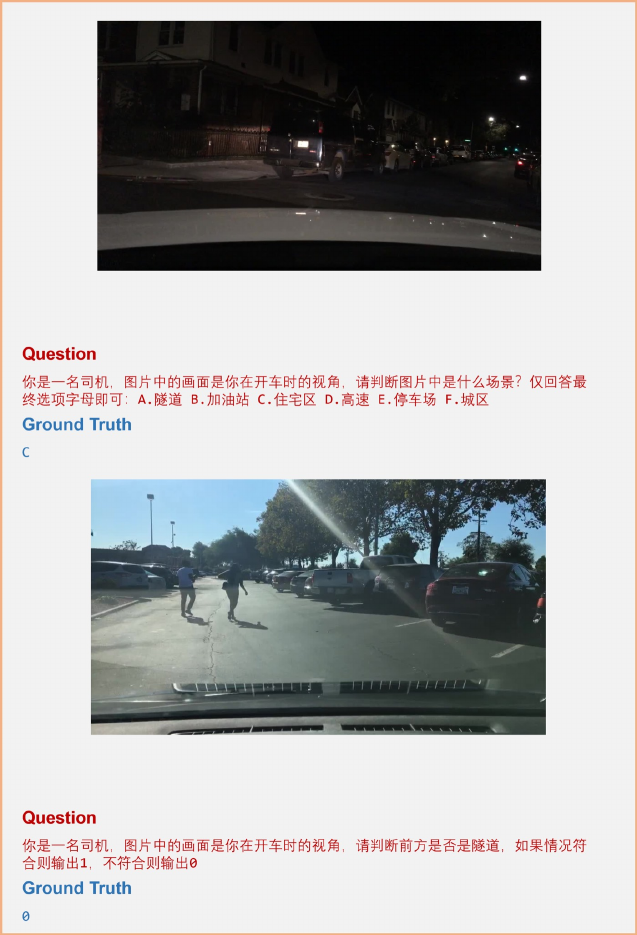}
  \caption{T1: Tiered data examples from Bdd.
  }
  \label{fig:data_tiered_bdd}
\end{figure*}

\begin{figure*}[tb]
  \centering
  \includegraphics[width=0.84\linewidth]{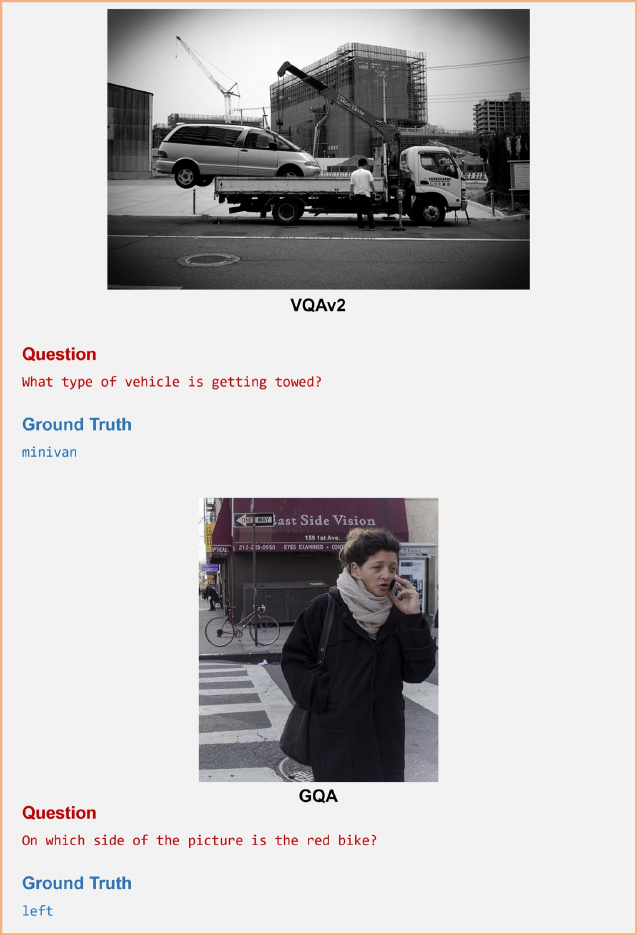}
  \caption{T1: Tiered data examples from VQAv2 and GQA.
  }
  \label{fig:data_tiered_gqavqa}
\end{figure*}

\begin{figure*}[tb]
  \centering
  \includegraphics[width=0.84\linewidth]{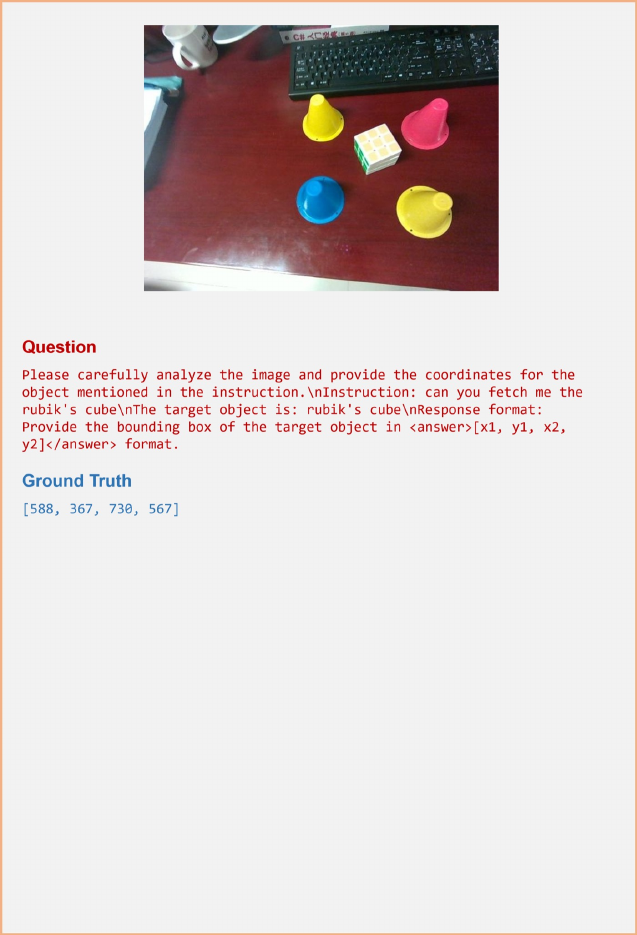}
  \caption{T2: Tiered data example from Roborefit.
  }
  \label{fig:data_tiered_roborefit}
\end{figure*}

\begin{figure*}[tb]
  \centering
  \includegraphics[width=0.84\linewidth]{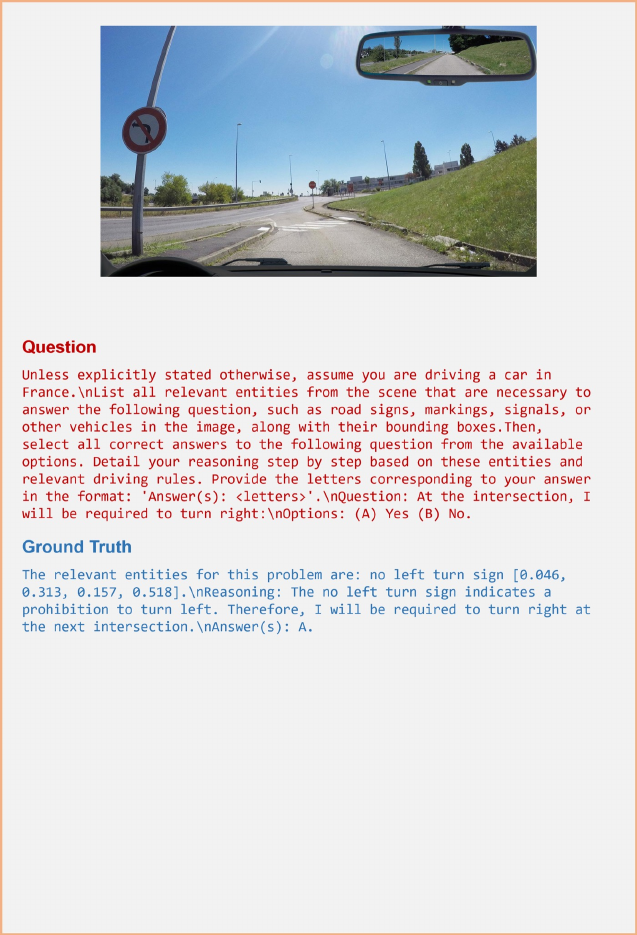}
  \caption{T2: Tiered data example from DrivingVQA.
  }
  \label{fig:data_tiered_drivingvqa}
\end{figure*}

\begin{figure*}[tb]
  \centering
  \includegraphics[width=0.84\linewidth]{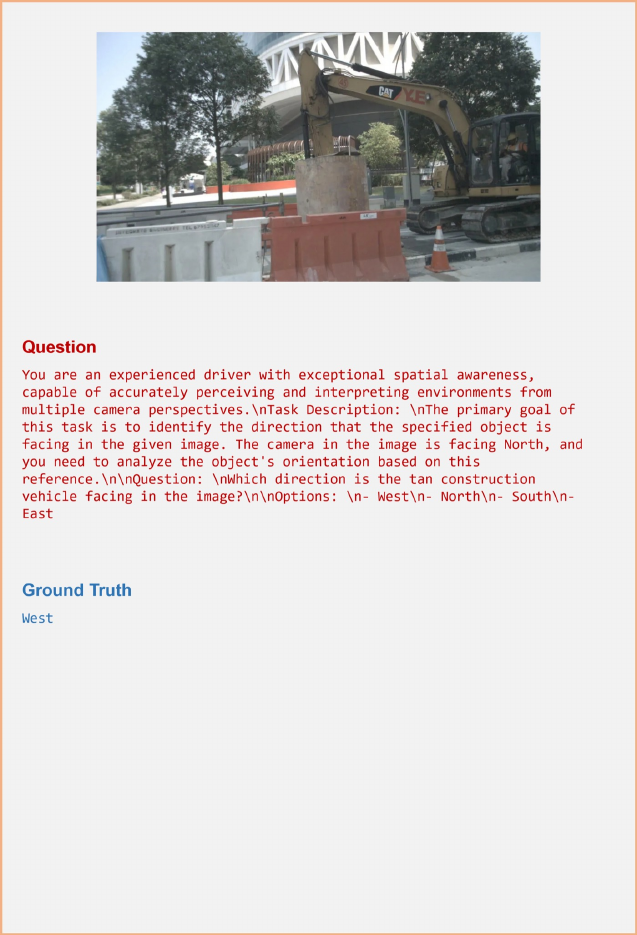}
  \caption{T2: Tiered data example from SURDS.
  }
  \label{fig:data_tiered_surds}
\end{figure*}

\begin{figure*}[tb]
  \centering
  \includegraphics[width=0.84\linewidth]{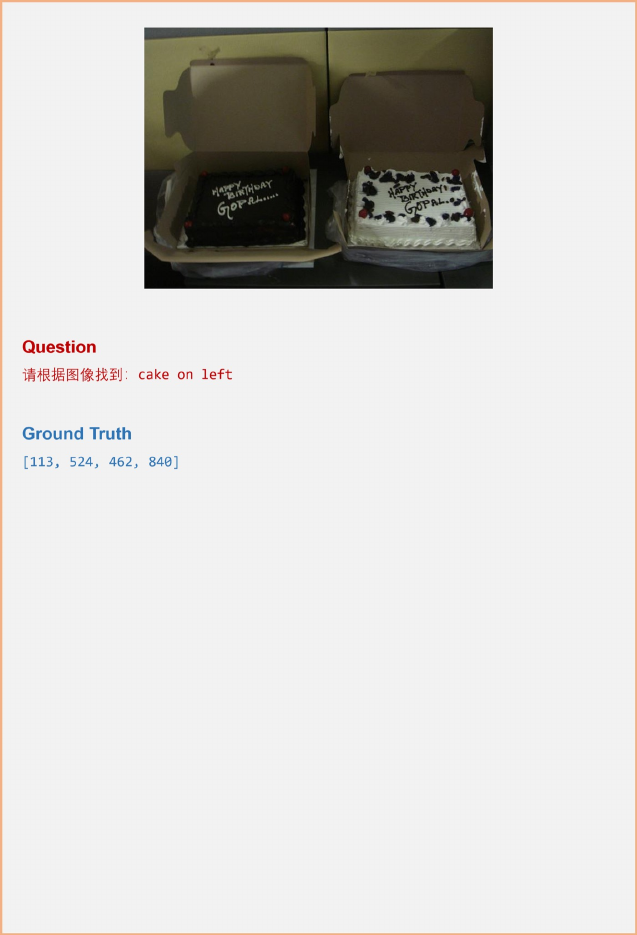}
  \caption{T2: Tiered data example from RefCOCO.
  }
  \label{fig:data_tiered_refcoco}
\end{figure*}


\begin{figure*}[tb]
  \centering
  \includegraphics[width=0.84\linewidth]{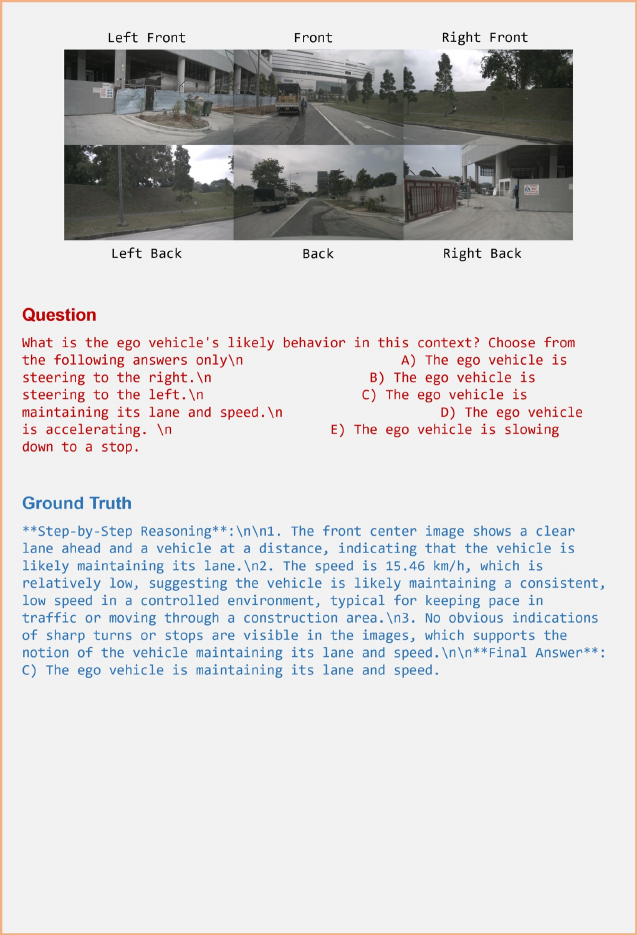}
  \caption{T3: Tiered data example from DriveLMM.
  }
  \label{fig:data_tiered_drivelmm}
\end{figure*}

\begin{figure*}[tb]
  \centering
  \includegraphics[width=0.84\linewidth]{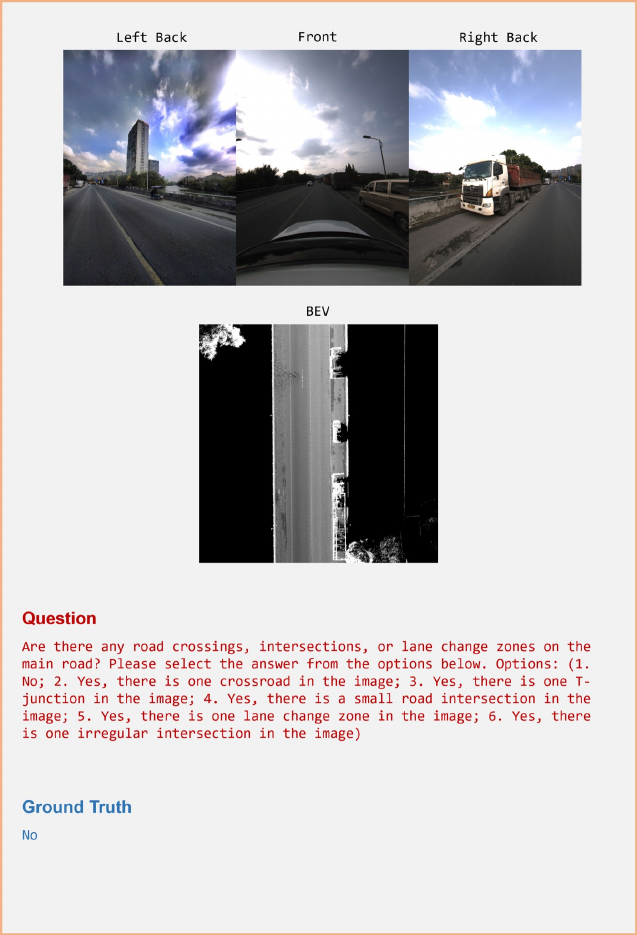}
  \caption{T3: Tiered data example from MapLMv2.
  }
  \label{fig:data_tiered_maplmv2}
\end{figure*}


\begin{figure*}[tb]
  \centering
  \includegraphics[width=0.84\linewidth]{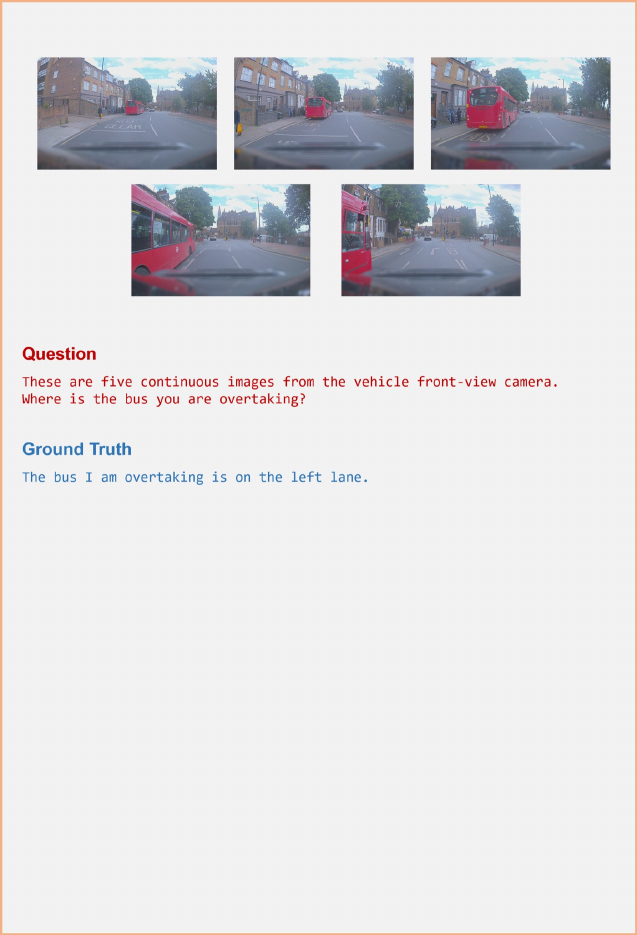}
  \caption{T4: Tiered data example from LingoQA.
  }
  \label{fig:data_tiered_lingoqa}
\end{figure*}


\begin{figure*}
  \includegraphics[width=0.84\linewidth]{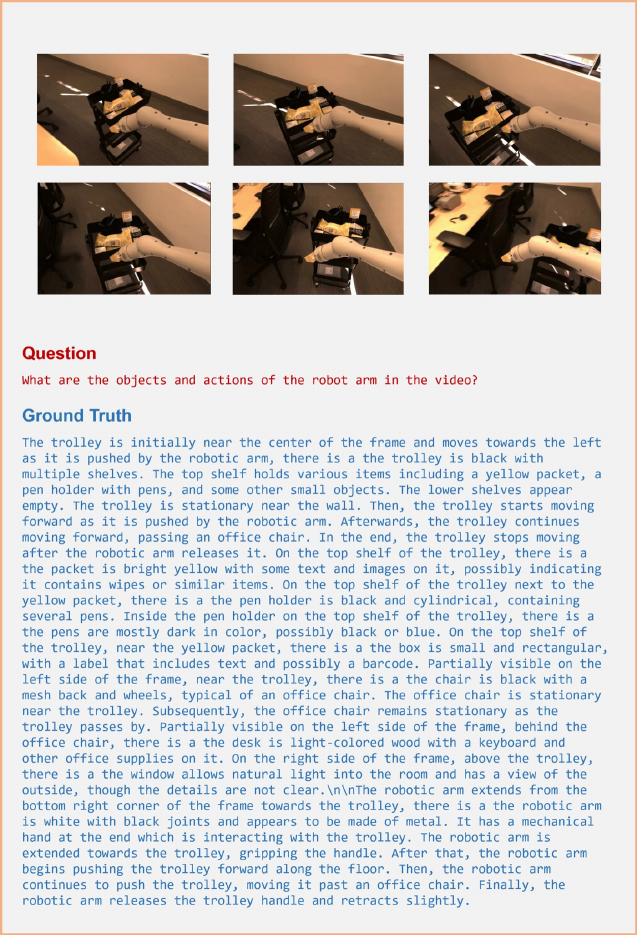}
  \caption{T4: Tiered data example from RoboVQA.
  }
  \label{fig:data_tiered_robovqa}
\end{figure*}


\subsubsection{Quality Assessment}
A Qwen3-VL-235B-based quality agent evaluates each sample across four dimensions (consistent with the main text):
\begin{enumerate}
    \item \textit{Correctness}: No factual errors or hallucinated content;
    \item \textit{Completeness}: Covers all core task information;
    \item \textit{Clarity}: No ambiguous, grammatically incorrect, or redundant language;
    \item \textit{Relevance}: No content irrelevant to task completion.
\end{enumerate}
Each dimension is scored from 0 to 1, and the total quality score is the average of the four dimensions. Samples are retained only if $\mathcal{H}(x_i) > \tau=0.2$ (filter empty/sparse scenes with no spatial information) and quality score $> \phi=0.85$ (filter low-quality annotations).

\begin{figure*}[tb]
  \centering
  \includegraphics[width=0.84\linewidth]{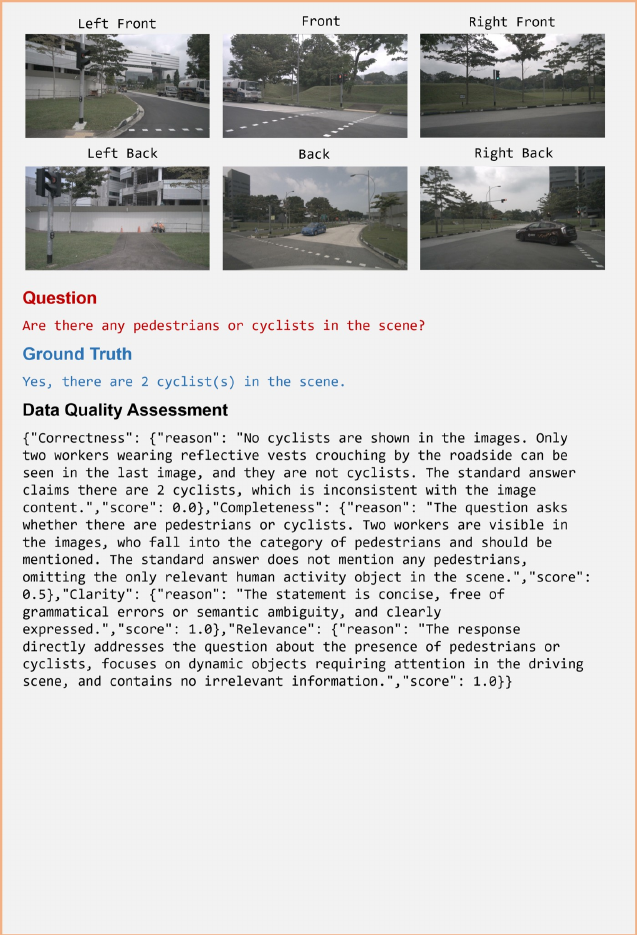}
  \caption{Data quality assessment example from OmniDrive.
  }
  \label{fig:data_filering_omnidrive}
\end{figure*}

\begin{figure*}[tb]
  \centering
  \includegraphics[width=0.84\linewidth]{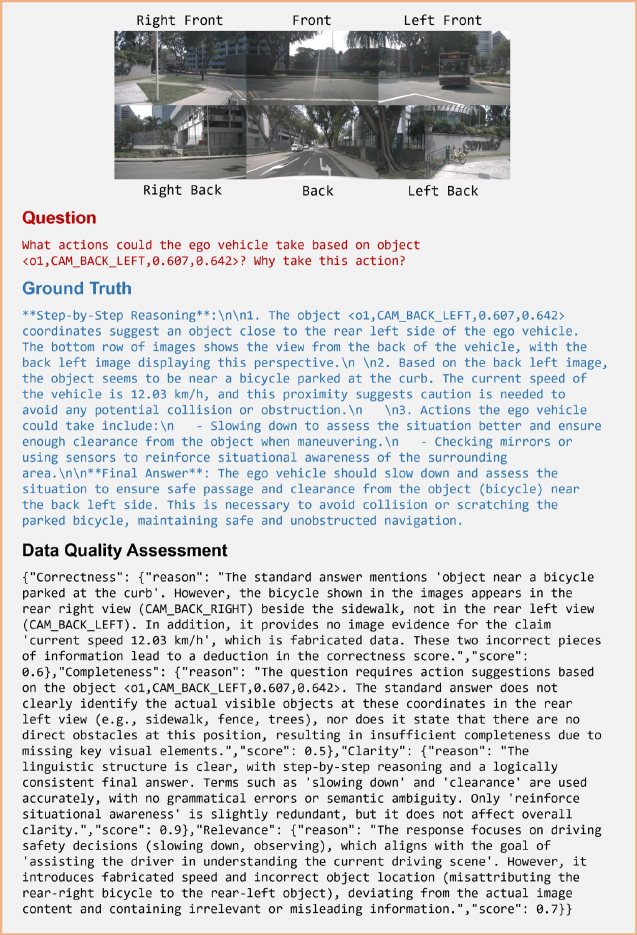}
  \caption{Data quality assessment example from DriveLMM-o1.
  }
  \label{fig:data_filering_drivelmm}
\end{figure*}

\begin{figure*}[tb]
  \centering
  \includegraphics[width=0.84\linewidth]{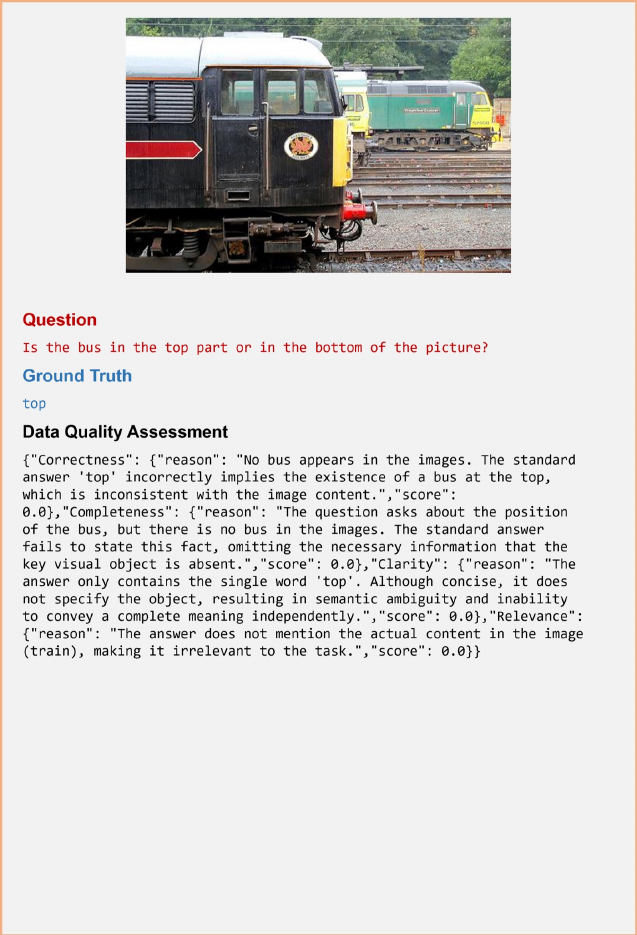}
  \caption{Data quality assessment example from GQA.
  }
  \label{fig:data_filering_GQA}
\end{figure*}

\begin{figure*}[tb]
  \centering
  \includegraphics[width=0.84\linewidth]{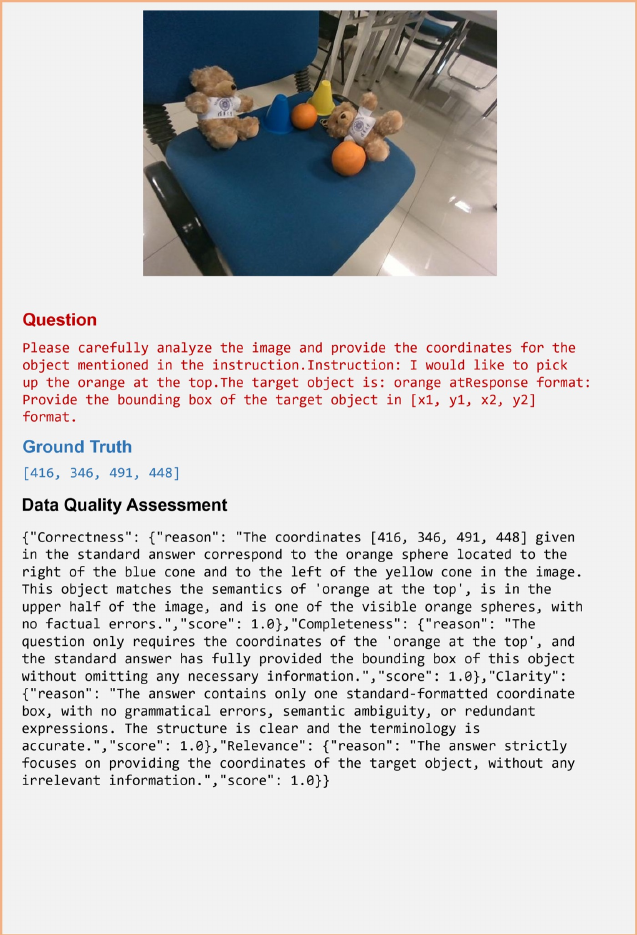}
  \caption{Data quality assessment example from Roborefit.
  }
  \label{fig:data_filering_Roborefit}
\end{figure*}

We present representative examples to demonstrate the results following data classification and filtering in Fig.~\ref{fig:data_filering_omnidrive}, Fig.~\ref{fig:data_filering_drivelmm}, Fig.~\ref{fig:data_filering_GQA}, Fig.~\ref{fig:data_filering_Roborefit}.

\subsection{Qualitative results}
\begin{figure*}[t]
  \centering
  \includegraphics[width=0.84\linewidth]{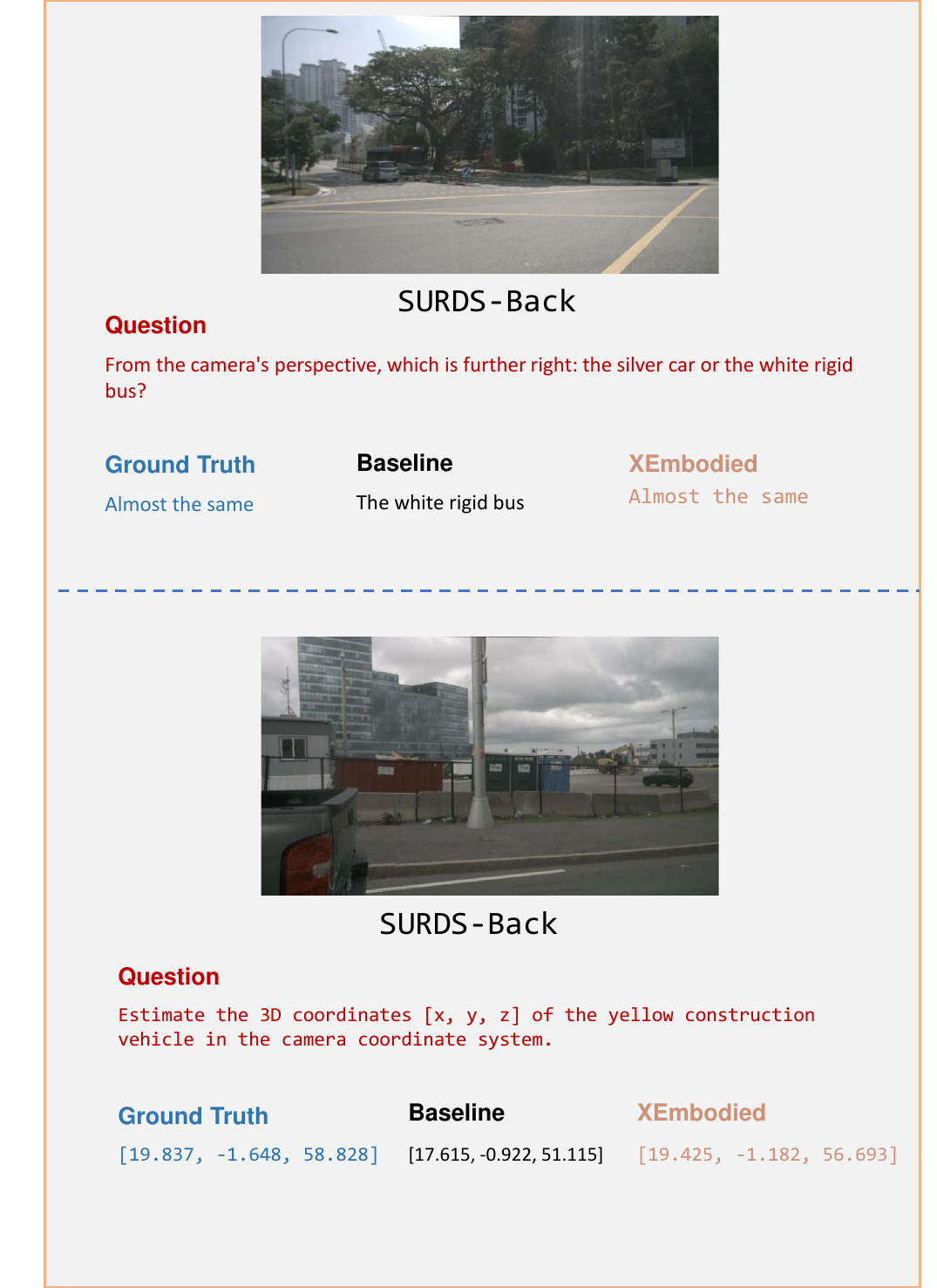}
  \caption{SURDS-Back: Spatial positioning judgment and 3D coordinate estimation}
  \label{fig:supp_qual_1}
\end{figure*}

\begin{figure*}[t]
  \centering
  \includegraphics[width=0.84\linewidth]{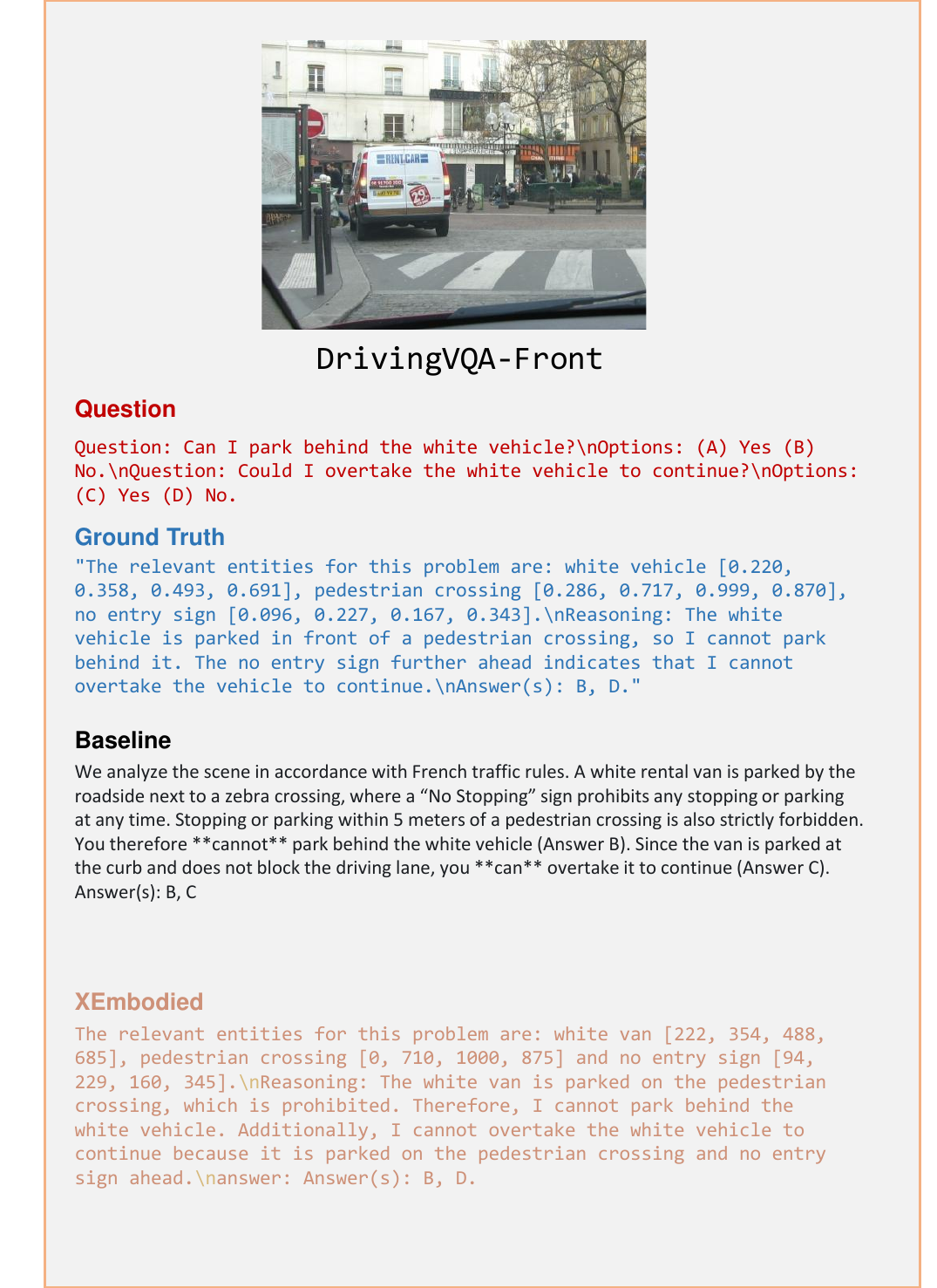}
  \caption{DrivingVQA-Front: French traffic rule reasoning for parking and overtaking}
  \label{fig:supp_qual_2}
\end{figure*}

\begin{figure*}[t]
  \centering
  \includegraphics[width=0.84\linewidth]{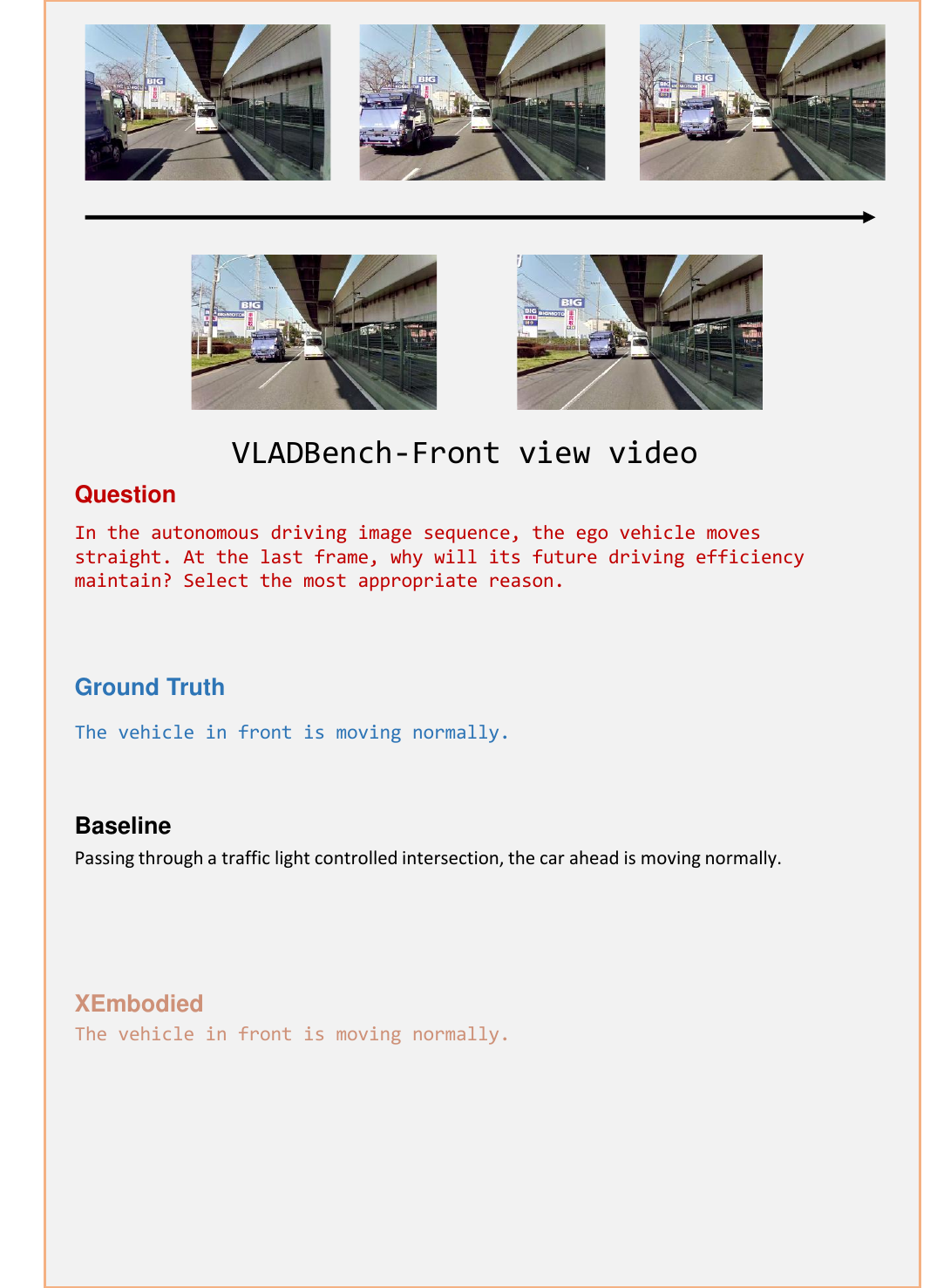}
  \caption{VLADBench-Front: Driving efficiency maintenance reason analysis}
  \label{fig:supp_qual_3}
\end{figure*}

\begin{figure*}[t]
  \centering
  \includegraphics[width=0.84\linewidth]{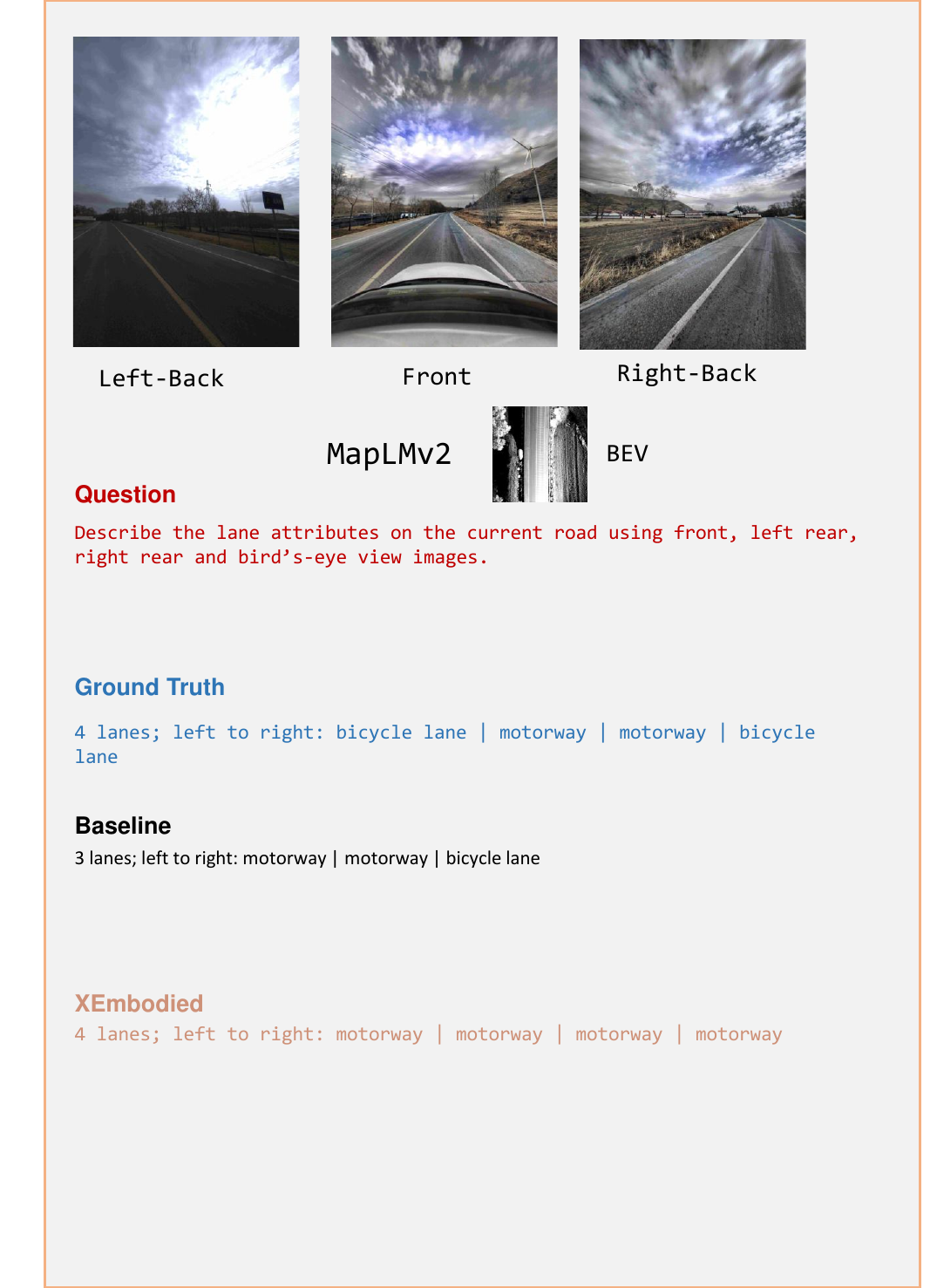}
  \caption{MapLMv2: Lane attribute description with multi-view images}
  \label{fig:supp_qual_4}
\end{figure*}

\begin{figure*}[t]
  \centering
  \includegraphics[width=0.84\linewidth]{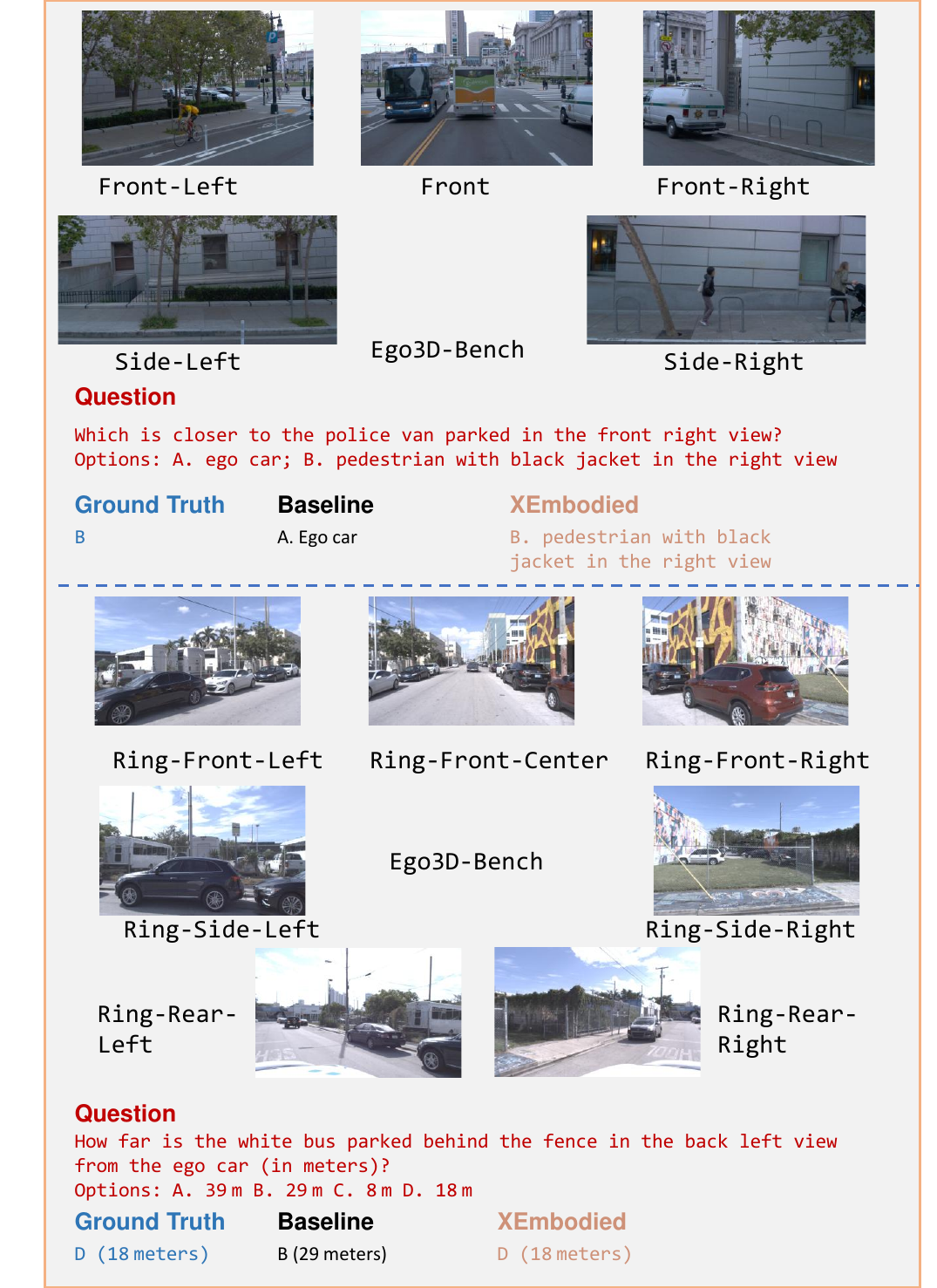}
  \caption{Ego3D-Bench: Spatial proximity judgment and distance estimation}
  \label{fig:supp_qual_5}
\end{figure*}

\begin{figure*}[t]
  \centering
  \includegraphics[width=0.84\linewidth]{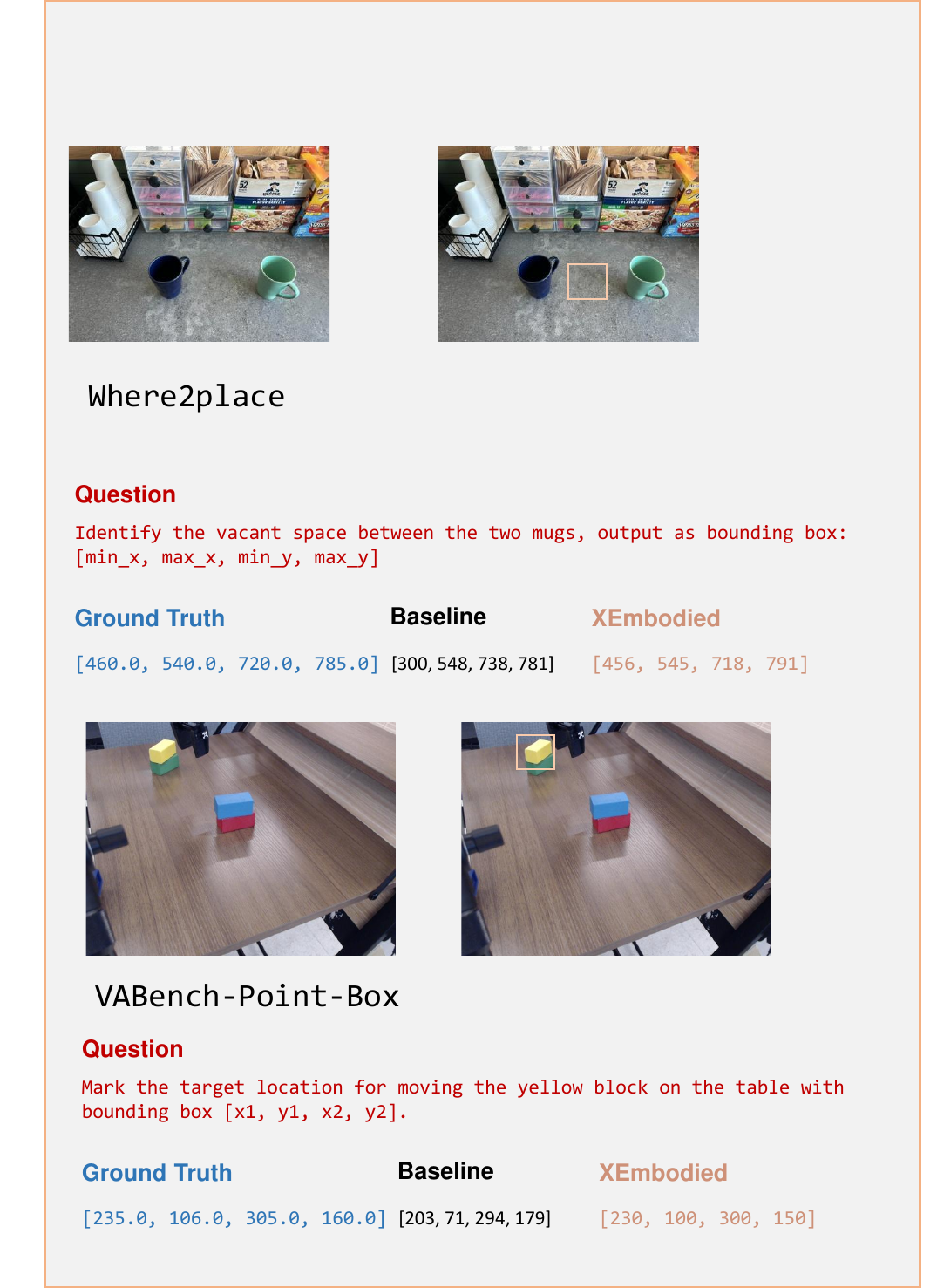}
  \caption{Where2place \& VABench-Point-Box: Robotic manipulation bounding box prediction}
  \label{fig:supp_qual_6}
\end{figure*}

\begin{figure*}[t]
  \centering
  \includegraphics[width=0.84\linewidth]{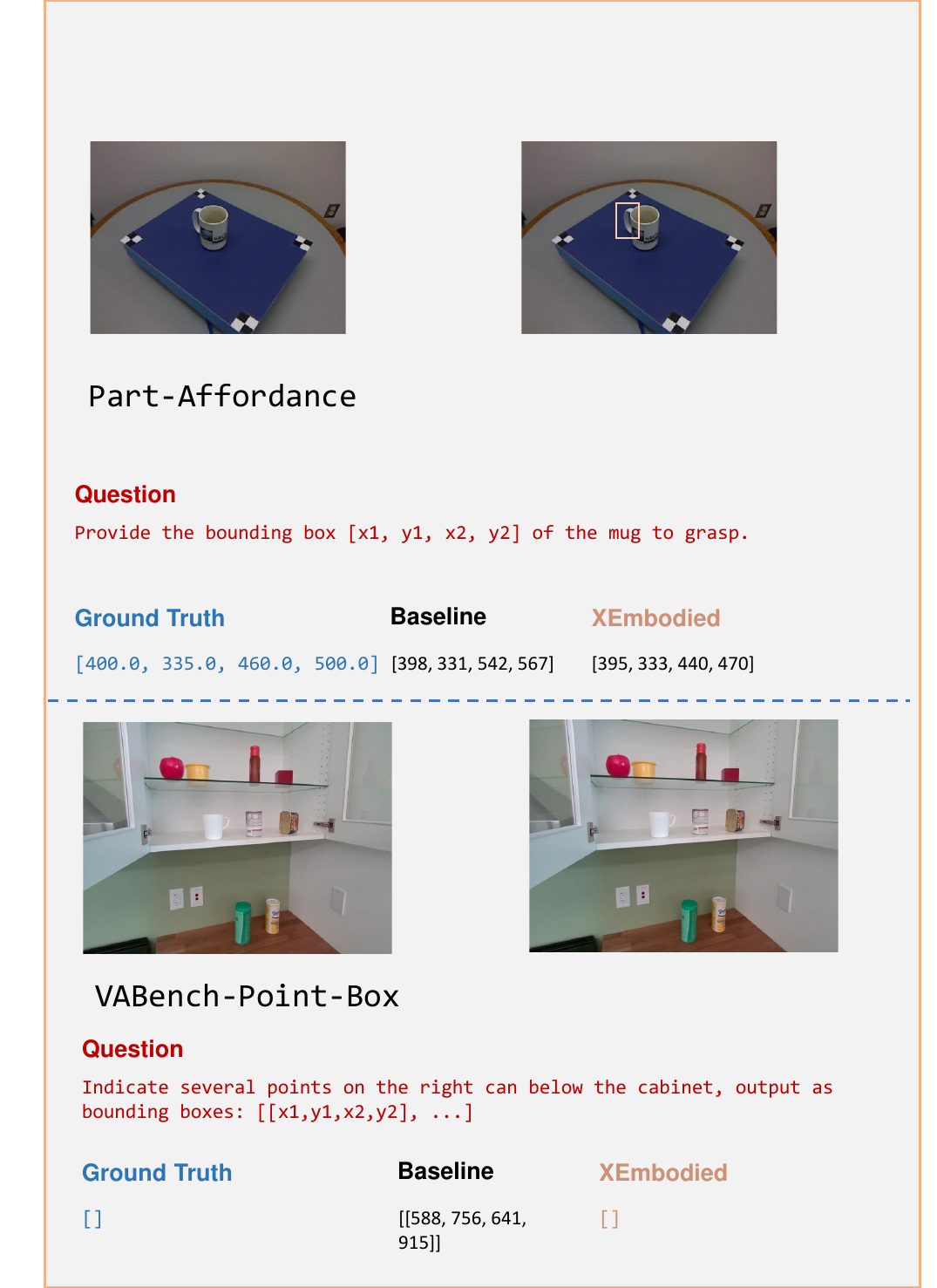}
  \caption{Part-Affordance \& VABench-Point-Box: Robotic grasping region detection}
  \label{fig:supp_qual_7}
\end{figure*}

In this section, we present comprehensive qualitative results of our XEmbodied model across diverse embodied intelligence tasks, covering autonomous driving spatial reasoning, traffic rule inference, 3D coordinate estimation, lane attribute description, and robotic manipulation fine-grained prediction. We compare XEmbodied with the baseline model against ground truth (GT) annotations on multiple benchmark datasets, and the visualized results are shown in Figs. \ref{fig:supp_qual_1} to \ref{fig:supp_qual_7}. The experimental results further verify the superiority of XEmbodied in spatial perception, scene semantic understanding, logical reasoning, and pixel-level prediction for embodied closed-loop learning tasks, while also revealing areas where the model still needs improvement.

Fig.~\ref{fig:supp_qual_1} shows the performance of XEmbodied and the baseline on the SURDS-Back dataset for two core autonomous driving spatial perception tasks. For the relative left-right positioning judgment between the silver car and the white rigid bus from the camera's perspective, XEmbodied accurately outputs the result of 'Almost the same' that is consistent with the ground truth, while the baseline misjudges the bus as being further right. For the 3D coordinate estimation of the yellow construction vehicle in the camera coordinate system, XEmbodied outputs $[19.425, -1.182, 56.693]$, which is significantly closer to the GT value $[19.837, -1.648, 58.828]$; in contrast, the baseline's prediction $[17.615, -0.922, 51.115]$ has a large deviation in all three dimensions. This demonstrates XEmbodied's strong capability in both qualitative spatial relationship judgment and quantitative 3D spatial regression.

Fig. \ref{fig:supp_qual_2} reports the traffic rule reasoning results on the DrivingVQA-Front dataset, focusing on parking and overtaking judgment under French traffic regulations. XEmbodied first accurately locates key traffic entities (white van, pedestrian crossing, no-entry sign) with pixel-level bounding box coordinates, then performs rigorous logical reasoning: it correctly infers that parking behind the van is prohibited due to the van being parked on the pedestrian crossing, and overtaking is not allowed because of the no-entry sign ahead, thus outputting the correct answers B and D that match the ground truth. The baseline only recognizes the no-stopping rule for the pedestrian crossing and the curb-parked state of the van, while ignoring the no-entry sign in the scene. As a result, it misjudges the overtaking permission and outputs incorrect answers B and C, reflecting its deficiency in comprehensive scene information extraction and multi-rule fusion reasoning.

Fig. \ref{fig:supp_qual_3} presents the reasoning result for the cause of ego vehicle driving efficiency maintenance on the VLADBench-Front view video dataset. The ground truth attributes the stable driving efficiency of the ego vehicle (moving straight) to The vehicle in front is moving normally. XEmbodied directly and accurately outputs the consistent result without any redundant information, which is crucial for concise and efficient autonomous driving decision-making. In contrast, the baseline unnecessarily adds the erroneous description of Passing through a traffic light controlled intersection that is not present in the scene, leading to an output inconsistent with the ground truth. This verifies that XEmbodied can perform scene reasoning in a fact-based and concise manner, avoiding the introduction of irrelevant or wrong semantic information.

Fig. \ref{fig:supp_qual_4} displays the lane attribute description results on the MapLMv2 dataset, where the model is required to analyze the number of lanes and their attributes from multi-view autonomous driving images (front, left rear, right rear, bird's-eye view). The ground truth defines the scene as 4 lanes with the attribute sequence bicycle lane | motorway | motorway | bicycle lane from left to right. The baseline not only undercounts the total number of lanes to 3 but also misdescribes the lane attributes, failing to recognize the bicycle lanes on both sides. By contrast, XEmbodied correctly identifies the total number of lanes as 4—a key improvement for high-precision map construction and autonomous driving path planning—but outputs the lane attribute sequence motorway | motorway | motorway | motorway, indicating a deviation in fine-grained lane-type recognition. This result reflects that while XEmbodied outperforms the baseline in basic lane quantity estimation, it still has room for improvement in detailed lane attribute classification to fully match the ground truth.

Fig. \ref{fig:supp_qual_5} shows the spatial reasoning results on the Ego3D-Bench dataset with multi-camera view inputs (front-left, front-right, side-left, side-right, ring views). Two typical spatial reasoning tasks are tested: police van relative proximity judgment and white bus distance estimation from the ego car. For the proximity judgment task, XEmbodied correctly infers that the pedestrian with black jacket in the right view (Option B) is closer to the police van, while the baseline misjudges it as the ego car (Option A). For the quantitative distance estimation task, XEmbodied accurately outputs the GT result of 18 meters (Option D) for the white bus, and the baseline outputs an erroneous 29 meters (Option B). These results fully demonstrate XEmbodied's outstanding ability in both qualitative spatial position judgment and quantitative spatial distance estimation for complex autonomous driving scenes with multi-view inputs.

Fig. \ref{fig:supp_qual_6} presents the robotic manipulation fine-grained bounding box prediction results on the Where2place and VABench-Point-Box datasets. For the vacant space detection task between two mugs on Where2place, XEmbodied outputs the bounding box $[456, 545, 718, 791]$, which is highly consistent with the GT $[460.0, 540.0, 720.0, 785.0]$ with only slight pixel-level deviations; the baseline's prediction $[300, 548, 738, 781]$ has a large offset in the min-x coordinate, leading to an inaccurate positioning of the vacant space. For the target location marking task of moving the yellow block on VABench-Point-Box, XEmbodied's bounding box $[230, 100, 300, 150]$ is almost aligned with the GT $[235.0, 106.0, 305.0, 160.0]$, while the baseline's $[203, 71, 294, 179]$ has obvious offsets in both x and y axes, failing to accurately mark the target placement position. This proves that XEmbodied has high precision in pixel-level fine-grained bounding box prediction for robotic manipulation tasks, which is essential for actual robotic operation.

Fig.~\ref{fig:supp_qual_7} reports the robotic grasping region detection results on the Part-Affordance and VABench-Point-Box datasets. For the mug grasping bounding box prediction task on Part-Affordance, XEmbodied outputs the compact and accurate box $[395, 333, 440, 470]$ that matches the GT $[400.0, 335.0, 460.0, 500.0]$, only covering the effective grasping region of the mug; the baseline's box is overly large and includes a large number of background regions, which is not conducive to actual robotic grasping operations. For the right can key point marking task under the cabinet on VABench-Point-Box, XEmbodied correctly identifies the non-target scene (no valid can in the specified area) and outputs an empty bounding box consistent with the ground truth; the baseline misdetects the background region as the target and outputs an erroneous bounding box, reflecting its poor ability in target recognition and negative sample judgment for robotic affordance tasks. XEmbodied's excellent performance in this task verifies its strong capability in identifying valid grasping regions and distinguishing non-target scenes for robotic manipulation.

\noindent Summary. The qualitative results across seven typical embodied intelligence tasks confirm XEmbodied outperforms the baseline model across most core capabilities, while also revealing specific areas for further improvement. Leveraging endogenous 3D geometric representation, physical clue interaction, and progressive domain curriculum learning, XEmbodied achieves superior spatial perception, scene semantic understanding, and logical reasoning in autonomous driving and robotic manipulation, alongside notable improvements in fine-grained prediction (with room for further refinement in certain detailed attribute recognition tasks like lane-type classification). The baseline model exhibits deficiencies like incomplete scene information extraction and low-precision prediction, verifying the rationality and effectiveness of XEmbodied's core design for embodied closed-loop learning, as well as the need for targeted optimization in fine-grained attribute recognition.
\clearpage

\bibliography{aaai2027}

\end{document}